\documentclass[sigconf]{acmart}
\usepackage{subfigure}
\usepackage{multirow}
\usepackage{threeparttable}
\usepackage{graphicx}
\usepackage{textcomp}
\usepackage{xcolor}
\usepackage{enumitem}
\usepackage{comment}
\usepackage{pifont}
\usepackage{amsthm}
\usepackage{algorithm}
\usepackage{hyperref} 
\usepackage{amsmath}
\usepackage{bm}
\usepackage{bbding}
\usepackage{amsthm}
\usepackage{algpseudocode}

\usepackage{fontawesome}

\newtheoremstyle{boldassumption}
  {\topsep}
  {\topsep}
  {\itshape}
  {}
  {\bfseries}
  {.}
  {.5em}
  {}

\theoremstyle{boldassumption}
\newtheorem{assumption}{Assumption}

\DeclareUnicodeCharacter{2217}{}

\AtBeginDocument{%
  }

\author{Yuan~Yuan}
\affiliation{
\institution{Department of Electronic Engineering, Tsinghua University}
\country{Beijing, China}
}
\email{y-yuan20@mails.tsinghua.edu.cn}

\author{Jingtao~Ding}
\affiliation{
\institution{Department of Electronic Engineering, Tsinghua University}
\country{Beijing, China}
}
\email{dingjt15@tsinghua.org.cn}

\author{Jie Feng}
\affiliation{
\institution{Department of Electronic Engineering, Tsinghua University}
\country{Beijing, China}
}
\email{fengj12ee@hotmail.com}

\author{Depeng~Jin}
\affiliation{
\institution{Department of Electronic Engineering, Tsinghua University}
\country{Beijing, China}
}
\email{jindp@tsinghua.edu.cn}

\author{Yong~Li}
\affiliation{
\institution{Department of Electronic Engineering,  Tsinghua University}
\country{Beijing, China}
}
\email{liyong07@tsinghua.edu.cn}

\copyrightyear{2024}
\acmYear{2024}
\setcopyright{rightsretained}
\acmConference[KDD '24]{Proceedings of the 30th ACM SIGKDD Conference on Knowledge Discovery and Data Mining}{August 25--29, 2024}{Barcelona, Spain}
\acmBooktitle{Proceedings of the 30th ACM SIGKDD Conference on Knowledge Discovery and Data Mining (KDD '24), August 25--29, 2024, Barcelona, Spain}\acmDOI{10.1145/3637528.3671662}
\acmISBN{979-8-4007-0490-1/24/08}

\makeatletter
\gdef\@copyrightpermission{
  \begin{minipage}{0.3\columnwidth}
   \href{https://creativecommons.org/licenses/by/4.0/}{\includegraphics[width=0.90\textwidth]{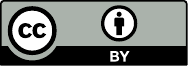}}
  \end{minipage}\hfill
  \begin{minipage}{0.7\columnwidth}
   \href{https://creativecommons.org/licenses/by/4.0/}{This work is licensed under a Creative Commons Attribution International 4.0 License.}
  \end{minipage}
  \vspace{5pt}
}
\makeatother

\begin{document}

\title{UniST: A Prompt-Empowered Universal Model for Urban Spatio-Temporal Prediction}



\begin{abstract}
Urban spatio-temporal prediction is crucial for informed decision-making, such as traffic management, resource optimization, and emergence response. Despite remarkable breakthroughs in pretrained natural language models that enable one model to handle diverse tasks, a universal solution for spatio-temporal prediction remains challenging.
Existing prediction approaches are typically  tailored for specific spatio-temporal scenarios, requiring task-specific model designs and extensive domain-specific training data. In this study, we introduce UniST, a universal model designed for general urban spatio-temporal prediction across a wide range of scenarios. Inspired by large language models, UniST achieves success through: (i) utilizing diverse spatio-temporal data from different scenarios, (ii) effective pre-training to capture complex spatio-temporal dynamics, (iii) knowledge-guided prompts  to enhance generalization capabilities. These designs together unlock the potential of building a universal model for various scenarios.
Extensive experiments on more than 20 spatio-temporal scenarios demonstrate UniST's efficacy in advancing state-of-the-art performance, especially in few-shot and zero-shot prediction. 
The datasets and code implementation are released on \textcolor{blue}{\url{https://github.com/tsinghua-fib-lab/UniST}}.
\end{abstract}



\maketitle

\section{Introduction}

Pre-trained foundation models have showcased remarkable success in Natural Language Processing (NLP)~\cite{touvron2023llama,brown2020language}, particularly excelling in few-shot and zero-shot settings~\cite{brown2020language,kojima2022large}.
However, similar breakthroughs have not yet been achieved in the field of urban spatio-temporal prediction~\cite{wang2020deep,zheng2014urban, gong2023empowering}.
In this paper, our goal is to establish a foundation model for general urban spatio-temporal prediction — specifically, to develop a universal model that offers superior performance and powerful generalization capabilities across diverse spatio-temporal scenarios. 
This entails training a single model capable of effectively handling various urban contexts, encompassing various domains such as human mobility, traffic and communication networks across different cities.

The significance of such a universal model lies in its ability to address prevalent data scarcity issues in urban areas.
The varying levels of digitalization across domains and cities  often result in imbalanced and incomplete datasets.
Despite notable advancements in existing spatio-temporal modeling approaches~\citep{bai2020adaptive, zhang2017deep, liu2018attentive, pan2019urban, zhou2023predicting, li2023learning},  their effectiveness is typically confined to specific domains within a single city.  The reliance on extensive training data further impedes the model's generalization potential. 
Consequently, current solutions are still far from ``universality'', and remain narrowly applicable.

A universal spatio-temporal model must possess two essential capabilities. \textit{Firstly, it must be capable of leveraging abundant and rich data from different urban scenarios for training.} The training of the foundational model should ensure the acquisition of ample and rich information~\cite{touvron2023llama,wang2023detecting,bai2023sequential}. 
\textit{Second, it should demonstrate robust generalization across different spatio-temporal scenarios}. Especially in scenarios with limited or no training data, the model can still work well without obvious performance degradation~\cite{garza2023timegpt,wang2023detecting}.

\begin{figure}[t]
    \centering
    \includegraphics[width=0.99\linewidth]{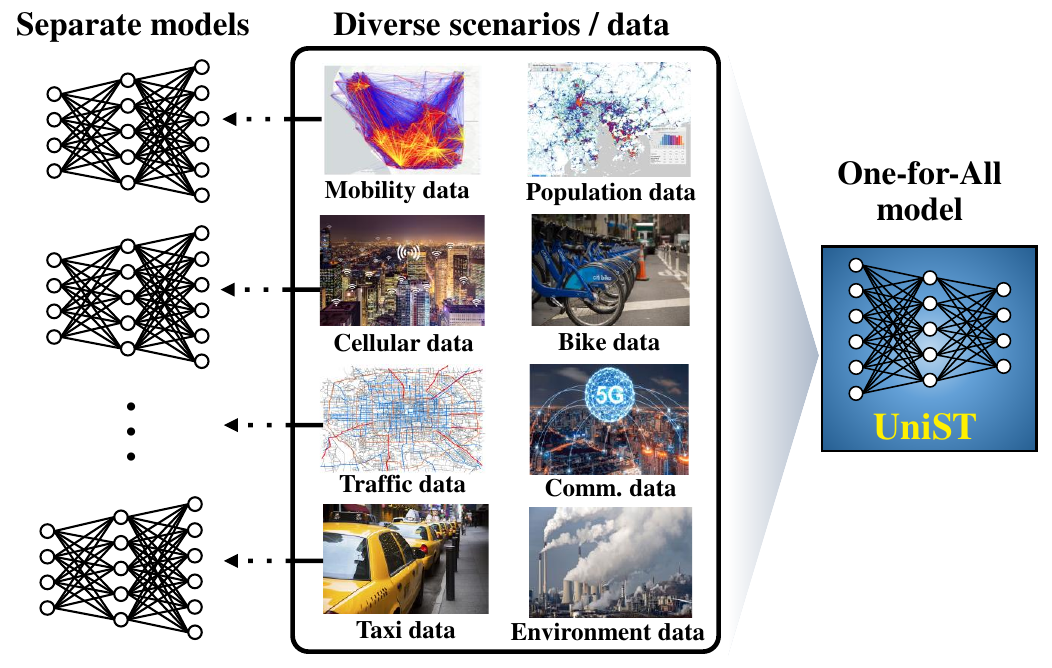}
    \caption{The transition from traditional separate deep learning models to a one-for-all universal model for urban spatio-temporal prediction.}
    \label{fig:compare_star}
\end{figure}

However, realizing the aforementioned capabilities encounters significant challenges specific to spatio-temporal data, which impede the direct application of current foundation models developed for language and vision domains.
The first challenge arises from the inherent \textit{diverse formats} of spatio-temporal datasets. Unlike languages with a natural and unified sequential structure or images and videos adhering to standardized dimensions, spatio-temporal data collected from different sources exhibit highly varied features. These include variable dimensions, temporal durations, and spatial coverages that differ significantly, posing difficulties in standardizing their structure.
The second challenge arises from \textit{high variations in data distributions across multiple scenarios}. Faced with highly distinct spatio-temporal patterns, the model may struggle to adapt to these differences. Unlike language, which benefits from a shared vocabulary, various scenarios of different domains and cities often operate on entirely different spatial and temporal scales, lacking common elements for effective training and generalization.

Although the displayed spatio-temporal patterns vary significantly, there are certain underlying laws that should be common among them. This principle arises from the intuition that human activity influences various spatio-temporal data generated in urban settings, leading to the existence of universal patterns. 
For example, traffic speed and communication networks exhibit distinct spatio-temporal patterns, yet both are influenced by human mobility and therefore adhere to similar underlying principles. 
Additionally, while temporal periodic patterns vary across domains, they share fundamental concept of repetition. 
Furthermore, city layouts vary considerably between different urban areas, but the relationships among various functional zones within cities may exhibit shared characteristics. 
Therefore,  the key to  building a one-for-all model is to capture, align and leverage these shared while underlying characteristics effectively.

To this end, we introduce \textbf{\textit{UniST}},  a \textbf{\underline{uni}}versal solution for urban  \textbf{\underline{s}}patio-\textbf{\underline{t}}emporal prediction through advanced pre-training and prompt learning. Notably, UniST achieves three essential capabilities of: 

\begin{enumerate}[leftmargin=*]
    \item  Scalability across scenarios with diverse  spatio-temporal data;
    \item  Effective pre-training to capture complex spatio-temporal relationships;
    \item  utilizing spatio-temporal prompts to align underlying shared patterns across scenarios.
\end{enumerate}

UniST achieves the above capabilities through its holistic design driven by four key components: \textit{data}, \textit{architecture}, \textit{pre-training}, and \textit{prompt learning}. 
Firstly, we harness the rich diversity inherent in spatio-temporal scenarios by leveraging extensive \textit{data} from various domains and cities.
Secondly,  we design spatio-temporal patching to unify diverse data into a sequential format, facilitating the utilization of the powerful Transformer \textit{architecture}. 
Thirdly, drawing inspiration from large language and vision models~\cite{devlin2018bert,he2022masked}, UniST adopts the widely-used generative \textit{pre-training} strategy – Masked Token Modeling (MTM).
We further enhance the model's capability to capture complex spatio-temporal relationships by employing multiple masking strategies that comprehensively address multi-perspective correlations.
Moreover, informed by the established domain knowledge in spatio-temporal modeling, we design an innovative prompt learning approach. The elaborated prompt network identifies underlying and shared spatio-temporal patterns, adapting dynamically to generate useful prompts.
In this way,  UniST  aligns distinct data distributions of various datasets and advances towards developing a one-for-all universal model.
We summarize our contributions as follows:
\begin{itemize}[leftmargin=*]
    \item To our best knowledge, this the first attempt to address universal spatio-temporal prediction by investigating the potential of a one-for-all model in diverse spatio-temporal scenarios. 
    \item  We propose UniST that harnesses data diversity and achieves universal spatio-temporal prediction through advanced pre-training and prompt learning. It has made a paradigm shift from traditional separate deep learning methods to a one-for-all model.
    \item Extensive experiments demonstrate the generality and universality of UniST. It achieves new state-of-the-art performance on various prediction tasks, particularly, superior few-shot and zero-shot capabilities.
\end{itemize}

\section{Related Work}

\subsubsection*{\textbf{Urban Spatio-Temporal Prediction.}}
Urban spatio-temporal prediction~\cite{wang2020deep,zheng2014urban} aims to model and forecast the dynamic patterns of urban activities over space and time. 
Deep learning techniques has propelled significant advancements.
A spectrum of models, including CNNs~\cite{li2018diffusion,zhang2017deep,liu2018attentive}, RNNs~\cite{wang2017predrnn, wang2018predrnn++,lin2020self}, ResNets~\cite{zhang2017deep}, MLPs~\cite{zhang2023mlpst,shao2022spatial}, GNNs~\cite{zhao2019t, bai2020adaptive,geng2019spatiotemporal}, Transformers~\cite{chen2022bidirectional,jiang2023pdformer,yu2020spatio,chen2021s2tnet},  and diffusion models~\cite{zhou2023towards, yuan2023spatio}, have been introduced to capture spatio-temporal patterns.
Simultaneously, cutting-edge techniques like meta-learning~\cite{lu2022spatio,yao2019learning}, contrastive learning~\cite{ji2023spatio,zhang2023mask}, and adversarial learning~\cite{ouyang2023citytrans,tang2022domain} are also utilized.
However, most approaches remain constrained by training separate models for each specific dataset. 
Some studies~\cite{yao2019learning,lu2022spatio,jin2022selective, liu2023cross} explore transfer learning between cities, however, a certain amount of data samples in the target city are still required. 
Current solutions are restrictive to specified spatio-temporal scenarios and require training data, while our model allows generalization across diverse scenarios and provides a one-for-all solution.

\begin{table}[t!]
\caption{Comparison of UniST with other spatio-temporal models regarding important properties. }\label{tbl:compare}
\begin{threeparttable}
\resizebox{\columnwidth}{!}{
\begin{tabular}{ccccc}
\hline
Model & Scalability\tnote{(1)}  & Few-shot & Zero-shot & Efficicency \\ \hline
PromptST~\cite{zhang2023promptst} &\textcolor{red}{\ding{55}}  & \textcolor{red}{\ding{55}} & \textcolor{red}{\ding{55}}  & \textcolor{green}{\ding{51} }  \\
GPT-ST~\cite{li2023generative} &  \textcolor{red}{\ding{55}}  & \textcolor{red}{\ding{55}}  &\textcolor{red}{\ding{55}}  & \textcolor{green}{\ding{51} }  \\
STEP~\cite{ShaoZWX22} & \textcolor{red}{\ding{55}}  & \textcolor{red}{\ding{55}} &\textcolor{red}{\ding{55}}  & \textcolor{green}{\ding{51} }   \\
ST-SSL~\cite{ji2023spatio} & \textcolor{red}{\ding{55}} & \textcolor{red}{\ding{55}}  &\textcolor{red}{\ding{55}}  & \textcolor{green}{\ding{51} }  \\
TrafficBERT~\cite{jin2021trafficbert} &   \textcolor{green}{\ding{51} }   & \textcolor{red}{\ding{55}} &\textcolor{red}{\ding{55}}  & \textcolor{green}{\ding{51} }  \\
TFM~\cite{wang2023building} &\textcolor{red}{\ding{55}}  & \textcolor{red}{\ding{55}} & \textcolor{red}{\ding{55}}  & \textcolor{green}{\ding{51} }  \\
UrbanGPT~\cite{li2024urbangpt} &   \textcolor{green}{\ding{51} }   &  \textcolor{green}{\ding{51} }\tnote{(2)}  &   \textcolor{green}{\ding{51}}\tnote{(2)}   & \textcolor{red}{\ding{55}} \\
STG-LLM~\cite{liu2024large} &\textcolor{red}{\ding{55}}  & \textcolor{red}{\ding{55}}  & \textcolor{red}{\ding{55}}  & \textcolor{red}{\ding{55}}  \\ \hline
UniST &  \textcolor{green}{\ding{51} }   &  \textcolor{green}{\ding{51} }   & \textcolor{green}{\ding{51} }   & \textcolor{green}{\ding{51} }   \\ \hline  
\end{tabular}}
\begin{tablenotes}
    \footnotesize
    \item[(1)] Whether can leverage diverse datasets with diverse formats.
    \item[(2)] Restricted in the same city.
\end{tablenotes}
\end{threeparttable}
\end{table}

\subsubsection*{\textbf{Foundation Models for Spatio-temporal Data and Time Series.}}
Inspired by the remarkable strides in foundation models for NLP~\cite{touvron2023llama,brown2020language} and CV~\cite{rombach2022high,bai2023sequential}, foundation models for urban prediction have emerged recently.
Some explorations unlock the potential of large language models (LLMs) in this context.
Intelligent urban systems like CityGPT~\cite{feng2014citygpt,xu2023urban}, CityBench~\cite{feng2014citybench} and UrbanGPT~\cite{li2024urbangpt} have demonstrated proficiency in addressing language-based tasks.
Additionally, LLMs are utilized  for describing urban-related images~\cite{yan2023urban} to benefit downstream tasks and predict user activities~\cite{gongKDDPITuning}. 
Moreover, the application of LLMs extends to  traffic signal control~\cite{lai2023large}, showcasing their utility in tackling complex spatio-temporal problems beyond languages.
Recently, there also has been great progress in foundation models for time series~\cite{jin2023large,cao2023tempo,jin2023time, zhou2023one}. Unlike time series characterized by a straightforward sequential structure, spatio-temporal data presents a more intricate nature with intertwined dependencies across both spatial and temporal dimensions.
While exploring the integration of LLMs is promising, it's important to recognize that spatio-temporal data is not inherently generated by language. Thus, developing foundation models specifically trained on pure spatio-temporal data is also an important direction.
In Table~~\ref{tbl:compare}, we compare the essential properties of UniST with other approaches employing pre-training, prompt learning, or LLMs. UniST encompasses all these essential capabilities, whereas other approaches have certain limitations.

\subsubsection*{\textbf{Prompt Learning}}
Prompt learning has achieved superior performance in large models~\cite{liu2023pre,qiao2022reasoning,jia2022visual,shen2024multitask}, with the goal of enhancing the generalization capability of pretrained models on specific tasks or domains.
 Typically, language models usually use a limited number of demonstrations as prompts and vision models often employ a learnable prompt network to generate useful prompts, known as prompt learning. 
Our research aligns with prompt learning, where spatio-temporal prompts are adaptively generated based on spatio-temporal patterns through a prompt network.

\begin{figure*}[t!]
    \centering
    \includegraphics[width=0.99\linewidth]{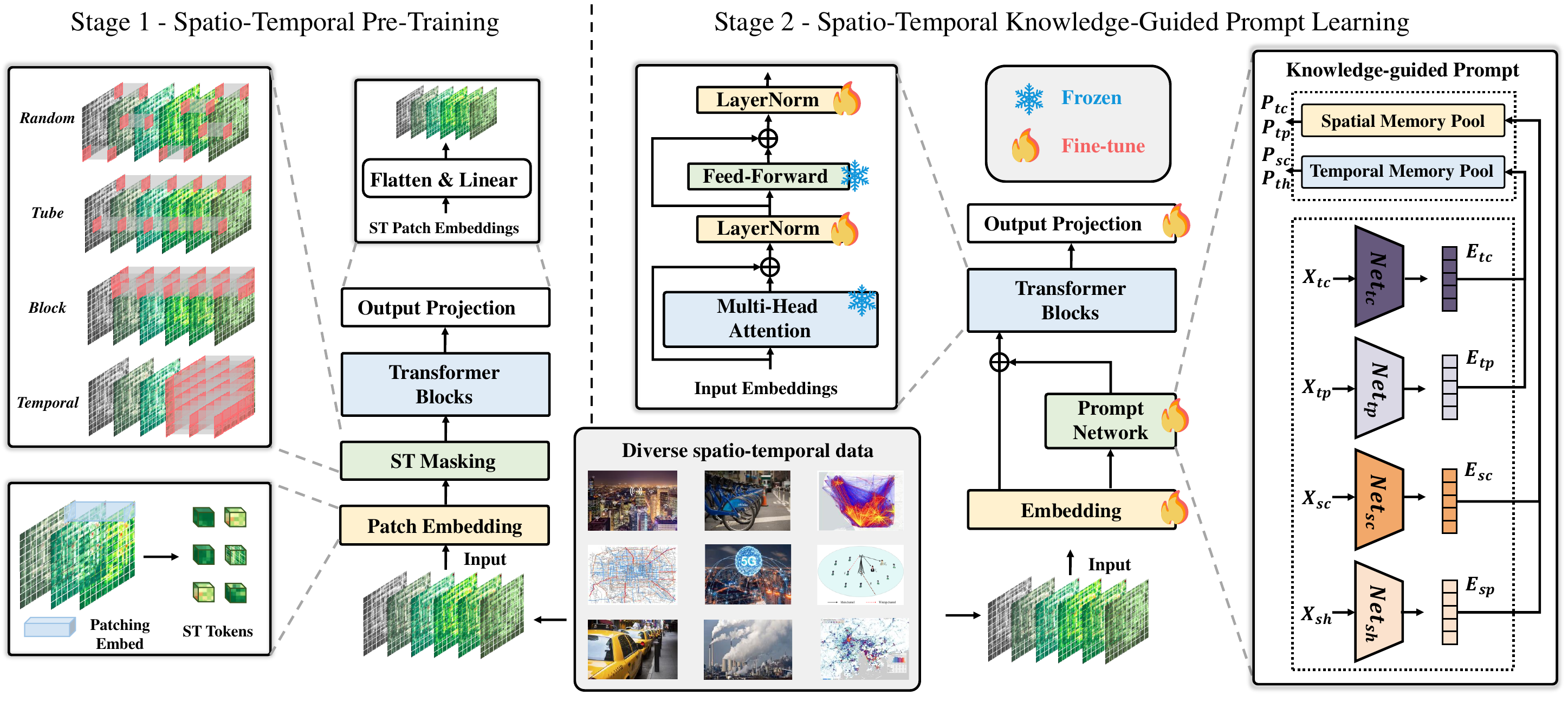}
    \caption{The overview architecture of UniST, which consists of two stages: (i) large-scale spatio-temporal pre-trianing, (ii) spatio-temporal knowledge-guided prompt learning.}
    \label{fig:model}
\end{figure*}

\section{Methodology}

\subsection{Preliminary}


\textbf{Spatial and Temporal Partitions.}  We use a grid system for spatial partitioning, dividing the city into equal, non-overlapping areas defined by longitude and latitude on an $H \times W$ map. For each area, the temporal dynamics are recorded at certain intervals.

\noindent\textbf{Spatio-Temporal Data.} A spatio-temporal data $X$ is defined as a four-dimensional tensor with dimensions $T \times C \times H \times W$, where $T$ represents time steps, $C$ represents the number of variables, $H$ and $W$ represent spatial grids. $T$, $C$, $H$, and $W$ can vary across different spatio-temporal scenarios.

\noindent\textbf{Spatio-Temporal Prediction.}
For a specific dataset, given $l_h$ historical observations for the grid map, we aim to predict the future $k$ steps. The spatio-temporal prediction task can be formulated as learning a $\theta$-parameterized model $\mathcal{F}$: $X_{[t:t+k]} = \mathcal{F}_\theta(X_{[t-l_h:t]})$.

\noindent\textbf{Few-Shot and Zero-Shot Predictions.}
The model is trained on multiple source datasets and then adapted to a target dataset. In few-shot learning, it is fine-tuned with a small amount of target samples; in zero-shot learning, it makes predictions without any fine-tuning.

\subsection{Pre-training and Prompt Learning}

Universal spatio-temporal prediction aims to empower a single model to effectively handle diverse spatio-temporal scenarios, requiring the unification of varied spatio-temporal data within a cohesive model.  This necessitates addressing significant distribution shifts across datasets of different scenarios. To achieve this goal, we propose a framework for pre-training and prompt learning, leading to a universal prediction model, UniST. Figure~\ref{fig:model} shows the overview architecture, detailing UniST with two stages:

\begin{itemize}[leftmargin=*]
    \item \textbf{Stage 1: Large-scale spatio-temporal pre-training.} Different from existing methods limited to a single dataset,  our approach utilizing extensive spatio-temporal data from a variety of domains and cities for pre-training.  

    \item \textbf{Stage 2: Spatio-temporal knowledge-guided prompt learning.} We introduces a prompt network for in-context learning, where the generation of prompts is adaptively guided by well-developed spatio-temporal domain knowledge, such as spatial hierarchy and temporal periodicity. 
\end{itemize}


\subsection{Base Model}

Our base model is a Transformer-based encoder-decoder architecture. Through spatio-temporal patching, it can handle diverse spatio-temporal data in a unified sequential format.  

\textbf{Spatio-Temporal Patching.}
The conventional Transformer architecture is designed for processing 1D sequential data. However, spatio-temporal data possesses a 4D structure.
 To accommodate this, we first split the data into channel-independent instances, which are 3D tensors. Then, we utilize spatio-temporal patching to transform the 3D tensor, denoted as $X \in \mathbb{R}^{L \times H \times W}$, into multiple smaller 3D tensors. 
If the original shape is $L\times H\times W$, and  the patch size is $(l, h, w)$,  the resulting sequence is given by $E_x \in \mathbb{R}^{L'\times H' \times W'}, L'=\frac{L}{l}, H' = \frac{H}{h} , W'= \frac{W}{w}$.

This transformation involves a 3D convolutional layer with a kernel size and stride both set to $(l, h, w)$. The process can be expressed as $E_x = \textsc{Conv}_{3d}(X)$,  where $E_x$ represents the converted 1D sequential data. The sequence length of $E_x$ is $L' \times H' \times W'$.

\textbf{Positional Encoding.}
As the original Transformer architecture does not consider the order of the sequence, we follow the common practice that incorporate positional encoding~\cite{devlin2018bert}. To enhance generalization, we choose sine and cosine functions rather than learnable parameters for positional encoding. 
This encoding is separately applied to the spatial and temporal dimensions.



\textbf{Encoder-Decoder Structure.}
The base model utilizes an encoder-decoder framework inspired by Masked Autoencoder (MAE)~\cite{he2022masked}. It processes input patches with a certain masking ratio, where the encoder takes the unmasked patches and the decoder reconstructs the image using the encoder's output and the masked patches. Our focus is on capturing comprehensive spatio-temporal dependencies, including both high-level and low-level relationships, with the goal of accurately predicting values at specific time and space coordinates. Unlike MAE, which uses a lightweight decoder for pre-training, our model employs a full-sized decoder that plays a crucial role in both pre-training and fine-tuning. It can be formulated as:

\begin{equation}
        E_{enc} =\textsc{Encoder}(E_x),\  Y_{dec} = \textsc{Decoder}(E_{enc}, E_{mask}),  \nonumber
\end{equation}

\noindent where $E_{mask}$ denotes the token embeddings for the masked patch.

\subsection{Spatio-Temporal Self-Supervised Pre-train}

In pretrained language models, the self-supervised learning task is either masking-reconstruction~\cite{devlin2018bert} or autoregressive prediction~\cite{brown2020language}.
Similarly, in vision models,  visual patches are randomly masked and the pre-training objective is to reconstruct the masked pixels. 
To further augment the model's capacity to capture intricate spatio-temporal relationships and intertwined dynamics, we introduce four distinct masking strategies during the pre-training phase, which are shown in the left box in the stage 1 of Figure~\ref{fig:model}. Suppose the masking percentage is $r$, we explain these strategies as follows:

\begin{itemize}[leftmargin=*]

    \item \textbf{Random masking.} This strategy is similar to the one used in MAE, where spatio-temporal patches are randomly masked. Its purpose is to capture fine-grained spatio-temporal relationships. 
    \begin{equation}
        M \sim \textbf{U}[0,1], \  E_x = E_x[M<1-r] , \ M \in \mathbb{R}^{L'\times H' \times W'}.  \nonumber
    \end{equation}
    
    \item \textbf{Tube masking.}  This strategy simulates scenarios where data for certain spatial units is entirely missing across all instances in time, mirroring real-world situations where some sensors may be nonfunctional—a common occurrence. The goal is to improve spatial extrapolation competence. 
    \begin{equation}
        M \sim \textbf{U}[0,1], \  E_x = E_x[\ :\ , M<1-r] , \ M \in \mathbb{R}^{ H' \times W'}.  \nonumber
    \end{equation} 
    
    \item \textbf{Block masking.}  A more challenging variant of tube masking, block masking involves the complete absence of an entire block of spatial units across all instances in time. The reconstruction task becomes more intricate due to limited context information, with the objective of enhancing spatial transferability.
    
   \begin{equation}
        M \sim \textbf{\textsc{Uniform}}(1,2), \  E_x = E_x[\ :\ ,\frac{M-1}{2}H':\frac{M}{2}H',\frac{M-1}{2}W':\frac{M}{2}W'].  \nonumber
    \end{equation} 

    \item \textbf{Temporal Masking.} In this approach, future data is masked, compelling the model to reconstruct the future based solely on historical information. The aim is to refine the model's capability to capture temporal dependencies from the past to the future.
    \begin{equation}
        M = \textsc{Concat}([\textbf{1}_{(1-r)L'\times H' \times W'}, \textbf{0}_{rL'\times H' \times W'}]), \ E_x = E_x[M=1].  \nonumber
    \end{equation}
    
\end{itemize}

By employing these diverse masking strategies, the model can systematically enhance its modeling capabilities from a comprehensive perspective, simultaneously addressing spatio-temporal, spatial, and temporal relationships.

\subsection{Spatio-Temporal Knowledge-Guided Prompt}

Prompt learning plays a critical role in enhancing UniST's generalization ability. Before delving into the details of our prompt design,  it is essential to discuss why pre-trained models can be applied to unseen scenarios.

\subsubsection{\textbf{Spatial-Temporal Generalization.}} In urban prediction tasks, the distributions of features and labels differ across domains and cities, denoted as $X_A\neq X_B, Y_A\neq Y_B$, where $X$ and $Y$ denote features and labels, while $A$ and $B$ represent different cities or domains. Taken $A$ and $B$ as a simple example, generalization involves leveraging knowledge acquired from the $A$ dataset and adapt it to the $B$ dataset.
The key point lies in identifying and aligning ``related'' patterns between $A$ and $B$ datasets. 
While finding similar patterns for an entire dataset may be challenging, we claim that identifying and aligning fine-grained patterns is feasible.
Specifically, we provide some assumptions that applies to prompt-empowered spatio-temporal generalization, which are expressed as follows:

\begin{assumption}
For a new dataset $\bf{B}$, it is possible to identify fine-grained patterns related to the training data $\bf{A}$.
 \begin{equation}
     \begin{aligned}
         &X_A\neq X_B,\  Y_A\neq Y_B, \\
         &\exists x_a \in X_A, y_a \in Y_A, \ \exists x_b \in X_B, y_b \in Y_B,  : x_a \approx x_b, y_a \approx y_b. \nonumber
     \end{aligned}
 \end{equation}
\end{assumption}

\begin{assumption}
Distinct spatio-temporal patterns correspond to customized prompts. 

\begin{equation}
    \begin{aligned}
        & P_i^* \neq P_j^* \quad \text{if} \quad D(x_i, x_j) > \epsilon, \\
        & D(P_i^*, P_j^*)>D(P_m^*, P_n^*) \quad \text{if} \quad D(x_i, x_j) > D(x_m, x_n),  \nonumber
    \end{aligned}
\end{equation}

\noindent where $x_i$ denotes the fine-grained spatio-temporal pattern,  $P_i^*$ represents the prompt of  $x_i$, and $D$ is the similarity  between $x_i$ and $x_j$.

\end{assumption}

\begin{assumption}
    There exists $f_\theta$ that captures the mapping relationship from the spatio-temporal pattern $x_i$ to prompt $P_i^*$. 
    $$P_i = f_\theta(x_i) \quad \text{where} \quad \theta = \underset{\theta}{\text{argmin}} \sum_i \textsc{Distance}(P_i^*, f_\theta(x_i)).$$ \nonumber
\end{assumption}

Based on these assumptions, our core idea is that for different inputs with distinct spatio-temporal patterns, customized prompts should be generated adaptively.

\subsubsection{\textbf{Spatio-Temporal Domain Knowledge}}

Given the aforementioned assumptions, a critical consideration is how to define the concept of ``similarity''  to  identify and align shared spatio-temporal patterns. 
Here we leverage insights from well-established domain knowledge in spatio-temporal modeling~\cite{zheng2014urban,zhang2017deep}, encompassing properties related to both space and time.
There are four aspects to consider when examining these properties:

\begin{itemize}[leftmargin=*]
    \item Spatial closeness: Nearby units may influence each other.
    \item Spatial hierarchy: The spatial hierarchical organization impacts the spatio-temporal dynamics, requiring a multi-level perception on the city structure.
    \item Temporal closeness: Recent dynamics affect future results, indicating a closeness dependence.
    \item Temporal period:  Daily or weekly patterns exhibit similarities, displaying a certain periodicity.
\end{itemize}

\begin{figure}[t]
    \centering
    \includegraphics[width=0.99\linewidth]{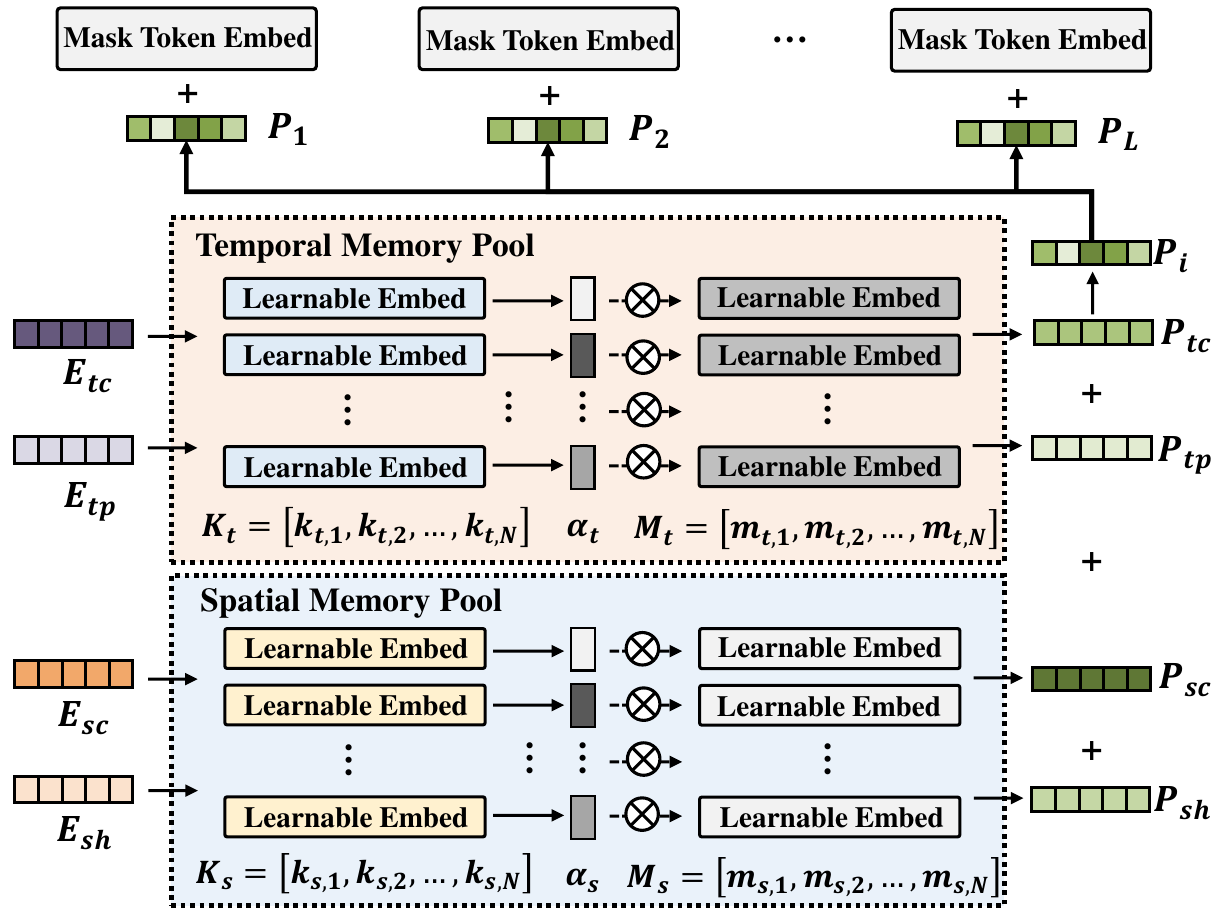}
    \caption{Illustration of the prompt generation process. }
    \label{fig:prompt}
\end{figure}

 For simplicity, we provide some straightforward implementations, which are shown in the four networks in Figure~\ref{fig:model}, \textit{i.e., } $\textsc{Net}_{tc}$, $\textsc{Net}_{tp}$, $\textsc{Net}_{sc}$, and $\textsc{Net}_{sh}$.
For the spatial dimension, we first employ an attention mechanism to merge the temporal dimension into a representation termed $E_{s}$.  Then, to capture spatial dependencies within close proximity, a two-dimensional convolutional neural network (CNN), \textit{i.e.,} $\textsc{Net}_{sc}$,  with a kernel size of 3 is employed. 
To capture spatial hierarchies, we utilize CNNs with larger kernel sizes,  \textit{i.e.,} $\textsc{Net}_{sh}$. These larger kernels enable the perception of spatial information on larger scales, which facilitate to construct a hierarchical perspective.
As for the temporal dimension, we employ an attention network, \textit{i.e.,} $\textsc{Net}_{tc}$, to  aggregate the previous M steps denoted as $X_c$. Regarding the temporal period, we select corresponding time points from the previous N days, denoted as $X_p$. Subsequently, we employ another attention network, \textit{i.e.,} $\textsc{Net}_{tp}$,  to aggregate the periodical sequence, which captures long-term temporal patterns. 
The overall process is formulated as follows:

\begin{align}
    & E_{sc} =  \textsc{Conv}_{2D}[3](X_s), \nonumber \\ 
    & E_{sh} = \{\textsc{Conv}_{2D}[2^i+1](X_s)\}, i \in\{2,3,4\}, \nonumber \\ 
    & E_{tc} = \textsc{Attention}(X_c), \nonumber \\ 
    & E_{tp} = \textsc{Attention}(X_p). \nonumber
\end{align}

\noindent It is essential to emphasize that the learning of $E_{sc}, E_{sh} , E_{tc}$, and $E_{tp} $ is not restricted by our practice. Practitioners have the flexibility to employ more complex designs to capture richer spatio-temporal properties. For example, Fourier-based approaches~\cite{wu2022timesnet,liu2023koopa} can be utilized to capture periodic patterns.

\subsubsection{\textbf{Spatio-Temporal Prompt Learner}}
Given the representations of properties derived from spatio-temporal domain knowledge, the pivotal question is how to generate prompts—\textit{how does spatio-temporal knowledge guide prompt generation?}
Here we utilize prompt learning techniques. 
While  prompt learning in computer vision~\cite{jia2022visual} often train fixed prompts for specific tasks such as segmentation, detection, and classification. 
Due to the high-dimensional and complex nature of spatio-temporal patterns, training a fixed prompt for each case becomes impractical. 

To tackle this issue, we draw inspirations from memory networks~\cite{sukhbaatar2015end} and propose a novel approach that learns a spatial memory pool and a temporal memory pool. In the prompt learning process, these memory pools are optimized to store valuable information about spatio-temporal domain knowledge. As shown in Figure~\ref{fig:prompt}, the spatial and memory pools are defined as follows:

\begin{equation}
\begin{aligned}
    & KM_s = \{(k_{s,0}, m_{s,0}), (k_{s,1}, m_{s,1}), ...,  (k_{s,N-1}, m_{s,N-1})\}, \\
    & KM_t = \{(k_{t,0}, m_{t,0}), (k_{t,1}, m_{t,1}), ...,  (k_{t,N-1}, m_{t,N-1})\}, \nonumber
\end{aligned}
\end{equation}

\noindent where $k_{s,i}, m_{s,i}, k_{t,i}, m_{t,i}, i\in\{0,1,...,N-1\}$ are all learnable parameters, and the memory is organized in a key-value structure following existing practice~\cite{sukhbaatar2015end,wang2022learning}.

\begin{table*}[t!]
\caption{Performance comparison of short-term prediction on seven datasets in terms of MAE and RMSE. We use the average prediction errors over all prediction steps. Bold denotes the best results and \underline{underline} denotes the second-best results.}
\label{tbl:short_term}
\begin{threeparttable}
\resizebox{2.1\columnwidth}{!}{
\begin{tabular}{ccccccccccccccc}
\toprule
& \multicolumn{2}{c}{\textbf{TaxiBJ}} & \multicolumn{2}{c}{\textbf{Crowd}} & \multicolumn{2}{c}{\textbf{Cellular}} & \multicolumn{2}{c}{\textbf{BikeNYC}}  & \multicolumn{2}{c}{\textbf{TrafficJN}} & \multicolumn{2}{c}{\textbf{TDrive}} &\multicolumn{2}{c}{\textbf{TrafficSH}} \\ 
 \cmidrule(lr){2-3} \cmidrule(lr){4-5} \cmidrule(lr){6-7} \cmidrule(lr){8-9} \cmidrule(lr){10-11} \cmidrule(lr){12-13}  \cmidrule(lr){14-15}
 \textbf{Model} & \textbf{RMSE}  & \textbf{MAE}    & \textbf{RMSE}       & \textbf{MAE}      & \textbf{RMSE}       & \textbf{MAE}      & \textbf{RMSE}    & \textbf{MAE}      & \textbf{RMSE}    & \textbf{MAE}    & \textbf{RMSE}  & \textbf{MAE}         & \textbf{RMSE}  & \textbf{MAE}        \\ 
\cmidrule(lr){1-1} \cmidrule(lr){2-3} \cmidrule(lr){4-5} \cmidrule(lr){6-7} \cmidrule(lr){8-9} \cmidrule(lr){10-11} \cmidrule(lr){12-13}  \cmidrule(lr){14-15}

  HA & 53.77 & 29.82	& 17.80 &6.79&72.94&27.57	& 11.41&3.43 & 1.38 & 0.690 & 150.2 & 74.5 &	1.24 & 0.771 \\
 ARIMA  & 56.70 & 39.53 & 21.87 & 10.23 & 81.31 & 40.22 & 12.37 & 3.86 & 1.20 & 0.651 & 211.3 & 108.5 & 1.17 & 0.769  \\
\cmidrule(lr){1-1} \cmidrule(lr){2-3} \cmidrule(lr){4-5} \cmidrule(lr){6-7} \cmidrule(lr){8-9} \cmidrule(lr){10-11} \cmidrule(lr){12-13}   \cmidrule(lr){14-15}

 STResNet & 45.17 & 30.87& 	5.355 &  3.382 &  24.30 & 
 14.32 & 	8.20  &  4.98 & 0.964 & 0.556 &  220.1 & 117.4 &	1.00 & 0.723 \\
 ACFM  &  37.77  & 21.59  &	4.17  & 2.34  &	22.79  & 12.00  &	\underline{3.93}  & 1.67 & 0.920 & 0.559&  98.1 & 51.9 &	0.833 & 0.566  \\

 STID & \underline{27.36} & \underline{14.01} & 	3.85 & 1.63 & 	18.77 & 8.24 & 	4.06 & 1.54 & 0.880 &0.495 &	 47.4 & 23.3 & \underline{0.742} & \underline{0.469}  \\
 STNorm &  29.37  & 15.71  & 	4.44  & 2.09  & 	19.77  & 8.19  & 	4.45  & 1.66 & 0.961 & 0.532 & 54.3  & 47.9  & 0.871 & 0.579  \\
 STGSP & 45.04 & 28.28 & 7.93 & 4.56 & 39.99& 21.40 & 5.00 & 1.69 & 0.882 & 0.490 &  94.6 &  47.8&	1.02 & 0.749  \\
 MC-STL & 29.14 & 15.83 & 4.75 &2.39 & 21.22 & 10.26 & 4.08 & 2.05 & 1.19 & 0.833 & 54.2 & 28.1 &	1.00 & 0.720\\
 PromptST  &   27.44 &  14.54 &  \underline{3.52} & \underline{1.54} &  \underline{15.74} & \underline{7.20} &  4.36 & \underline{1.57} &   0.953 & 0.490 &  47.5 &   22.8 &   0.811  &  0.523  \\
\cmidrule(lr){1-1} \cmidrule(lr){2-3} \cmidrule(lr){4-5} \cmidrule(lr){6-7} \cmidrule(lr){8-9} \cmidrule(lr){10-11} \cmidrule(lr){12-13}   \cmidrule(lr){14-15}
MAU & 38.14 &20.13  &	4.94  &2.35  &	39.09  &18.73  &	5.22  &2.06 & 1.28 & 0.697 &  48.8 & 22.1 &	1.37 & 0.991 \\
PredRNN &  27.50 & 14.29 &	5.13 & 2.36 &	24.15 & 10.44 &	5.00 & 1.74 & \underline{0.852} & \underline{0.463} &  54.9 & 25.2 &	0.748 & 0.469 \\
MIM  & 28.62  & 14.77  &	5.66 & 2.27 &	21.38  & 9.37 &	4.40 & 1.62 & 1.17 & 0.650 &  51.4 & \underline{22.7} &	0.760 & 0.505 \\
SimVP &  32.66  & 17.67  & 	3.91  & 1.96  & 	16.48  & 8.23  & 	4.11  & 1.67 & 0.969 & 0.556 &  \underline{46.8} & 22.9  & 0.814 & 0.569  \\
TAU  &  33.90  & 19.37  & 	4.09  & 2.11  & 	17.94  & 8.91  & 	4.30  & 1.83 & 0.993 & 0.566 &  51.6 & 28.1 &	0.820 & 0.557 \\
\cmidrule(lr){1-1} \cmidrule(lr){2-3} \cmidrule(lr){4-5} \cmidrule(lr){6-7} \cmidrule(lr){8-9} \cmidrule(lr){10-11} \cmidrule(lr){12-13}   \cmidrule(lr){14-15}
PatchTST &  42.74 & 22.23 & 10.25 & 3.62 & 43.40  & 15.74 &  5.27 & 1.65 & 1.25 & 0.616 & 106.4 & 51.3 &	1.10 &  0.663 \\
iTransformer  & 36.97  &19.14  &9.40  & 3.40   &	37.01  & 13.93	& 7.74  &2.53  & 1.11 & 0.570 &  86.3 & 42.6 &	1.04 & 0.655 \\
\cmidrule(lr){1-1} \cmidrule(lr){2-3} \cmidrule(lr){4-5} \cmidrule(lr){6-7} \cmidrule(lr){8-9} \cmidrule(lr){10-11} \cmidrule(lr){12-13}   \cmidrule(lr){14-15}
PatchTST(one-for-all) &  43.66 & 23.16 & 13.51 & 5.00 & 56.80 & 20.56 & 9.97 & 3.05 & 1.30 & 0.645 & 127.0 & 59.26 & 1.13 & 0.679 \\
\cmidrule(lr){1-1} \cmidrule(lr){2-3} \cmidrule(lr){4-5} \cmidrule(lr){6-7} \cmidrule(lr){8-9} \cmidrule(lr){10-11} \cmidrule(lr){12-13}   \cmidrule(lr){14-15}
\textbf{UniST(one-for-all)}  & \textbf{26.84}  & \textbf{13.95} & \textbf{3.00} & \textbf{1.38} & 	\textbf{14.29} & \textbf{6.50} & 	\textbf{3.50} & \textbf{ 1.27} & \textbf{0.843} & \textbf{0.430} &  \textbf{44.97} & \textbf{19.67} & \textbf{0.665} &  \textbf{0.405} \\ 
\bottomrule
\end{tabular}}
\end{threeparttable}
\end{table*}

Subsequently, useful prompts are generated based on these optimized memories. This involves using the representations of spatio-temporal properties as queries to extract valuable memory knowledge, \textit{i.e.,} pertinent embeddings from the memory pool. Figure~\ref{fig:prompt} illustrates the process, and it is formulated as follows:

\begin{equation}
    \begin{aligned}
        & \alpha_{sc} =  [k_{s,0};k_{s,1}; ..., k_{s,N-1}]  E_{sc} ^T, \ P_{sc} = \sum_i \alpha_{sc,i} m_{s,i}, \\
        & \alpha_{sh} = [k_{s,0};k_{s,1}; ..., k_{s,N-1}]  E_{sh} ^T, \ P_{sh} = \sum_i \alpha_{sh,i} m_{s,i},  \\
        & \alpha_{tc} = [k_{t,0};k_{t,1}; ..., k_{t,N-1}] E_{tc}^T, \ P_{tc} = \sum_i \alpha_{tc,i} m_{t,i},  \\
        & \alpha_{tp} = [k_{t,0};k_{t,1}; ..., k_{t,N-1}] E_{tp}^T, \ P_{tp} = \sum_i \alpha_{tp,i} m_{t,i},  \\ \nonumber
    \end{aligned}
\end{equation}

\noindent where $E_{sc}, E_{sh}, E_{tc}, E_{tp}$ represent four representations related to four types of spatio-temporal domain knowledge, and $P_{sc}, P_{sh}, P_{tc}, P_{tp}$ are the extracted prompts. 
This allows the model to adaptively select the most useful information for prediction.
These prompts are then integrated into the input space of the Transformer architecture, which are displayed in the upper part of  Figure~\ref{fig:prompt}.


\section{Performance Evaluations}

\subsection{Experimental Setup}\label{exp:exp_set}

To evaluate the performance of UniST, we conducted extensive experiments on more than 20 spatio-temporal datasets.

\textbf{Datasets.} The datasets we used cover multiple cities, spanning various domains such as crowd flow, dynamic population, traffic speed, cellular network usage, taxi trips, and bike demand.   Appendix Table~\ref{tbl:dataset_info}  and Table~\ref{tbl:dataset_st} provide a summary of the datasets we used. These spatio-temporal datasets originate from distinct domains and cities, and have variations in the number of variables, sampling frequency, spatial scale, temporal duration, and data size.

\textbf{Baselines.} We compare UniST with a broad collection of state-of-the-art models for spatio-temporal prediction, which can be categorized into five groups:

\begin{itemize}[leftmargin=*]
    \item \textbf{Heuristic approaches.} History average (HA) and ARIMA.
    \item \textbf{Deep urban prediction approaches.} We consider state-of-the-art  urban ST prediction models, including STResNet~\cite{zhang2017deep}, ACFM~\cite{liu2018attentive}, MC-STL~\cite{zhang2023mask}, STGSP~\cite{zhao2022st}, STNorm~\cite{deng2021st},  STID~\cite{shao2022spatial}, and  PromptST~\cite{zhang2023promptst}. 
    \item \textbf{Video prediction approaches.} We compare with competitive video prediction models from the popular benchmark, including PredRNN~\cite{wang2017predrnn}, MAU~\cite{chang2021mau}, MIM~\cite{wang2019memory}, SimVP~\cite{gao2022simvp}, and TAU~\cite{tan2023temporal}. 
    \item \textbf{Multivariate time series forecasting approaches.} We consider state-of-the-art multivariate time series forecasting models, including PatchTST~\cite{nie2022time} and iTransformer~\cite{liu2023itransformer}. For a fair comparison, we also train PatchTST for all datasets, denoted as PatchTST(one-for-all).
    \item \textbf{Meta learning approaches.} To evaluate the generalization capability, we consider meta-learning approaches including MAML~\cite{finn2017model} and MetaST~\cite{yao2019learning}.
\end{itemize}

\textbf{Metrics.} We employed commonly used regression metrics, including Mean Absolute Error (MAE) and Root Mean Squared Error (RMSE). 
For more detailed information of the datasets, baselines, and metrics, please refer to Appendix~\ref{sup:dataset}, Appendix~\ref{sup:baseline}, and Appendix~\ref{sup:imp_detail}.

\subsection{Short-Term Prediction}

\subsubsection*{\textbf{Setups.}} 
Following previous practices~\cite{jin2023time,nie2022time}, both the input step and prediction horizon are set as 6, i.e., $6\rightarrow 6$.  For baselines, we train a dedicated model for each dataset, while UniST is evaluated  across all datasets.

\subsubsection*{\textbf{Results.}} Table~\ref{tbl:short_term} presents the short-term prediction results, with a selection of datasets due to space constraints. The complete results can be found in Table~\ref{tbl:short1} and Table~\ref{tbl:short2} in Appendix~\ref{sup:add_result}. As we can observe from Table~\ref{tbl:short_term}, UniST consistently outperforms all baselines across all datasets. Compared with the best baseline of each dataset, it showcases a notable average improvement. 
Notably, time series approaches such as PatchTST and iTransformer exhibit inferior performance compared to spatio-temporal methods.
This underscores the importance of incorporating spatial dependency as prior knowledge for spatio-temporal prediction tasks.
Another observation is that PatchTST(one-for-all) performs worse than PatchTST dedicated for each dataset, suggesting that the model struggles to directly adept to these distinct data distributions. 
Moreover, baseline approaches exhibit inconsistent performance across diverse datasets, indicating their instability across scenarios. The consistent superior performance of UniST across all scenarios underscores the significant potential and benefits of a one-for-all model. Moreover, it demonstrates UniST's capability to orchestrate diverse data, where different datasets can benefit each other.

\subsection{Long-Term Prediction}

\begin{table}[t!]
\caption{Performance comparison of long-term prediction on three datasets in terms of MAE and RMSE.  We use the average prediction errors over all prediction steps. Bold denotes the best results and \underline{underline} denotes the second-best results.
}\label{tbl:long_term}
\begin{threeparttable}
\resizebox{\columnwidth}{!}{
\begin{tabular}{ccccccc}
\toprule
& \multicolumn{2}{c}{\textbf{TaxiNYC}} & \multicolumn{2}{c}{\textbf{Crowd}} & \multicolumn{2}{c}{\textbf{BikeNYC}}  \\ 
 \cmidrule(lr){2-3} \cmidrule(lr){4-5}  \cmidrule(lr){6-7} 
 \textbf{Model} & \textbf{RMSE}  & \textbf{MAE}    & \textbf{RMSE}    & \textbf{MAE}   & \textbf{RMSE}    & \textbf{MAE}     \\ 
\cmidrule(lr){1-1} \cmidrule(lr){2-3} \cmidrule(lr){4-5} \cmidrule(lr){6-7} 

  HA &  61.03 & 21.33  & 19.57 &  8.49 & 11.00 & 3.66\\
 ARIMA  & 68.0 & 28.66 & 21.34 & 8.93  & 11.59 & 3.98 \\
\cmidrule(lr){1-1} \cmidrule(lr){2-3} \cmidrule(lr){4-5}  \cmidrule(lr){6-7} 

 STResNet & 29.54 & 14.46 & 8.75 & 5.58 & 7.15 & 3.87\\
 ACFM  & 32.91 & 13.72 &6.16 & 3.35 & 4.56 & 1.86 \\ 
 STID & 24.74 & 11.01 & \underline{4.91} & \underline{2.63} & 4.78 & 2.24 \\
STNorm & 31.81 & 11.99 & 9.62 & 4.30 &  6.45 & 2.18\\
 STGSP &  28.65 & 10.38 & 17.03 & 8.21 & 4.71 & \underline{1.54} \\
 MC-STL &  29.29 & 17.36 & 9.01 & 6.32  & 4.97 & 2.61\\
\cmidrule(lr){1-1} \cmidrule(lr){2-3} \cmidrule(lr){4-5}  \cmidrule(lr){6-7} 
MAU &  26.28 & 9.07 & 20.13 & 8.49 & 6.18 & 2.13 \\
PredRNN & 21.17 & \underline{7.31} & 19.70 & 10.66 & 5.86 & 1.97\\
MIM  & 63.36 & 29.83 & 15.70 & 8.81 & 7.58 & 2.81\\
SimVP &  \underline{20.18} & 9.78 & 5.50 & 3.13 & 4.10 & 1.71  \\
TAU  &  24.97 & 10.93 & 5.31 & 2.81 & \underline{3.89} &  1.73  \\
\cmidrule(lr){1-1} \cmidrule(lr){2-3} \cmidrule(lr){4-5}  \cmidrule(lr){6-7} 
PatchTST & 30.64 & 17.49 & 5.25 & 2.83 & 5.27 & 1.65   \\

iTransformer  & 33.81 & 11.48 & 6.94 & 2.63  & 6.00 & 2.02  \\
\cmidrule(lr){1-1} \cmidrule(lr){2-3} \cmidrule(lr){4-5}  \cmidrule(lr){6-7} 
PatchTST(one-for-all) &  34.50 & 10.63 & 6.39 & 2.92 & 6.02 & 1.83   \\
\cmidrule(lr){1-1} \cmidrule(lr){2-3} \cmidrule(lr){4-5}  \cmidrule(lr){6-7} 
\textbf{UniST (one-for-all)}  & \textbf{19.83} & \textbf{6.71} & \textbf{4.25} & \textbf{2.26} & \textbf{3.56} & \textbf{1.31} \\ 
\bottomrule
\end{tabular}}
\end{threeparttable}
\end{table}

\subsubsection*{\textbf{Setups.}} 
Here we extend the input step and prediction horizon to 64 following~\cite{jin2023time,nie2022time}.
This configuration accommodates prolonged temporal dependencies, allowing us to gauge the model's proficiency in capturing extended patterns over time.
Similar to the short-term prediction, UniST is directly evaluated across all datasets, while specific models are individually trained for each baseline on respective datasets.

\subsubsection*{\textbf{Results.}} Table~\ref{tbl:long_term} shows the long-term prediction results. Even with a more extended prediction horizon, UniST still consistently outperforms all baseline approaches across all datasets.  Compared with the best baseline of each dataset, it yields an average improvement of 10.1\%.  
This highlights UniST's capability to comprehend temporal patterns effectively and its robustness in generalizing across extended durations. 
Table~\ref{tbl:long1} in Appendix~\ref{sup:add_result} illustrates the complete results.

\begin{figure}[t]
    \centering
    \includegraphics[width=0.99\linewidth]{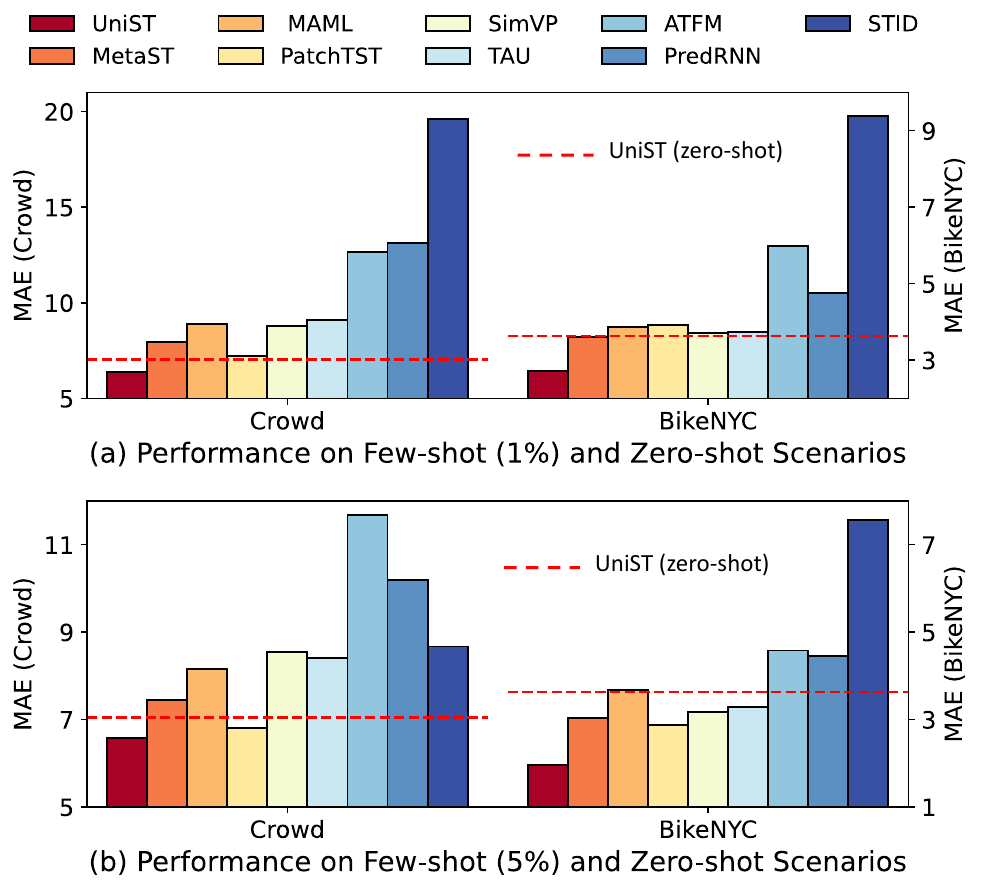}
    \caption{ (a) Few-shot performance of UniST and baselines on Crowd and BikeNYC datasets using only 1\% of the training data. (b) Few-shot performance of UniST and baselines using only 5\% of the training data. The Dashed red lines denote the zero-shot performance of UniST.}
    \label{fig:few_zero}
\end{figure}

\subsection{Few-Shot Prediction}

\subsubsection*{\textbf{Setups.}} The hallmark of large foundation models lies in their exceptional generalization ability. 
The few-shot and zero-shot evaluations are commonly employed to characterize the ultimate tasks for universal time series forecasting~\cite{yuan2024spatiotemporal,zhou2023one}.  Likewise, the few-shot and zero-shot prediction capability is crucial for a universal spatio-temporal model. In this section, we assess the few-shot learning performance of UniST. Each dataset is partitioned into three segments: training data, validation data, and test data.
In few-shot learning scenarios, when confronted with an unseen dataset during the training process, we utilized a restricted amount of training data, specifically, \textit{1\%, 5\%, 10\%} of the training data. We choose some baselines with relatively good performance for the few-shot setting evaluation, We also compare with meta-learning baselines, \textit{i.e.,} MAML and MetaST, and pretraining and finetuning-based time series method, \textit{i.e.,} PatchTST.

\subsubsection*{\textbf{Results.}} Appendix Table~\ref{tbl:few_shot_crowd} to Table~\ref{tbl:few_shot_taxibj} illustrate the overall few-shot results. Due to the space limit, Figure~\ref{fig:few_zero} only illustrates the 1\% few-shot learning results on two datasets. In these cases, UniST still outperforms all baselines, it achieves a larger relative improvement over baselines compared to long-term and short-term predictions.  The transferability can be attributed to successful knowledge transfer in our spatio-temporal prompt.

\subsection{Zero-Shot Prediction}

\subsubsection*{\textbf{Setups.}}
 Zero-shot inference serves as the ultimate task for assessing a model's adaptation ability. In this context, after training on a diverse collection of datasets, we evaluate UniST on an entirely novel dataset—\textit{i.e., without any prior training data from it}. The test data used in this scenario aligns with that of normal prediction and few-shot prediction.

\subsubsection*{\textbf{Results.}} Figure~\ref{fig:few_zero} also compares the performance of UniST (zero-shot) and baselines (few-shot). As observed, UniST achieves remarkable zero-shot performance, even surpassing many baselines trained with training data that are highlighted by red dashed lines. We attribute these surprising results to the powerful spatio-temporal transfer capability. It suggests that for a completely new scenario, even when the displayed overall patterns are dissimilar to the data encountered during the training process, UniST can extract fine-grained similar patterns from our defined spatial and temporal properties. The few-shot and zero-shot results demonstrate the powerful generalization capability of UniST.

\section{Study and Analysis on UniST}

\begin{figure}[t]
    \centering
    \includegraphics[width=0.95\linewidth]{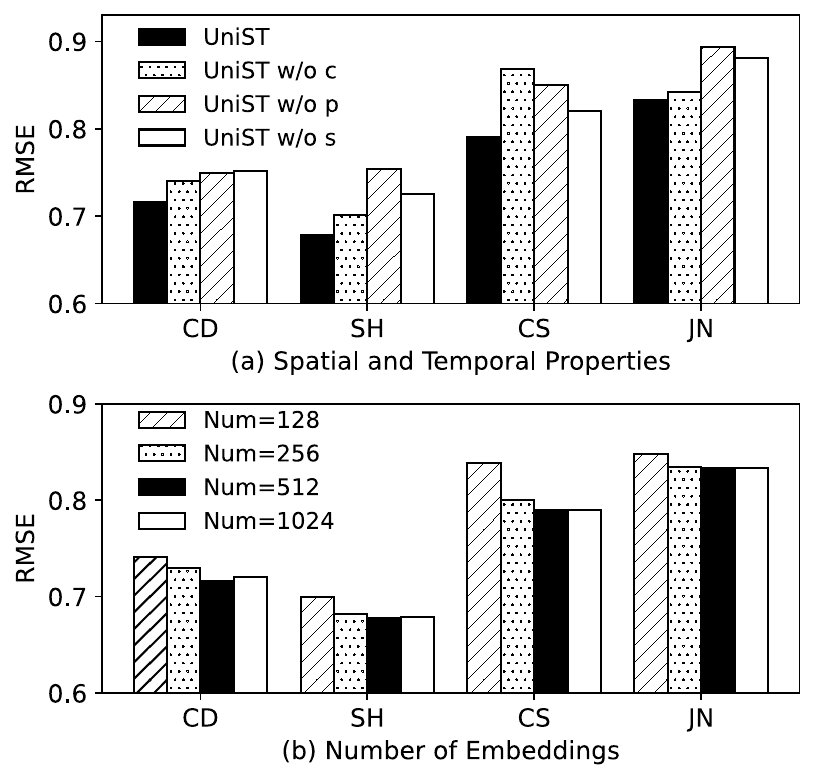}
    \caption{Ablation studies on four traffic speed datasets: Chengdu (CD), Shanghai (SH), Changsha (CS), and Jinan (JN). (a) illustrates the results of removing a prompt guided by one type of spatio-temporal knowledge. (b) presents the results of varying the number of learnable embeddings in the temporal and spatial memory pools.}
    \label{fig:ablation}
\end{figure}

\subsection{Ablation Study}

The prompts play an essential role in our UniST model. Here we investigate whether the designed spatial and temporal properties contribute to the overall performance. We use `s' to denote spatial closeness and hierarchy, `p' for temporal periodicity, and `c' for temporal closeness.
we compare the overall design that incorporates all three properties with three degraded versions that individually remove `s', `p', or `c'. 
Figure~\ref{fig:ablation}(a) shows the results on four traffic speed datasets. As we can observe, removing any property results in a performance decrease. 
The contributions of each spatial and temporal property vary across different datasets, highlighting the necessity of each property for the spatio-temporal design.

Additionally, we explore how the number of embeddings in the memory pools affects the final performance. 
As seen in Figure~\ref{fig:ablation}(b), increasing the number from 128 to 512 improves performance across the four datasets. When further increasing the number to 1024, the performance remains similar to 512, suggesting that 512 is the optimal choice.

\begin{figure}[t]
    \centering
    \includegraphics[width=0.99\linewidth]{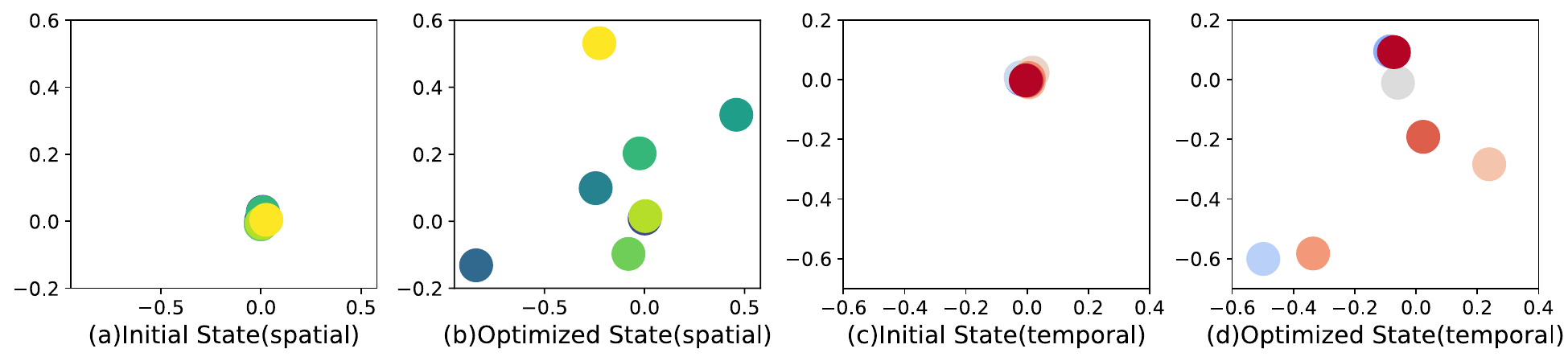}
    \caption{Embeddings visualization of spatial and temporal memory pools at the initial and final optimized states.  The embeddings exhibit obvious divergence. }
    \label{fig:case_tsne}
\end{figure}

\begin{figure}[t]
    \centering
    \includegraphics[width=0.99\linewidth]{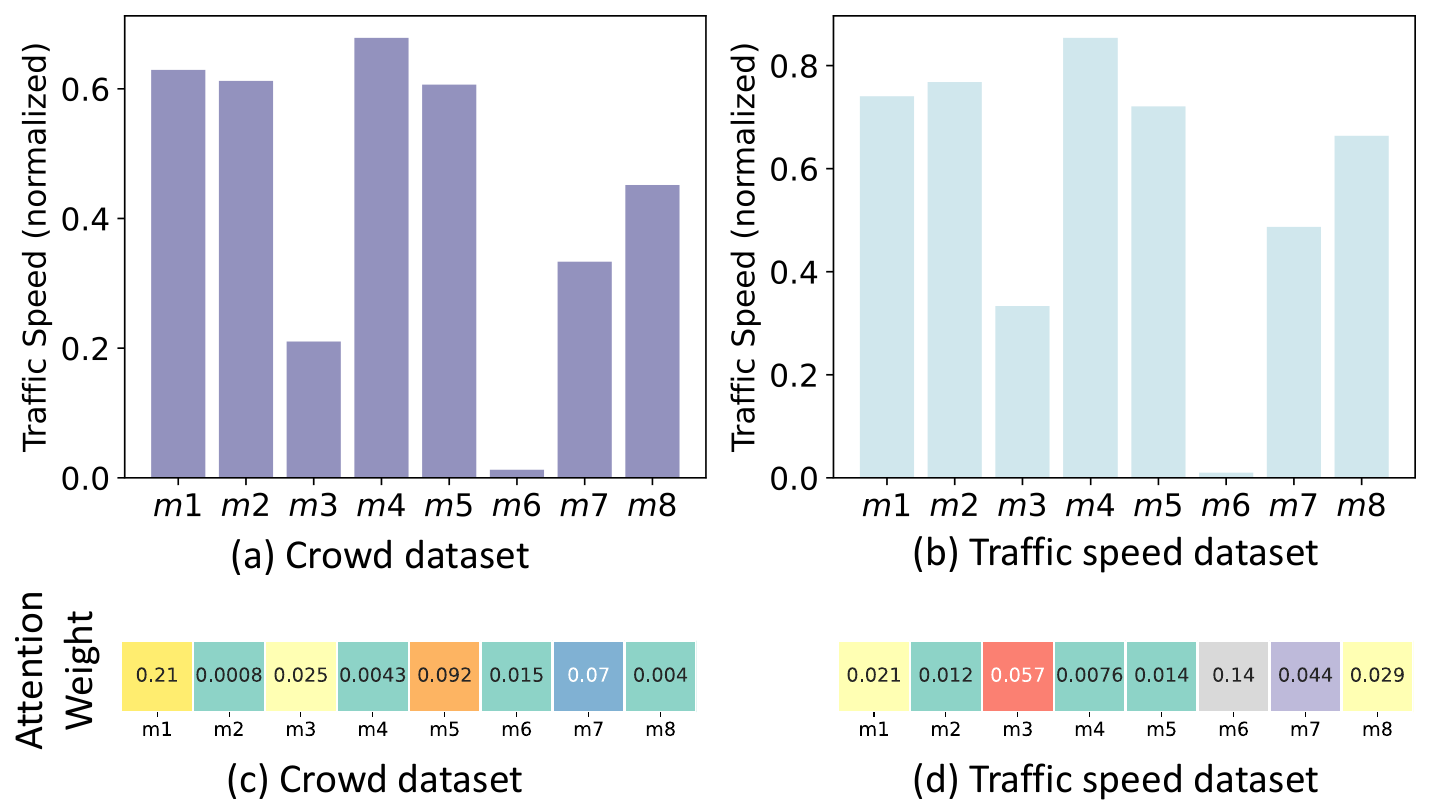}
    \caption{(a) and (b): Comparison of the mean value of inputs in each memory embedding, where the inputs assign the highest attention weight to the memory embedding. (c) and (d): Comparison of the attention weight on each memory embedding for two distinct datasets.}
    \label{fig:case_uni}
\end{figure}

\subsection{Prompt Learner}

In this section, we conduct in-depth analyses of the prompt learner.
To provide a clearer understanding, we leverage t-Distributed Stochastic Neighbor Embedding (t-SNE) to visualize the embeddings of both the spatial and temporal memory pools.
Specifically, we plot the initial state and the optimized state in Figure~\ref{fig:case_tsne}. Notably, from the start state to the final optimized state, the embeddings gradually become diverged in different directions. 
This suggests that, throughout the optimization process, the memory pools progressively store and encapsulate personalized information.

Next, we delve into the memorized patterns of each embedding within the temporal memory pool.  Specifically, we first select the inputs based on the attention weights. For each embedding, we aggregate the corresponding input spatio-temporal data with the highest attention weight. Then, we calculate the mean value of the extracted spatio-temporal data. Figure~\ref{fig:case_uni}(a) and Figure~\ref{fig:case_uni}(b) illustrate the results for two datasets (Crowd and TrafficSH). As we can see, the memorized patterns revealed in the prompt tool exhibit remarkable consistency across different urban scenarios.
This not only affirms that each embedding is meticulously optimized to memorize unique spatio-temporal patterns, but also underscores the robustness of the spatial and temporal memory pools across different scenarios.

Moreover,  we examine the extracted spatio-temporal prompts for two distinct domains.  Specifically, we calculate the mean attention weight for each embedding in the context of each dataset. Figure~\ref{fig:case_uni}(c) and Figure~\ref{fig:case_uni}(d)  illustrate the comparison results. As we can observe,  the depicted attention weight distributions for the two datasets manifest striking dissimilarities.
The observed distinctiveness in attention weight distributions implies a dynamic and responsive nature in the model's ability to tailor its focus based on the characteristics of the input data.
The ability to dynamically adjust the attention weights reinforces UniST's versatility and universality for diverse datasets.

\begin{figure}[t]
    \centering
    \includegraphics[width=0.99\linewidth]{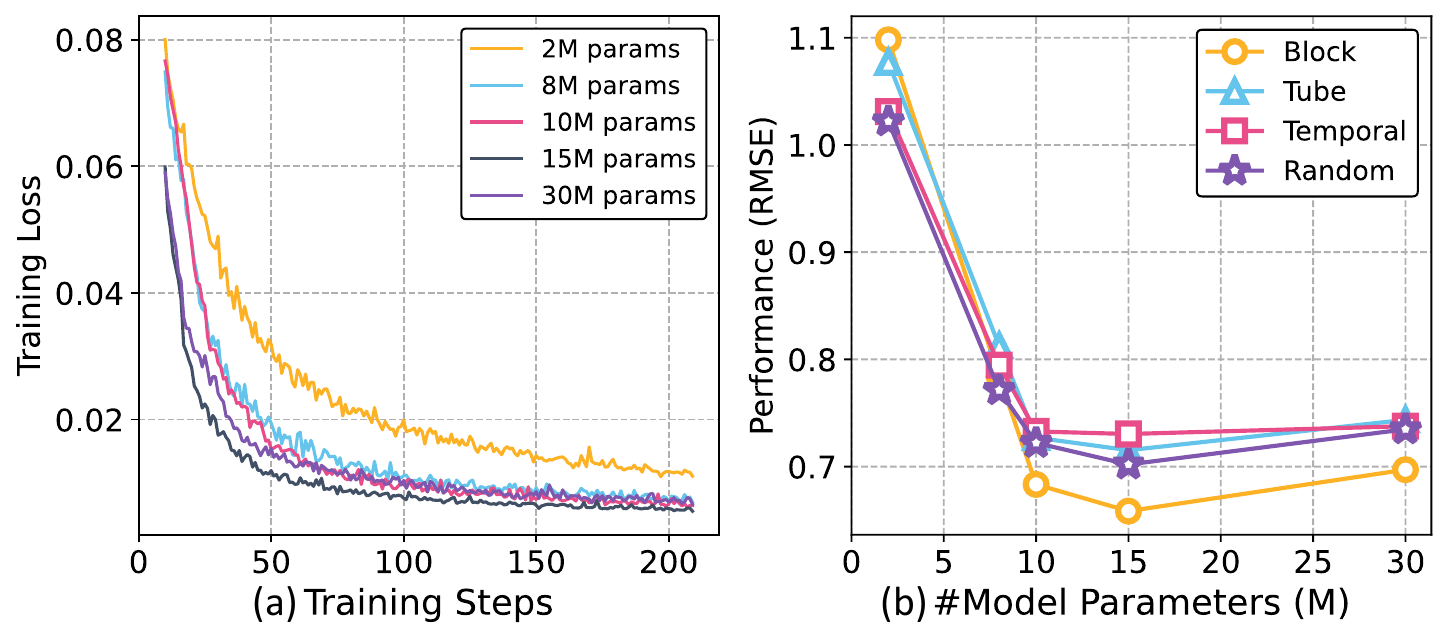}
    \caption{(a) Training loss across five models with varying parameter sizes. (b) Performance evaluation of masked patch reconstruction by increasing parameter sizes. }
    \label{fig:scaling}
\end{figure}

\subsection{Scalability}
Scalability is a crucial characteristic for universal models, therefore,  we explore the scaling behavior of our UniST model. Our investigation specifically concentrates on observing changes in training loss and prediction performance as we vary the model parameter size.
Figure~\ref{fig:scaling} depicts the training loss and testing RMSE of UniST with varying parameter sizes. Regarding training loss (left
 figure), several key observations emerge: (i) across different parameter sizes, the training loss consistently decreases and gradually converges with increasing training steps; (ii) increasing the parameter size accelerates the convergence of the training loss; (iii) there exist diminishing marginal returns, suggesting that reducing the training loss becomes progressively harder as parameter size increases.  The right figure illustrates the reconstruction RMSE on the testing set, showing similar trends to the training loss. 

 These observations indicate that UniST has shown scalability behaviors,wherein larger models generally exhibit improved performance. However, unlike large language and vision models~\cite{kaplan2020scaling, bai2023sequential}, the scalability in spatio-temporal prediction shows diminishing marginal returns.  This may stem from the relative lack of diversity in spatio-temporal data compared to language or visual datasets.

\section{Conclusion}

In this work, we address an important problem of building a universal model UniST for urban spatio-temporal prediction. By leveraging the diversity of spatio-temporal data from multiple sources, and discerning and aligning underlying shared spatio-temporal patterns across multiple scenarios, UniST demonstrates a powerful capability to predict across all scenarios, particularly in few-shot and zero-shot settings.   A promising direction for future work entails the integration of various spatio-temporal data formats, such as grid, sequence, and graph data. 
Our study inspires future research in spatio-temporal modeling towards the universal direction.

\section*{Acknowledgments}
This work was supported in part by the National Key Research and Development Program of China under grant 2020YFA0711403  and the National Natural Science Foundation of China under 62171260 and 62272260.

\bibliographystyle{ACM-Reference-Format}
\bibliography{8.reference}


\begin{thebibliography}{76}


\ifx \showCODEN    \undefined \def \showCODEN     #1{\unskip}     \fi
\ifx \showDOI      \undefined \def \showDOI       #1{#1}\fi
\ifx \showISBNx    \undefined \def \showISBNx     #1{\unskip}     \fi
\ifx \showISBNxiii \undefined \def \showISBNxiii  #1{\unskip}     \fi
\ifx \showISSN     \undefined \def \showISSN      #1{\unskip}     \fi
\ifx \showLCCN     \undefined \def \showLCCN      #1{\unskip}     \fi
\ifx \shownote     \undefined \def \shownote      #1{#1}          \fi
\ifx \showarticletitle \undefined \def \showarticletitle #1{#1}   \fi
\ifx \showURL      \undefined \def \showURL       {\relax}        \fi
\providecommand\bibfield[2]{#2}
\providecommand\bibinfo[2]{#2}
\providecommand\natexlab[1]{#1}
\providecommand\showeprint[2][]{arXiv:#2}

\bibitem[Bai et~al\mbox{.}(2020)]%
        {bai2020adaptive}
\bibfield{author}{\bibinfo{person}{Lei Bai}, \bibinfo{person}{Lina Yao}, \bibinfo{person}{Can Li}, \bibinfo{person}{Xianzhi Wang}, {and} \bibinfo{person}{Can Wang}.} \bibinfo{year}{2020}\natexlab{}.
\newblock \showarticletitle{Adaptive graph convolutional recurrent network for traffic forecasting}.
\newblock \bibinfo{journal}{\emph{Advances in neural information processing systems}}  \bibinfo{volume}{33} (\bibinfo{year}{2020}), \bibinfo{pages}{17804--17815}.
\newblock


\bibitem[Bai et~al\mbox{.}(2023)]%
        {bai2023sequential}
\bibfield{author}{\bibinfo{person}{Yutong Bai}, \bibinfo{person}{Xinyang Geng}, \bibinfo{person}{Karttikeya Mangalam}, \bibinfo{person}{Amir Bar}, \bibinfo{person}{Alan Yuille}, \bibinfo{person}{Trevor Darrell}, \bibinfo{person}{Jitendra Malik}, {and} \bibinfo{person}{Alexei~A Efros}.} \bibinfo{year}{2023}\natexlab{}.
\newblock \showarticletitle{Sequential modeling enables scalable learning for large vision models}.
\newblock \bibinfo{journal}{\emph{arXiv preprint arXiv:2312.00785}} (\bibinfo{year}{2023}).
\newblock


\bibitem[Brown et~al\mbox{.}(2020)]%
        {brown2020language}
\bibfield{author}{\bibinfo{person}{Tom Brown}, \bibinfo{person}{Benjamin Mann}, \bibinfo{person}{Nick Ryder}, \bibinfo{person}{Melanie Subbiah}, \bibinfo{person}{Jared~D Kaplan}, \bibinfo{person}{Prafulla Dhariwal}, \bibinfo{person}{Arvind Neelakantan}, \bibinfo{person}{Pranav Shyam}, \bibinfo{person}{Girish Sastry}, \bibinfo{person}{Amanda Askell}, {et~al\mbox{.}}} \bibinfo{year}{2020}\natexlab{}.
\newblock \showarticletitle{Language models are few-shot learners}.
\newblock \bibinfo{journal}{\emph{Advances in neural information processing systems}}  \bibinfo{volume}{33} (\bibinfo{year}{2020}), \bibinfo{pages}{1877--1901}.
\newblock


\bibitem[Cao et~al\mbox{.}(2023)]%
        {cao2023tempo}
\bibfield{author}{\bibinfo{person}{Defu Cao}, \bibinfo{person}{Furong Jia}, \bibinfo{person}{Sercan~O Arik}, \bibinfo{person}{Tomas Pfister}, \bibinfo{person}{Yixiang Zheng}, \bibinfo{person}{Wen Ye}, {and} \bibinfo{person}{Yan Liu}.} \bibinfo{year}{2023}\natexlab{}.
\newblock \showarticletitle{Tempo: Prompt-based generative pre-trained transformer for time series forecasting}.
\newblock \bibinfo{journal}{\emph{arXiv preprint arXiv:2310.04948}} (\bibinfo{year}{2023}).
\newblock


\bibitem[Chang et~al\mbox{.}(2021)]%
        {chang2021mau}
\bibfield{author}{\bibinfo{person}{Zheng Chang}, \bibinfo{person}{Xinfeng Zhang}, \bibinfo{person}{Shanshe Wang}, \bibinfo{person}{Siwei Ma}, \bibinfo{person}{Yan Ye}, \bibinfo{person}{Xiang Xinguang}, {and} \bibinfo{person}{Wen Gao}.} \bibinfo{year}{2021}\natexlab{}.
\newblock \showarticletitle{Mau: A motion-aware unit for video prediction and beyond}.
\newblock \bibinfo{journal}{\emph{Advances in Neural Information Processing Systems}}  \bibinfo{volume}{34} (\bibinfo{year}{2021}), \bibinfo{pages}{26950--26962}.
\newblock


\bibitem[Chen et~al\mbox{.}(2022)]%
        {chen2022bidirectional}
\bibfield{author}{\bibinfo{person}{Changlu Chen}, \bibinfo{person}{Yanbin Liu}, \bibinfo{person}{Ling Chen}, {and} \bibinfo{person}{Chengqi Zhang}.} \bibinfo{year}{2022}\natexlab{}.
\newblock \showarticletitle{Bidirectional spatial-temporal adaptive transformer for Urban traffic flow forecasting}.
\newblock \bibinfo{journal}{\emph{IEEE Transactions on Neural Networks and Learning Systems}} (\bibinfo{year}{2022}).
\newblock


\bibitem[Chen et~al\mbox{.}(2021)]%
        {chen2021s2tnet}
\bibfield{author}{\bibinfo{person}{Weihuang Chen}, \bibinfo{person}{Fangfang Wang}, {and} \bibinfo{person}{Hongbin Sun}.} \bibinfo{year}{2021}\natexlab{}.
\newblock \showarticletitle{S2tnet: Spatio-temporal transformer networks for trajectory prediction in autonomous driving}. In \bibinfo{booktitle}{\emph{Asian Conference on Machine Learning}}. PMLR, \bibinfo{pages}{454--469}.
\newblock


\bibitem[Deng et~al\mbox{.}(2021)]%
        {deng2021st}
\bibfield{author}{\bibinfo{person}{Jinliang Deng}, \bibinfo{person}{Xiusi Chen}, \bibinfo{person}{Renhe Jiang}, \bibinfo{person}{Xuan Song}, {and} \bibinfo{person}{Ivor~W Tsang}.} \bibinfo{year}{2021}\natexlab{}.
\newblock \showarticletitle{St-norm: Spatial and temporal normalization for multi-variate time series forecasting}. In \bibinfo{booktitle}{\emph{Proceedings of the 27th ACM SIGKDD conference on knowledge discovery \& data mining}}. \bibinfo{pages}{269--278}.
\newblock


\bibitem[Devlin et~al\mbox{.}(2018)]%
        {devlin2018bert}
\bibfield{author}{\bibinfo{person}{Jacob Devlin}, \bibinfo{person}{Ming-Wei Chang}, \bibinfo{person}{Kenton Lee}, {and} \bibinfo{person}{Kristina Toutanova}.} \bibinfo{year}{2018}\natexlab{}.
\newblock \showarticletitle{Bert: Pre-training of deep bidirectional transformers for language understanding}.
\newblock \bibinfo{journal}{\emph{arXiv preprint arXiv:1810.04805}} (\bibinfo{year}{2018}).
\newblock


\bibitem[Feng et~al\mbox{.}(2024a)]%
        {feng2014citygpt}
\bibfield{author}{\bibinfo{person}{Jie Feng}, \bibinfo{person}{Yuwei Du}, \bibinfo{person}{Tianhui Liu}, \bibinfo{person}{Siqi Guo}, \bibinfo{person}{Yuming Lin}, {and} \bibinfo{person}{Yong Li}.} \bibinfo{year}{2024}\natexlab{a}.
\newblock \showarticletitle{CityGPT: Empowering Urban Spatial Cognition of Large Language Models}.
\newblock \bibinfo{journal}{\emph{arXiv preprint arXiv:2406.13948}} (\bibinfo{year}{2024}).
\newblock


\bibitem[Feng et~al\mbox{.}(2024b)]%
        {feng2014citybench}
\bibfield{author}{\bibinfo{person}{Jie Feng}, \bibinfo{person}{Jun Zhang}, \bibinfo{person}{Junbo Yan}, \bibinfo{person}{Xin Zhang}, \bibinfo{person}{Tianjian Ouyang}, \bibinfo{person}{Tianhui Liu}, \bibinfo{person}{Yuwei Du}, \bibinfo{person}{Siqi Guo}, {and} \bibinfo{person}{Yong Li}.} \bibinfo{year}{2024}\natexlab{b}.
\newblock \showarticletitle{CityBench: Evaluating the Capabilities of Large Language Model as World Model}.
\newblock \bibinfo{journal}{\emph{arXiv preprint arXiv:2406.13945}} (\bibinfo{year}{2024}).
\newblock


\bibitem[Finn et~al\mbox{.}(2017)]%
        {finn2017model}
\bibfield{author}{\bibinfo{person}{Chelsea Finn}, \bibinfo{person}{Pieter Abbeel}, {and} \bibinfo{person}{Sergey Levine}.} \bibinfo{year}{2017}\natexlab{}.
\newblock \showarticletitle{Model-agnostic meta-learning for fast adaptation of deep networks}. In \bibinfo{booktitle}{\emph{International conference on machine learning}}. PMLR, \bibinfo{pages}{1126--1135}.
\newblock


\bibitem[Gao et~al\mbox{.}(2022)]%
        {gao2022simvp}
\bibfield{author}{\bibinfo{person}{Zhangyang Gao}, \bibinfo{person}{Cheng Tan}, \bibinfo{person}{Lirong Wu}, {and} \bibinfo{person}{Stan~Z Li}.} \bibinfo{year}{2022}\natexlab{}.
\newblock \showarticletitle{Simvp: Simpler yet better video prediction}. In \bibinfo{booktitle}{\emph{Proceedings of the IEEE/CVF Conference on Computer Vision and Pattern Recognition}}. \bibinfo{pages}{3170--3180}.
\newblock


\bibitem[Garza and Mergenthaler-Canseco(2023)]%
        {garza2023timegpt}
\bibfield{author}{\bibinfo{person}{Azul Garza} {and} \bibinfo{person}{Max Mergenthaler-Canseco}.} \bibinfo{year}{2023}\natexlab{}.
\newblock \showarticletitle{TimeGPT-1}.
\newblock \bibinfo{journal}{\emph{arXiv preprint arXiv:2310.03589}} (\bibinfo{year}{2023}).
\newblock


\bibitem[Geng et~al\mbox{.}(2019)]%
        {geng2019spatiotemporal}
\bibfield{author}{\bibinfo{person}{Xu Geng}, \bibinfo{person}{Yaguang Li}, \bibinfo{person}{Leye Wang}, \bibinfo{person}{Lingyu Zhang}, \bibinfo{person}{Qiang Yang}, \bibinfo{person}{Jieping Ye}, {and} \bibinfo{person}{Yan Liu}.} \bibinfo{year}{2019}\natexlab{}.
\newblock \showarticletitle{Spatiotemporal multi-graph convolution network for ride-hailing demand forecasting}. In \bibinfo{booktitle}{\emph{Proceedings of the AAAI conference on artificial intelligence}}, Vol.~\bibinfo{volume}{33}. \bibinfo{pages}{3656--3663}.
\newblock


\bibitem[Gong et~al\mbox{.}(2024)]%
        {gongKDDPITuning}
\bibfield{author}{\bibinfo{person}{Jiahui Gong}, \bibinfo{person}{Jingtao Ding}, \bibinfo{person}{Fanjin Meng}, \bibinfo{person}{Guilong Chen}, \bibinfo{person}{Hong Chen}, \bibinfo{person}{Shen Zhao}, \bibinfo{person}{Haisheng Lu}, {and} \bibinfo{person}{Yong Li}.} \bibinfo{year}{2024}\natexlab{}.
\newblock \showarticletitle{A Population-to-individual Tuning Framework for Adapting Pretrained LM to On-device User Intent Prediction}. In \bibinfo{booktitle}{\emph{Proceedings of the 30th ACM SIGKDD Conference on Knowledge Discovery and Data Mining}}. \bibinfo{publisher}{Association for Computing Machinery}, \bibinfo{address}{New York, NY, USA}.
\newblock
\urldef\tempurl%
\url{https://doi.org/10.1145/3637528.3671984}
\showDOI{\tempurl}


\bibitem[Gong et~al\mbox{.}(2023)]%
        {gong2023empowering}
\bibfield{author}{\bibinfo{person}{Jiahui Gong}, \bibinfo{person}{Yu Liu}, \bibinfo{person}{Tong Li}, \bibinfo{person}{Haoye Chai}, \bibinfo{person}{Xing Wang}, \bibinfo{person}{Junlan Feng}, \bibinfo{person}{Chao Deng}, \bibinfo{person}{Depeng Jin}, {and} \bibinfo{person}{Yong Li}.} \bibinfo{year}{2023}\natexlab{}.
\newblock \showarticletitle{Empowering spatial knowledge graph for mobile traffic prediction}. In \bibinfo{booktitle}{\emph{Proceedings of the 31st ACM International Conference on Advances in Geographic Information Systems}}. \bibinfo{pages}{1--11}.
\newblock


\bibitem[He et~al\mbox{.}(2022)]%
        {he2022masked}
\bibfield{author}{\bibinfo{person}{Kaiming He}, \bibinfo{person}{Xinlei Chen}, \bibinfo{person}{Saining Xie}, \bibinfo{person}{Yanghao Li}, \bibinfo{person}{Piotr Doll{\'a}r}, {and} \bibinfo{person}{Ross Girshick}.} \bibinfo{year}{2022}\natexlab{}.
\newblock \showarticletitle{Masked autoencoders are scalable vision learners}. In \bibinfo{booktitle}{\emph{Proceedings of the IEEE/CVF conference on computer vision and pattern recognition}}. \bibinfo{pages}{16000--16009}.
\newblock


\bibitem[Ji et~al\mbox{.}(2023)]%
        {ji2023spatio}
\bibfield{author}{\bibinfo{person}{Jiahao Ji}, \bibinfo{person}{Jingyuan Wang}, \bibinfo{person}{Chao Huang}, \bibinfo{person}{Junjie Wu}, \bibinfo{person}{Boren Xu}, \bibinfo{person}{Zhenhe Wu}, \bibinfo{person}{Zhang Junbo}, {and} \bibinfo{person}{Yu Zheng}.} \bibinfo{year}{2023}\natexlab{}.
\newblock \showarticletitle{Spatio-Temporal Self-Supervised Learning for Traffic Flow Prediction}.
\newblock \bibinfo{journal}{\emph{Proceedings of the AAAI Conference on Artificial Intelligence}} \bibinfo{volume}{37}, \bibinfo{number}{4} (\bibinfo{year}{2023}), \bibinfo{pages}{4356--4364}.
\newblock


\bibitem[Jia et~al\mbox{.}(2022)]%
        {jia2022visual}
\bibfield{author}{\bibinfo{person}{Menglin Jia}, \bibinfo{person}{Luming Tang}, \bibinfo{person}{Bor-Chun Chen}, \bibinfo{person}{Claire Cardie}, \bibinfo{person}{Serge Belongie}, \bibinfo{person}{Bharath Hariharan}, {and} \bibinfo{person}{Ser-Nam Lim}.} \bibinfo{year}{2022}\natexlab{}.
\newblock \showarticletitle{Visual prompt tuning}. In \bibinfo{booktitle}{\emph{European Conference on Computer Vision}}. Springer, \bibinfo{pages}{709--727}.
\newblock


\bibitem[Jiang et~al\mbox{.}(2023)]%
        {jiang2023pdformer}
\bibfield{author}{\bibinfo{person}{Jiawei Jiang}, \bibinfo{person}{Chengkai Han}, \bibinfo{person}{Wayne~Xin Zhao}, {and} \bibinfo{person}{Jingyuan Wang}.} \bibinfo{year}{2023}\natexlab{}.
\newblock \showarticletitle{PDFormer: Propagation Delay-aware Dynamic Long-range Transformer for Traffic Flow Prediction}.
\newblock \bibinfo{journal}{\emph{arXiv preprint arXiv:2301.07945}} (\bibinfo{year}{2023}).
\newblock


\bibitem[Jin et~al\mbox{.}(2021)]%
        {jin2021trafficbert}
\bibfield{author}{\bibinfo{person}{KyoHoon Jin}, \bibinfo{person}{JeongA Wi}, \bibinfo{person}{EunJu Lee}, \bibinfo{person}{ShinJin Kang}, \bibinfo{person}{SooKyun Kim}, {and} \bibinfo{person}{YoungBin Kim}.} \bibinfo{year}{2021}\natexlab{}.
\newblock \showarticletitle{TrafficBERT: Pre-trained model with large-scale data for long-range traffic flow forecasting}.
\newblock \bibinfo{journal}{\emph{Expert Systems with Applications}}  \bibinfo{volume}{186} (\bibinfo{year}{2021}), \bibinfo{pages}{115738}.
\newblock


\bibitem[Jin et~al\mbox{.}(2023a)]%
        {jin2023time}
\bibfield{author}{\bibinfo{person}{Ming Jin}, \bibinfo{person}{Shiyu Wang}, \bibinfo{person}{Lintao Ma}, \bibinfo{person}{Zhixuan Chu}, \bibinfo{person}{James~Y Zhang}, \bibinfo{person}{Xiaoming Shi}, \bibinfo{person}{Pin-Yu Chen}, \bibinfo{person}{Yuxuan Liang}, \bibinfo{person}{Yuan-Fang Li}, \bibinfo{person}{Shirui Pan}, {et~al\mbox{.}}} \bibinfo{year}{2023}\natexlab{a}.
\newblock \showarticletitle{Time-llm: Time series forecasting by reprogramming large language models}.
\newblock \bibinfo{journal}{\emph{arXiv preprint arXiv:2310.01728}} (\bibinfo{year}{2023}).
\newblock


\bibitem[Jin et~al\mbox{.}(2023b)]%
        {jin2023large}
\bibfield{author}{\bibinfo{person}{Ming Jin}, \bibinfo{person}{Qingsong Wen}, \bibinfo{person}{Yuxuan Liang}, \bibinfo{person}{Chaoli Zhang}, \bibinfo{person}{Siqiao Xue}, \bibinfo{person}{Xue Wang}, \bibinfo{person}{James Zhang}, \bibinfo{person}{Yi Wang}, \bibinfo{person}{Haifeng Chen}, \bibinfo{person}{Xiaoli Li}, {et~al\mbox{.}}} \bibinfo{year}{2023}\natexlab{b}.
\newblock \showarticletitle{Large models for time series and spatio-temporal data: A survey and outlook}.
\newblock \bibinfo{journal}{\emph{arXiv preprint arXiv:2310.10196}} (\bibinfo{year}{2023}).
\newblock


\bibitem[Jin et~al\mbox{.}(2022)]%
        {jin2022selective}
\bibfield{author}{\bibinfo{person}{Yilun Jin}, \bibinfo{person}{Kai Chen}, {and} \bibinfo{person}{Qiang Yang}.} \bibinfo{year}{2022}\natexlab{}.
\newblock \showarticletitle{Selective cross-city transfer learning for traffic prediction via source city region re-weighting}. In \bibinfo{booktitle}{\emph{Proceedings of the 28th ACM SIGKDD Conference on Knowledge Discovery and Data Mining}}. \bibinfo{pages}{731--741}.
\newblock


\bibitem[Kaplan et~al\mbox{.}(2020)]%
        {kaplan2020scaling}
\bibfield{author}{\bibinfo{person}{Jared Kaplan}, \bibinfo{person}{Sam McCandlish}, \bibinfo{person}{Tom Henighan}, \bibinfo{person}{Tom~B Brown}, \bibinfo{person}{Benjamin Chess}, \bibinfo{person}{Rewon Child}, \bibinfo{person}{Scott Gray}, \bibinfo{person}{Alec Radford}, \bibinfo{person}{Jeffrey Wu}, {and} \bibinfo{person}{Dario Amodei}.} \bibinfo{year}{2020}\natexlab{}.
\newblock \showarticletitle{Scaling laws for neural language models}.
\newblock \bibinfo{journal}{\emph{arXiv preprint arXiv:2001.08361}} (\bibinfo{year}{2020}).
\newblock


\bibitem[Kojima et~al\mbox{.}(2022)]%
        {kojima2022large}
\bibfield{author}{\bibinfo{person}{Takeshi Kojima}, \bibinfo{person}{Shixiang~Shane Gu}, \bibinfo{person}{Machel Reid}, \bibinfo{person}{Yutaka Matsuo}, {and} \bibinfo{person}{Yusuke Iwasawa}.} \bibinfo{year}{2022}\natexlab{}.
\newblock \showarticletitle{Large language models are zero-shot reasoners}.
\newblock \bibinfo{journal}{\emph{Advances in neural information processing systems}}  \bibinfo{volume}{35} (\bibinfo{year}{2022}), \bibinfo{pages}{22199--22213}.
\newblock


\bibitem[Lai et~al\mbox{.}(2023)]%
        {lai2023large}
\bibfield{author}{\bibinfo{person}{Siqi Lai}, \bibinfo{person}{Zhao Xu}, \bibinfo{person}{Weijia Zhang}, \bibinfo{person}{Hao Liu}, {and} \bibinfo{person}{Hui Xiong}.} \bibinfo{year}{2023}\natexlab{}.
\newblock \showarticletitle{Large Language Models as Traffic Signal Control Agents: Capacity and Opportunity}.
\newblock \bibinfo{journal}{\emph{arXiv preprint arXiv:2312.16044}} (\bibinfo{year}{2023}).
\newblock


\bibitem[Li et~al\mbox{.}(2023a)]%
        {li2023learning}
\bibfield{author}{\bibinfo{person}{Ruikun Li}, \bibinfo{person}{Huandong Wang}, {and} \bibinfo{person}{Yong Li}.} \bibinfo{year}{2023}\natexlab{a}.
\newblock \showarticletitle{Learning slow and fast system dynamics via automatic separation of time scales}. In \bibinfo{booktitle}{\emph{Proceedings of the 29th ACM SIGKDD Conference on Knowledge Discovery and Data Mining}}. \bibinfo{pages}{4380--4390}.
\newblock


\bibitem[Li et~al\mbox{.}(2018)]%
        {li2018diffusion}
\bibfield{author}{\bibinfo{person}{Yaguang Li}, \bibinfo{person}{Rose Yu}, \bibinfo{person}{Cyrus Shahabi}, {and} \bibinfo{person}{Yan Liu}.} \bibinfo{year}{2018}\natexlab{}.
\newblock \showarticletitle{Diffusion Convolutional Recurrent Neural Network: Data-Driven Traffic Forecasting}. In \bibinfo{booktitle}{\emph{International Conference on Learning Representations}}.
\newblock


\bibitem[Li et~al\mbox{.}(2024)]%
        {li2024urbangpt}
\bibfield{author}{\bibinfo{person}{Zhonghang Li}, \bibinfo{person}{Lianghao Xia}, \bibinfo{person}{Jiabin Tang}, \bibinfo{person}{Yong Xu}, \bibinfo{person}{Lei Shi}, \bibinfo{person}{Long Xia}, \bibinfo{person}{Dawei Yin}, {and} \bibinfo{person}{Chao Huang}.} \bibinfo{year}{2024}\natexlab{}.
\newblock \bibinfo{title}{UrbanGPT: Spatio-Temporal Large Language Models}.
\newblock
\newblock
\showeprint[arxiv]{2403.00813}~[cs.CL]


\bibitem[Li et~al\mbox{.}(2023b)]%
        {li2023generative}
\bibfield{author}{\bibinfo{person}{Zhonghang Li}, \bibinfo{person}{Lianghao Xia}, \bibinfo{person}{Yong Xu}, {and} \bibinfo{person}{Chao Huang}.} \bibinfo{year}{2023}\natexlab{b}.
\newblock \showarticletitle{Generative Pre-Training of Spatio-Temporal Graph Neural Networks}. In \bibinfo{booktitle}{\emph{Thirty-seventh Conference on Neural Information Processing Systems}}.
\newblock
\urldef\tempurl%
\url{https://openreview.net/forum?id=nMH5cUaSj8}
\showURL{%
\tempurl}


\bibitem[Lin et~al\mbox{.}(2020)]%
        {lin2020self}
\bibfield{author}{\bibinfo{person}{Zhihui Lin}, \bibinfo{person}{Maomao Li}, \bibinfo{person}{Zhuobin Zheng}, \bibinfo{person}{Yangyang Cheng}, {and} \bibinfo{person}{Chun Yuan}.} \bibinfo{year}{2020}\natexlab{}.
\newblock \showarticletitle{Self-attention convlstm for spatiotemporal prediction}. In \bibinfo{booktitle}{\emph{Proceedings of the AAAI conference on artificial intelligence}}, Vol.~\bibinfo{volume}{34}. \bibinfo{pages}{11531--11538}.
\newblock


\bibitem[Liu et~al\mbox{.}(2024)]%
        {liu2024large}
\bibfield{author}{\bibinfo{person}{Lei Liu}, \bibinfo{person}{Shuo Yu}, \bibinfo{person}{Runze Wang}, \bibinfo{person}{Zhenxun Ma}, {and} \bibinfo{person}{Yanming Shen}.} \bibinfo{year}{2024}\natexlab{}.
\newblock \bibinfo{title}{How Can Large Language Models Understand Spatial-Temporal Data?}
\newblock
\newblock
\showeprint[arxiv]{2401.14192}~[cs.LG]


\bibitem[Liu et~al\mbox{.}(2018)]%
        {liu2018attentive}
\bibfield{author}{\bibinfo{person}{Lingbo Liu}, \bibinfo{person}{Ruimao Zhang}, \bibinfo{person}{Jiefeng Peng}, \bibinfo{person}{Guanbin Li}, \bibinfo{person}{Bowen Du}, {and} \bibinfo{person}{Liang Lin}.} \bibinfo{year}{2018}\natexlab{}.
\newblock \showarticletitle{Attentive crowd flow machines}. In \bibinfo{booktitle}{\emph{Proceedings of the 26th ACM international conference on Multimedia}}. \bibinfo{pages}{1553--1561}.
\newblock


\bibitem[Liu et~al\mbox{.}(2023c)]%
        {liu2023pre}
\bibfield{author}{\bibinfo{person}{Pengfei Liu}, \bibinfo{person}{Weizhe Yuan}, \bibinfo{person}{Jinlan Fu}, \bibinfo{person}{Zhengbao Jiang}, \bibinfo{person}{Hiroaki Hayashi}, {and} \bibinfo{person}{Graham Neubig}.} \bibinfo{year}{2023}\natexlab{c}.
\newblock \showarticletitle{Pre-train, prompt, and predict: A systematic survey of prompting methods in natural language processing}.
\newblock \bibinfo{journal}{\emph{Comput. Surveys}} \bibinfo{volume}{55}, \bibinfo{number}{9} (\bibinfo{year}{2023}), \bibinfo{pages}{1--35}.
\newblock


\bibitem[Liu et~al\mbox{.}(2023a)]%
        {liu2023itransformer}
\bibfield{author}{\bibinfo{person}{Yong Liu}, \bibinfo{person}{Tengge Hu}, \bibinfo{person}{Haoran Zhang}, \bibinfo{person}{Haixu Wu}, \bibinfo{person}{Shiyu Wang}, \bibinfo{person}{Lintao Ma}, {and} \bibinfo{person}{Mingsheng Long}.} \bibinfo{year}{2023}\natexlab{a}.
\newblock \showarticletitle{itransformer: Inverted transformers are effective for time series forecasting}.
\newblock \bibinfo{journal}{\emph{arXiv preprint arXiv:2310.06625}} (\bibinfo{year}{2023}).
\newblock


\bibitem[Liu et~al\mbox{.}(2023b)]%
        {liu2023koopa}
\bibfield{author}{\bibinfo{person}{Yong Liu}, \bibinfo{person}{Chenyu Li}, \bibinfo{person}{Jianmin Wang}, {and} \bibinfo{person}{Mingsheng Long}.} \bibinfo{year}{2023}\natexlab{b}.
\newblock \showarticletitle{Koopa: Learning Non-stationary Time Series Dynamics with Koopman Predictors}.
\newblock \bibinfo{journal}{\emph{arXiv preprint arXiv:2305.18803}} (\bibinfo{year}{2023}).
\newblock


\bibitem[Liu et~al\mbox{.}(2023d)]%
        {liu2023cross}
\bibfield{author}{\bibinfo{person}{Zhanyu Liu}, \bibinfo{person}{Guanjie Zheng}, {and} \bibinfo{person}{Yanwei Yu}.} \bibinfo{year}{2023}\natexlab{d}.
\newblock \showarticletitle{Cross-city Few-Shot Traffic Forecasting via Traffic Pattern Bank}. In \bibinfo{booktitle}{\emph{Proceedings of the 32nd ACM International Conference on Information and Knowledge Management}}. \bibinfo{pages}{1451--1460}.
\newblock


\bibitem[Lu et~al\mbox{.}(2022)]%
        {lu2022spatio}
\bibfield{author}{\bibinfo{person}{Bin Lu}, \bibinfo{person}{Xiaoying Gan}, \bibinfo{person}{Weinan Zhang}, \bibinfo{person}{Huaxiu Yao}, \bibinfo{person}{Luoyi Fu}, {and} \bibinfo{person}{Xinbing Wang}.} \bibinfo{year}{2022}\natexlab{}.
\newblock \showarticletitle{Spatio-Temporal Graph Few-Shot Learning with Cross-City Knowledge Transfer}. In \bibinfo{booktitle}{\emph{Proceedings of the 28th ACM SIGKDD Conference on Knowledge Discovery and Data Mining}}. \bibinfo{pages}{1162--1172}.
\newblock


\bibitem[Nie et~al\mbox{.}(2022)]%
        {nie2022time}
\bibfield{author}{\bibinfo{person}{Yuqi Nie}, \bibinfo{person}{Nam~H Nguyen}, \bibinfo{person}{Phanwadee Sinthong}, {and} \bibinfo{person}{Jayant Kalagnanam}.} \bibinfo{year}{2022}\natexlab{}.
\newblock \showarticletitle{A time series is worth 64 words: Long-term forecasting with transformers}.
\newblock \bibinfo{journal}{\emph{arXiv preprint arXiv:2211.14730}} (\bibinfo{year}{2022}).
\newblock


\bibitem[Ouyang et~al\mbox{.}(2023)]%
        {ouyang2023citytrans}
\bibfield{author}{\bibinfo{person}{Xiaocao Ouyang}, \bibinfo{person}{Yan Yang}, \bibinfo{person}{Wei Zhou}, \bibinfo{person}{Yiling Zhang}, \bibinfo{person}{Hao Wang}, {and} \bibinfo{person}{Wei Huang}.} \bibinfo{year}{2023}\natexlab{}.
\newblock \showarticletitle{CityTrans: Domain-Adversarial Training with Knowledge Transfer for Spatio-Temporal Prediction across Cities}.
\newblock \bibinfo{journal}{\emph{IEEE Transactions on Knowledge and Data Engineering}} (\bibinfo{year}{2023}).
\newblock


\bibitem[Pan et~al\mbox{.}(2019)]%
        {pan2019urban}
\bibfield{author}{\bibinfo{person}{Zheyi Pan}, \bibinfo{person}{Yuxuan Liang}, \bibinfo{person}{Weifeng Wang}, \bibinfo{person}{Yong Yu}, \bibinfo{person}{Yu Zheng}, {and} \bibinfo{person}{Junbo Zhang}.} \bibinfo{year}{2019}\natexlab{}.
\newblock \showarticletitle{Urban traffic prediction from spatio-temporal data using deep meta learning}. In \bibinfo{booktitle}{\emph{Proceedings of the 25th ACM SIGKDD international conference on knowledge discovery \& data mining}}. \bibinfo{pages}{1720--1730}.
\newblock


\bibitem[Qiao et~al\mbox{.}(2022)]%
        {qiao2022reasoning}
\bibfield{author}{\bibinfo{person}{Shuofei Qiao}, \bibinfo{person}{Yixin Ou}, \bibinfo{person}{Ningyu Zhang}, \bibinfo{person}{Xiang Chen}, \bibinfo{person}{Yunzhi Yao}, \bibinfo{person}{Shumin Deng}, \bibinfo{person}{Chuanqi Tan}, \bibinfo{person}{Fei Huang}, {and} \bibinfo{person}{Huajun Chen}.} \bibinfo{year}{2022}\natexlab{}.
\newblock \showarticletitle{Reasoning with language model prompting: A survey}.
\newblock \bibinfo{journal}{\emph{arXiv preprint arXiv:2212.09597}} (\bibinfo{year}{2022}).
\newblock


\bibitem[Rombach et~al\mbox{.}(2022)]%
        {rombach2022high}
\bibfield{author}{\bibinfo{person}{Robin Rombach}, \bibinfo{person}{Andreas Blattmann}, \bibinfo{person}{Dominik Lorenz}, \bibinfo{person}{Patrick Esser}, {and} \bibinfo{person}{Bj{\"o}rn Ommer}.} \bibinfo{year}{2022}\natexlab{}.
\newblock \showarticletitle{High-resolution image synthesis with latent diffusion models}. In \bibinfo{booktitle}{\emph{Proceedings of the IEEE/CVF conference on computer vision and pattern recognition}}. \bibinfo{pages}{10684--10695}.
\newblock


\bibitem[Shao et~al\mbox{.}(2022b)]%
        {shao2022spatial}
\bibfield{author}{\bibinfo{person}{Zezhi Shao}, \bibinfo{person}{Zhao Zhang}, \bibinfo{person}{Fei Wang}, \bibinfo{person}{Wei Wei}, {and} \bibinfo{person}{Yongjun Xu}.} \bibinfo{year}{2022}\natexlab{b}.
\newblock \showarticletitle{Spatial-temporal identity: A simple yet effective baseline for multivariate time series forecasting}. In \bibinfo{booktitle}{\emph{Proceedings of the 31st ACM International Conference on Information \& Knowledge Management}}. \bibinfo{pages}{4454--4458}.
\newblock


\bibitem[Shao et~al\mbox{.}(2022a)]%
        {ShaoZWX22}
\bibfield{author}{\bibinfo{person}{Zezhi Shao}, \bibinfo{person}{Zhao Zhang}, \bibinfo{person}{Fei Wang}, {and} \bibinfo{person}{Yongjun Xu}.} \bibinfo{year}{2022}\natexlab{a}.
\newblock \showarticletitle{Pre-training Enhanced Spatial-temporal Graph Neural Network for Multivariate Time Series Forecasting}. In \bibinfo{booktitle}{\emph{{KDD} '22: The 28th {ACM} {SIGKDD} Conference on Knowledge Discovery and Data Mining, Washington, DC, USA, August 14 - 18, 2022}}. \bibinfo{publisher}{{ACM}}, \bibinfo{pages}{1567--1577}.
\newblock


\bibitem[Shen et~al\mbox{.}(2024)]%
        {shen2024multitask}
\bibfield{author}{\bibinfo{person}{Sheng Shen}, \bibinfo{person}{Shijia Yang}, \bibinfo{person}{Tianjun Zhang}, \bibinfo{person}{Bohan Zhai}, \bibinfo{person}{Joseph~E Gonzalez}, \bibinfo{person}{Kurt Keutzer}, {and} \bibinfo{person}{Trevor Darrell}.} \bibinfo{year}{2024}\natexlab{}.
\newblock \showarticletitle{Multitask vision-language prompt tuning}. In \bibinfo{booktitle}{\emph{Proceedings of the IEEE/CVF Winter Conference on Applications of Computer Vision}}. \bibinfo{pages}{5656--5667}.
\newblock


\bibitem[Sukhbaatar et~al\mbox{.}(2015)]%
        {sukhbaatar2015end}
\bibfield{author}{\bibinfo{person}{Sainbayar Sukhbaatar}, \bibinfo{person}{Jason Weston}, \bibinfo{person}{Rob Fergus}, {et~al\mbox{.}}} \bibinfo{year}{2015}\natexlab{}.
\newblock \showarticletitle{End-to-end memory networks}.
\newblock \bibinfo{journal}{\emph{Advances in neural information processing systems}}  \bibinfo{volume}{28} (\bibinfo{year}{2015}).
\newblock


\bibitem[Tan et~al\mbox{.}(2023)]%
        {tan2023temporal}
\bibfield{author}{\bibinfo{person}{Cheng Tan}, \bibinfo{person}{Zhangyang Gao}, \bibinfo{person}{Lirong Wu}, \bibinfo{person}{Yongjie Xu}, \bibinfo{person}{Jun Xia}, \bibinfo{person}{Siyuan Li}, {and} \bibinfo{person}{Stan~Z Li}.} \bibinfo{year}{2023}\natexlab{}.
\newblock \showarticletitle{Temporal attention unit: Towards efficient spatiotemporal predictive learning}. In \bibinfo{booktitle}{\emph{Proceedings of the IEEE/CVF Conference on Computer Vision and Pattern Recognition}}. \bibinfo{pages}{18770--18782}.
\newblock


\bibitem[Tang et~al\mbox{.}(2022)]%
        {tang2022domain}
\bibfield{author}{\bibinfo{person}{Yihong Tang}, \bibinfo{person}{Ao Qu}, \bibinfo{person}{Andy~HF Chow}, \bibinfo{person}{William~HK Lam}, \bibinfo{person}{SC Wong}, {and} \bibinfo{person}{Wei Ma}.} \bibinfo{year}{2022}\natexlab{}.
\newblock \showarticletitle{Domain adversarial spatial-temporal network: a transferable framework for short-term traffic forecasting across cities}. In \bibinfo{booktitle}{\emph{Proceedings of the 31st ACM International Conference on Information \& Knowledge Management}}. \bibinfo{pages}{1905--1915}.
\newblock


\bibitem[Touvron et~al\mbox{.}(2023)]%
        {touvron2023llama}
\bibfield{author}{\bibinfo{person}{Hugo Touvron}, \bibinfo{person}{Louis Martin}, \bibinfo{person}{Kevin Stone}, \bibinfo{person}{Peter Albert}, \bibinfo{person}{Amjad Almahairi}, \bibinfo{person}{Yasmine Babaei}, \bibinfo{person}{Nikolay Bashlykov}, \bibinfo{person}{Soumya Batra}, \bibinfo{person}{Prajjwal Bhargava}, \bibinfo{person}{Shruti Bhosale}, {et~al\mbox{.}}} \bibinfo{year}{2023}\natexlab{}.
\newblock \showarticletitle{Llama 2: Open foundation and fine-tuned chat models}.
\newblock \bibinfo{journal}{\emph{arXiv preprint arXiv:2307.09288}} (\bibinfo{year}{2023}).
\newblock


\bibitem[Wang et~al\mbox{.}(2020)]%
        {wang2020deep}
\bibfield{author}{\bibinfo{person}{Senzhang Wang}, \bibinfo{person}{Jiannong Cao}, {and} \bibinfo{person}{S~Yu Philip}.} \bibinfo{year}{2020}\natexlab{}.
\newblock \showarticletitle{Deep learning for spatio-temporal data mining: A survey}.
\newblock \bibinfo{journal}{\emph{IEEE transactions on knowledge and data engineering}} \bibinfo{volume}{34}, \bibinfo{number}{8} (\bibinfo{year}{2020}), \bibinfo{pages}{3681--3700}.
\newblock


\bibitem[Wang et~al\mbox{.}(2023b)]%
        {wang2023building}
\bibfield{author}{\bibinfo{person}{Xuhong Wang}, \bibinfo{person}{Ding Wang}, \bibinfo{person}{Liang Chen}, {and} \bibinfo{person}{Yilun Lin}.} \bibinfo{year}{2023}\natexlab{b}.
\newblock \bibinfo{title}{Building Transportation Foundation Model via Generative Graph Transformer}.
\newblock
\newblock
\showeprint[arxiv]{2305.14826}~[cs.LG]


\bibitem[Wang et~al\mbox{.}(2018)]%
        {wang2018predrnn++}
\bibfield{author}{\bibinfo{person}{Yunbo Wang}, \bibinfo{person}{Zhifeng Gao}, \bibinfo{person}{Mingsheng Long}, \bibinfo{person}{Jianmin Wang}, {and} \bibinfo{person}{S~Yu Philip}.} \bibinfo{year}{2018}\natexlab{}.
\newblock \showarticletitle{Predrnn++: Towards a resolution of the deep-in-time dilemma in spatiotemporal predictive learning}. In \bibinfo{booktitle}{\emph{International Conference on Machine Learning}}. PMLR, \bibinfo{pages}{5123--5132}.
\newblock


\bibitem[Wang et~al\mbox{.}(2017)]%
        {wang2017predrnn}
\bibfield{author}{\bibinfo{person}{Yunbo Wang}, \bibinfo{person}{Mingsheng Long}, \bibinfo{person}{Jianmin Wang}, \bibinfo{person}{Zhifeng Gao}, {and} \bibinfo{person}{Philip~S Yu}.} \bibinfo{year}{2017}\natexlab{}.
\newblock \showarticletitle{Predrnn: Recurrent neural networks for predictive learning using spatiotemporal lstms}.
\newblock \bibinfo{journal}{\emph{Advances in neural information processing systems}}  \bibinfo{volume}{30} (\bibinfo{year}{2017}).
\newblock


\bibitem[Wang et~al\mbox{.}(2019)]%
        {wang2019memory}
\bibfield{author}{\bibinfo{person}{Yunbo Wang}, \bibinfo{person}{Jianjin Zhang}, \bibinfo{person}{Hongyu Zhu}, \bibinfo{person}{Mingsheng Long}, \bibinfo{person}{Jianmin Wang}, {and} \bibinfo{person}{Philip~S Yu}.} \bibinfo{year}{2019}\natexlab{}.
\newblock \showarticletitle{Memory in memory: A predictive neural network for learning higher-order non-stationarity from spatiotemporal dynamics}. In \bibinfo{booktitle}{\emph{Proceedings of the IEEE/CVF conference on computer vision and pattern recognition}}. \bibinfo{pages}{9154--9162}.
\newblock


\bibitem[Wang et~al\mbox{.}(2023a)]%
        {wang2023detecting}
\bibfield{author}{\bibinfo{person}{Zhenyu Wang}, \bibinfo{person}{Yali Li}, \bibinfo{person}{Xi Chen}, \bibinfo{person}{Ser-Nam Lim}, \bibinfo{person}{Antonio Torralba}, \bibinfo{person}{Hengshuang Zhao}, {and} \bibinfo{person}{Shengjin Wang}.} \bibinfo{year}{2023}\natexlab{a}.
\newblock \showarticletitle{Detecting everything in the open world: Towards universal object detection}. In \bibinfo{booktitle}{\emph{Proceedings of the IEEE/CVF Conference on Computer Vision and Pattern Recognition}}. \bibinfo{pages}{11433--11443}.
\newblock


\bibitem[Wang et~al\mbox{.}(2022)]%
        {wang2022learning}
\bibfield{author}{\bibinfo{person}{Zifeng Wang}, \bibinfo{person}{Zizhao Zhang}, \bibinfo{person}{Chen-Yu Lee}, \bibinfo{person}{Han Zhang}, \bibinfo{person}{Ruoxi Sun}, \bibinfo{person}{Xiaoqi Ren}, \bibinfo{person}{Guolong Su}, \bibinfo{person}{Vincent Perot}, \bibinfo{person}{Jennifer Dy}, {and} \bibinfo{person}{Tomas Pfister}.} \bibinfo{year}{2022}\natexlab{}.
\newblock \showarticletitle{Learning to prompt for continual learning}. In \bibinfo{booktitle}{\emph{Proceedings of the IEEE/CVF Conference on Computer Vision and Pattern Recognition}}. \bibinfo{pages}{139--149}.
\newblock


\bibitem[Wu et~al\mbox{.}(2022)]%
        {wu2022timesnet}
\bibfield{author}{\bibinfo{person}{Haixu Wu}, \bibinfo{person}{Tengge Hu}, \bibinfo{person}{Yong Liu}, \bibinfo{person}{Hang Zhou}, \bibinfo{person}{Jianmin Wang}, {and} \bibinfo{person}{Mingsheng Long}.} \bibinfo{year}{2022}\natexlab{}.
\newblock \showarticletitle{Timesnet: Temporal 2d-variation modeling for general time series analysis}.
\newblock \bibinfo{journal}{\emph{arXiv preprint arXiv:2210.02186}} (\bibinfo{year}{2022}).
\newblock


\bibitem[Xu et~al\mbox{.}(2023)]%
        {xu2023urban}
\bibfield{author}{\bibinfo{person}{Fengli Xu}, \bibinfo{person}{Jun Zhang}, \bibinfo{person}{Chen Gao}, \bibinfo{person}{Jie Feng}, {and} \bibinfo{person}{Yong Li}.} \bibinfo{year}{2023}\natexlab{}.
\newblock \showarticletitle{Urban Generative Intelligence (UGI): A Foundational Platform for Agents in Embodied City Environment}.
\newblock \bibinfo{journal}{\emph{arXiv preprint arXiv:2312.11813}} (\bibinfo{year}{2023}).
\newblock


\bibitem[Yan et~al\mbox{.}(2023)]%
        {yan2023urban}
\bibfield{author}{\bibinfo{person}{Yibo Yan}, \bibinfo{person}{Haomin Wen}, \bibinfo{person}{Siru Zhong}, \bibinfo{person}{Wei Chen}, \bibinfo{person}{Haodong Chen}, \bibinfo{person}{Qingsong Wen}, \bibinfo{person}{Roger Zimmermann}, {and} \bibinfo{person}{Yuxuan Liang}.} \bibinfo{year}{2023}\natexlab{}.
\newblock \showarticletitle{When Urban Region Profiling Meets Large Language Models}.
\newblock \bibinfo{journal}{\emph{arXiv preprint arXiv:2310.18340}} (\bibinfo{year}{2023}).
\newblock


\bibitem[Yao et~al\mbox{.}(2019)]%
        {yao2019learning}
\bibfield{author}{\bibinfo{person}{Huaxiu Yao}, \bibinfo{person}{Yiding Liu}, \bibinfo{person}{Ying Wei}, \bibinfo{person}{Xianfeng Tang}, {and} \bibinfo{person}{Zhenhui Li}.} \bibinfo{year}{2019}\natexlab{}.
\newblock \showarticletitle{Learning from multiple cities: A meta-learning approach for spatial-temporal prediction}. In \bibinfo{booktitle}{\emph{The world wide web conference}}. \bibinfo{pages}{2181--2191}.
\newblock


\bibitem[Yu et~al\mbox{.}(2020)]%
        {yu2020spatio}
\bibfield{author}{\bibinfo{person}{Cunjun Yu}, \bibinfo{person}{Xiao Ma}, \bibinfo{person}{Jiawei Ren}, \bibinfo{person}{Haiyu Zhao}, {and} \bibinfo{person}{Shuai Yi}.} \bibinfo{year}{2020}\natexlab{}.
\newblock \showarticletitle{Spatio-temporal graph transformer networks for pedestrian trajectory prediction}. In \bibinfo{booktitle}{\emph{Computer Vision--ECCV 2020: 16th European Conference, Glasgow, UK, August 23--28, 2020, Proceedings, Part XII 16}}. Springer, \bibinfo{pages}{507--523}.
\newblock


\bibitem[Yuan et~al\mbox{.}(2023)]%
        {yuan2023spatio}
\bibfield{author}{\bibinfo{person}{Yuan Yuan}, \bibinfo{person}{Jingtao Ding}, \bibinfo{person}{Chenyang Shao}, \bibinfo{person}{Depeng Jin}, {and} \bibinfo{person}{Yong Li}.} \bibinfo{year}{2023}\natexlab{}.
\newblock \showarticletitle{Spatio-temporal Diffusion Point Processes}.
\newblock \bibinfo{journal}{\emph{arXiv preprint arXiv:2305.12403}} (\bibinfo{year}{2023}).
\newblock


\bibitem[Yuan et~al\mbox{.}(2024)]%
        {yuan2024spatiotemporal}
\bibfield{author}{\bibinfo{person}{Yuan Yuan}, \bibinfo{person}{Chenyang Shao}, \bibinfo{person}{Jingtao Ding}, \bibinfo{person}{Depeng Jin}, {and} \bibinfo{person}{Yong Li}.} \bibinfo{year}{2024}\natexlab{}.
\newblock \showarticletitle{Spatio-Temporal Few-Shot Learning via Diffusive Neural Network Generation}. In \bibinfo{booktitle}{\emph{The Twelfth International Conference on Learning Representations}}.
\newblock
\urldef\tempurl%
\url{https://openreview.net/forum?id=QyFm3D3Tzi}
\showURL{%
\tempurl}


\bibitem[Zhang et~al\mbox{.}(2017)]%
        {zhang2017deep}
\bibfield{author}{\bibinfo{person}{Junbo Zhang}, \bibinfo{person}{Yu Zheng}, {and} \bibinfo{person}{Dekang Qi}.} \bibinfo{year}{2017}\natexlab{}.
\newblock \showarticletitle{Deep spatio-temporal residual networks for citywide crowd flows prediction}. In \bibinfo{booktitle}{\emph{Proceedings of the AAAI conference on artificial intelligence}}, Vol.~\bibinfo{volume}{31}.
\newblock


\bibitem[Zhang et~al\mbox{.}(2023a)]%
        {zhang2023mask}
\bibfield{author}{\bibinfo{person}{Xu Zhang}, \bibinfo{person}{Yongshun Gong}, \bibinfo{person}{Xinxin Zhang}, \bibinfo{person}{Xiaoming Wu}, \bibinfo{person}{Chengqi Zhang}, {and} \bibinfo{person}{Xiangjun Dong}.} \bibinfo{year}{2023}\natexlab{a}.
\newblock \showarticletitle{Mask-and Contrast-Enhanced Spatio-Temporal Learning for Urban Flow Prediction}. In \bibinfo{booktitle}{\emph{Proceedings of the 32nd ACM International Conference on Information and Knowledge Management}}. \bibinfo{pages}{3298--3307}.
\newblock


\bibitem[Zhang et~al\mbox{.}(2023b)]%
        {zhang2023mlpst}
\bibfield{author}{\bibinfo{person}{Zijian Zhang}, \bibinfo{person}{Ze Huang}, \bibinfo{person}{Zhiwei Hu}, \bibinfo{person}{Xiangyu Zhao}, \bibinfo{person}{Wanyu Wang}, \bibinfo{person}{Zitao Liu}, \bibinfo{person}{Junbo Zhang}, \bibinfo{person}{S~Joe Qin}, {and} \bibinfo{person}{Hongwei Zhao}.} \bibinfo{year}{2023}\natexlab{b}.
\newblock \showarticletitle{MLPST: MLP is All You Need for Spatio-Temporal Prediction}. In \bibinfo{booktitle}{\emph{Proceedings of the 32nd ACM International Conference on Information and Knowledge Management}}. \bibinfo{pages}{3381--3390}.
\newblock


\bibitem[Zhang et~al\mbox{.}(2023c)]%
        {zhang2023promptst}
\bibfield{author}{\bibinfo{person}{Zijian Zhang}, \bibinfo{person}{Xiangyu Zhao}, \bibinfo{person}{Qidong Liu}, \bibinfo{person}{Chunxu Zhang}, \bibinfo{person}{Qian Ma}, \bibinfo{person}{Wanyu Wang}, \bibinfo{person}{Hongwei Zhao}, \bibinfo{person}{Yiqi Wang}, {and} \bibinfo{person}{Zitao Liu}.} \bibinfo{year}{2023}\natexlab{c}.
\newblock \showarticletitle{PromptST: Prompt-Enhanced Spatio-Temporal Multi-Attribute Prediction}. In \bibinfo{booktitle}{\emph{Proceedings of the 32nd ACM International Conference on Information and Knowledge Management}}. \bibinfo{pages}{3195--3205}.
\newblock


\bibitem[Zhao et~al\mbox{.}(2022)]%
        {zhao2022st}
\bibfield{author}{\bibinfo{person}{Liang Zhao}, \bibinfo{person}{Min Gao}, {and} \bibinfo{person}{Zongwei Wang}.} \bibinfo{year}{2022}\natexlab{}.
\newblock \showarticletitle{St-gsp: Spatial-temporal global semantic representation learning for urban flow prediction}. In \bibinfo{booktitle}{\emph{Proceedings of the Fifteenth ACM International Conference on Web Search and Data Mining}}. \bibinfo{pages}{1443--1451}.
\newblock


\bibitem[Zhao et~al\mbox{.}(2019)]%
        {zhao2019t}
\bibfield{author}{\bibinfo{person}{Ling Zhao}, \bibinfo{person}{Yujiao Song}, \bibinfo{person}{Chao Zhang}, \bibinfo{person}{Yu Liu}, \bibinfo{person}{Pu Wang}, \bibinfo{person}{Tao Lin}, \bibinfo{person}{Min Deng}, {and} \bibinfo{person}{Haifeng Li}.} \bibinfo{year}{2019}\natexlab{}.
\newblock \showarticletitle{T-gcn: A temporal graph convolutional network for traffic prediction}.
\newblock \bibinfo{journal}{\emph{IEEE transactions on intelligent transportation systems}} \bibinfo{volume}{21}, \bibinfo{number}{9} (\bibinfo{year}{2019}), \bibinfo{pages}{3848--3858}.
\newblock


\bibitem[Zheng et~al\mbox{.}(2014)]%
        {zheng2014urban}
\bibfield{author}{\bibinfo{person}{Yu Zheng}, \bibinfo{person}{Licia Capra}, \bibinfo{person}{Ouri Wolfson}, {and} \bibinfo{person}{Hai Yang}.} \bibinfo{year}{2014}\natexlab{}.
\newblock \showarticletitle{Urban computing: concepts, methodologies, and applications}.
\newblock \bibinfo{journal}{\emph{ACM Transactions on Intelligent Systems and Technology (TIST)}} \bibinfo{volume}{5}, \bibinfo{number}{3} (\bibinfo{year}{2014}), \bibinfo{pages}{1--55}.
\newblock


\bibitem[Zhou et~al\mbox{.}(2023b)]%
        {zhou2023one}
\bibfield{author}{\bibinfo{person}{Tian Zhou}, \bibinfo{person}{Peisong Niu}, \bibinfo{person}{Xue Wang}, \bibinfo{person}{Liang Sun}, {and} \bibinfo{person}{Rong Jin}.} \bibinfo{year}{2023}\natexlab{b}.
\newblock \showarticletitle{One Fits All: Power General Time Series Analysis by Pretrained LM}.
\newblock \bibinfo{journal}{\emph{arXiv preprint arXiv:2302.11939}} (\bibinfo{year}{2023}).
\newblock


\bibitem[Zhou et~al\mbox{.}(2023a)]%
        {zhou2023towards}
\bibfield{author}{\bibinfo{person}{Zhilun Zhou}, \bibinfo{person}{Jingtao Ding}, \bibinfo{person}{Yu Liu}, \bibinfo{person}{Depeng Jin}, {and} \bibinfo{person}{Yong Li}.} \bibinfo{year}{2023}\natexlab{a}.
\newblock \showarticletitle{Towards Generative Modeling of Urban Flow through Knowledge-enhanced Denoising Diffusion}. In \bibinfo{booktitle}{\emph{Proceedings of the 31st ACM International Conference on Advances in Geographic Information Systems}}. \bibinfo{pages}{1--12}.
\newblock


\bibitem[Zhou et~al\mbox{.}(2023c)]%
        {zhou2023predicting}
\bibfield{author}{\bibinfo{person}{Zhengyang Zhou}, \bibinfo{person}{Kuo Yang}, \bibinfo{person}{Yuxuan Liang}, \bibinfo{person}{Binwu Wang}, \bibinfo{person}{Hongyang Chen}, {and} \bibinfo{person}{Yang Wang}.} \bibinfo{year}{2023}\natexlab{c}.
\newblock \showarticletitle{Predicting collective human mobility via countering spatiotemporal heterogeneity}.
\newblock \bibinfo{journal}{\emph{IEEE Transactions on Mobile Computing}} (\bibinfo{year}{2023}).
\newblock


\end{thebibliography}



\section*{Appendix}

\appendix

\section{Datasets}\label{sup:dataset}

\subsection{Basic Information}

Here we provide more details of the used datasets in our study. 
We collect various spatio-temporal data from multiple cities and domains. Table~\ref{tbl:dataset_info} summarizes the basic information of the used datasets, and Table~\ref{tbl:dataset_st} reports the basic statistics.  
Specifically, values for Crowd and Cellular datasets in Table~\ref{tbl:short_term}, Table~\ref{tbl:long_term}, Table~\ref{tbl:long1}, Table~\ref{tbl:few_shot_crowd} and Figure~\ref{fig:few_zero} should be scaled by a factor of $10^3$.

\subsection{Data Preprocessing}

For each dataset, We split it into three non-overlapping periods: the first 70\% of the period was used as the training set, the next 15\% as the validation set, and the final 15\% as the test set. To ensure no overlap between train/val/test sets, we removed intermediate sequences. We have normalized all datasets to the range $[-1, 1]$. The reported prediction results are denormalized results.

\begin{table*}[t!]
\caption{The basic information of the used datasets. 
}\label{tbl:dataset_info}
\begin{threeparttable}
\resizebox{2\columnwidth}{!}{
\begin{tabular}{cccccc}
\toprule
Dataset  & Domain & City & Temporal Duration & Temporal interval & Spatial partition  \\ 
\cmidrule(lr){1-1} \cmidrule(lr){2-2}   \cmidrule(lr){3-3}    \cmidrule(lr){4-4}   \cmidrule(lr){5-5}  \cmidrule(lr){6-6} 
\multirow{4}{*}{TaxiBJ} &  \multirow{4}{*}{Taxi GPS} & \multirow{4}{*}{Beijing, China} &  20130601-20131030  & \multirow{4}{*}{Half an hour} & \multirow{4}{*}{$32 \times 32$}  \\ 
 & &  & 20140301-20140630 & &  \\ 
  & &  & 20150301-20150630 & &  \\ 
   & &  & 20151101-20160410 & &  \\ 
\cmidrule(lr){1-1} \cmidrule(lr){2-2}   \cmidrule(lr){3-3}    \cmidrule(lr){4-4}   \cmidrule(lr){5-5}  \cmidrule(lr){6-6}   
Cellular & Cellular usage & Nanjing, China & 20201111-20210531  & Half an hour & 16 * 20  \\ 
\cmidrule(lr){1-1} \cmidrule(lr){2-2}   \cmidrule(lr){3-3}    \cmidrule(lr){4-4}   \cmidrule(lr){5-5}  \cmidrule(lr){6-6}   
TaxiNYC-1 & Taxi OD & New York City, USA   & 20160101-20160229  &Half an hour &  16 * 12 \\ 
\cmidrule(lr){1-1} \cmidrule(lr){2-2}   \cmidrule(lr){3-3}    \cmidrule(lr){4-4}   \cmidrule(lr){5-5}  \cmidrule(lr){6-6}   
TaxiNYC-2 &  Taxi OD & New York City, USA   & 20150101-20150301 &Half an hour & 20 * 10  \\ 
\cmidrule(lr){1-1} \cmidrule(lr){2-2}   \cmidrule(lr){3-3}    \cmidrule(lr){4-4}   \cmidrule(lr){5-5}  \cmidrule(lr){6-6}   
BikeNYC-1 & Bike usage & New York City, USA    &20160801-20160929  & One hour & 16 * 8 \\ 
\cmidrule(lr){1-1} \cmidrule(lr){2-2}   \cmidrule(lr){3-3}    \cmidrule(lr){4-4}   \cmidrule(lr){5-5}  \cmidrule(lr){6-6}   
BikeNYC-2 & Bike usage &  New York City, USA   & 20160701-20160829  & Half an hour&  10 * 20\\ 
\cmidrule(lr){1-1} \cmidrule(lr){2-2}   \cmidrule(lr){3-3}    \cmidrule(lr){4-4}   \cmidrule(lr){5-5}  \cmidrule(lr){6-6}   
TDrive & Taxi trajectory & New York City, USA   & 20150201-20160602  & One hour & $32 \times 32$  \\ 
\cmidrule(lr){1-1} \cmidrule(lr){2-2}   \cmidrule(lr){3-3}    \cmidrule(lr){4-4}   \cmidrule(lr){5-5}  \cmidrule(lr){6-6}   
Crowd & Crowd flow &  Nanjing, China  & 20201111-20210531  & Half an hour &  16 * 20  \\ 
\cmidrule(lr){1-1} \cmidrule(lr){2-2}   \cmidrule(lr){3-3}    \cmidrule(lr){4-4}   \cmidrule(lr){5-5}  \cmidrule(lr){6-6}   
TrafficCS & Traffic speed  & Changsha, China &20220305-20220405 & Five minutes  &  $28\times 28$ \\
\cmidrule(lr){1-1} \cmidrule(lr){2-2}   \cmidrule(lr){3-3}    \cmidrule(lr){4-4}   \cmidrule(lr){5-5}  \cmidrule(lr){6-6}  
TrafficWH & Traffic speed  & Wuhan, China& 20220305-20220405 & Five minutes  & $30\times 28$  \\
\cmidrule(lr){1-1} \cmidrule(lr){2-2}   \cmidrule(lr){3-3}    \cmidrule(lr){4-4}   \cmidrule(lr){5-5}  \cmidrule(lr){6-6}  
TrafficCD & Traffic speed  & Chengdu, China& 20220305-20220405 & Five minutes  &  $28 \times 26$ \\
\cmidrule(lr){1-1} \cmidrule(lr){2-2}   \cmidrule(lr){3-3}    \cmidrule(lr){4-4}   \cmidrule(lr){5-5}  \cmidrule(lr){6-6}  
TrafficJN & Traffic speed  & Jinan, China& 20220305-20220405& Five minutes  &   $32\times 18$ \\
\cmidrule(lr){1-1} \cmidrule(lr){2-2}   \cmidrule(lr){3-3}    \cmidrule(lr){4-4}   \cmidrule(lr){5-5}  \cmidrule(lr){6-6}  
TrafficNJ & Traffic speed  & Nanjing, China& 20220305-20220405 & Five minutes  &  $32\times 24 $ \\
\cmidrule(lr){1-1} \cmidrule(lr){2-2}   \cmidrule(lr){3-3}    \cmidrule(lr){4-4}   \cmidrule(lr){5-5}  \cmidrule(lr){6-6}  
TrafficSH & Traffic speed  & Shanghai, China& 20220127-20220227& Five minutes  & $28\times 32$  \\
\cmidrule(lr){1-1} \cmidrule(lr){2-2}   \cmidrule(lr){3-3}    \cmidrule(lr){4-4}   \cmidrule(lr){5-5}  \cmidrule(lr){6-6}  
TrafficSZ & Traffic speed  & Shenzhen, China& 20220305-20220405 & Five minutes  & $24\times 18$  \\
\cmidrule(lr){1-1} \cmidrule(lr){2-2}   \cmidrule(lr){3-3}    \cmidrule(lr){4-4}   \cmidrule(lr){5-5}  \cmidrule(lr){6-6}  
TrafficGZ & Traffic speed  & Guangzhou, China& 20220305-20220405& Five minutes  &  $32\times 26$ \\
\cmidrule(lr){1-1} \cmidrule(lr){2-2}   \cmidrule(lr){3-3}    \cmidrule(lr){4-4}   \cmidrule(lr){5-5}  \cmidrule(lr){6-6}  
TrafficGY & Traffic speed  & Guiyang, China& 20220305-20220405 & Five minutes  &  $26\times 28$  \\
\cmidrule(lr){1-1} \cmidrule(lr){2-2}   \cmidrule(lr){3-3}    \cmidrule(lr){4-4}   \cmidrule(lr){5-5}  \cmidrule(lr){6-6}  
TrafficTJ & Traffic speed  & Tianjin, China& 20220305-20220405 & Five minutes  &  $24\times 30$  \\
\cmidrule(lr){1-1} \cmidrule(lr){2-2}   \cmidrule(lr){3-3}    \cmidrule(lr){4-4}   \cmidrule(lr){5-5}  \cmidrule(lr){6-6}  
TrafficHZ & Traffic speed  & Hangzhou, China& 20220305-20220405 & Five minutes  &  $28\times 24$ \\
\cmidrule(lr){1-1} \cmidrule(lr){2-2}   \cmidrule(lr){3-3}    \cmidrule(lr){4-4}   \cmidrule(lr){5-5}  \cmidrule(lr){6-6}  
TrafficZZ & Traffic speed  & Zhengzhou, China & 20220305-20220405 & Five minutes  & $26\times 26$  \\
\cmidrule(lr){1-1} \cmidrule(lr){2-2}   \cmidrule(lr){3-3}    \cmidrule(lr){4-4}   \cmidrule(lr){5-5}  \cmidrule(lr){6-6}  
TrafficBJ & Traffic speed  & Beijing, China & 20220305-20220405 & Five minutes  &  $30\times 32$  \\
\bottomrule
\end{tabular}}
\end{threeparttable}
\end{table*}

\section{Baselines}\label{sup:baseline}

\begin{itemize}[leftmargin=*]
    \item \textbf{HA}: History average uses the mean value of historical data for future predictions. Here we use historical data of corresponding periods in the past days. 
    \item \textbf{ARIMA}:  Auto-regressive Integrated Moving Average model a widely used statistical method for time series forecasting. It is a powerful tool for analyzing and predicting time series data, which are observations collected at regular intervals over time.
    \item \textbf{STResNet}~\cite{zhang2017deep}: It is a spatio-temporal model for crowd flow prediction, which utilizes residual neural networks to model the temporal closeness, period, and trend properties. 
    \item \textbf{ACFM}~\cite{liu2018attentive}: Attentive Crowd Flow Machine model is proposed to predict the dynamics of the crowd flows. It learns the dynamics by leveraging an attention mechanism to adaptively aggregate the sequential patterns and the periodic patterns. 
    \item \textbf{STGSP}~\cite{zhao2022st}: This model propose that the global information and positional information in the temporal dimension are important for spatio-temporal prediction. To this end, it leverages a semantic flow encoder to model the temporal relative positional signals. Besides, it utilizes an attention mechanism to capture the multi-scale temporal dependencies. 
    \item \textbf{MC-STL}~\cite{zhang2023mask}: It leverages an state-of-the-art training techniques for spatio-temporal predition, the mask-enhanced contrastive learning, which can effectively capture the relationships on the spatio-temporal dimension. 
    \item \textbf{MAU}~\cite{chang2021mau}: Motion-aware unit is a video prediction model. it broadens the temporal receptive fields of prediction units, which can facilitates to capture inter-frame motion correlations. It consists of an attention module and a fusion module. 
    \item \textbf{PredRNN}~\cite{wang2017predrnn}: PredRNN is a recurrent network-based model. In this model, the memory cells are explicitly decoupled, and they calculate in independent transition manners. Besides, different from the memory cell of LSTM, this network leverages zigzap memory flow, which facilitates to learn at distinct levels. 
    \item \textbf{MIM}~\cite{wang2019memory}: Memory  utilize the differential information between adjacent recurrent states, which facilitates to model the non-stationary  properties. Stacked multiple MIM blocks make it possible to model high-order non-stationarity.
    \item \textbf{SimVP}~\cite{gao2022simvp}: It is a simple yet very effective video prediction model. It is completely built based on convolutional neural networks and uses MSE loss. It serves as a solid baseline in video prediction tasks. 
    \item \textbf{TAU}~\cite{tan2023temporal}: Temporal Attention Unit is the state-of-the-art video prediction model. It decomposes the temporal attention into two parts: intra-frame attention and inter-frame attention, which are static and dynamical, respectively.  Besides, it introduces a novel regularization, \textit{i.e.,} differential divergence regularization, to consider the impact of inter-frame variations. 
    \item \textbf{STID}~\cite{shao2022spatial}:  It is a MLP-based spatio-temporal prediction model, which is simple yet effective. Its superior performance comes from the identification of the indistinguishability of samples in spatio-temporal dimensions. It demonstrates that it is promising to design efficient and effective models in spatio-temporal predictions.
    \item \textbf{STNorm}~\cite{deng2021st}: It proposed two types of normalization modules: spatial normalization and temporal normalization.  These two normalization methods can separately consider high-frequency components and local components.
    \item \textbf{PatchTST}~\cite{nie2022time}: It first employed patching and self-supervised learning in multivariate time series forecasting. It has two essential designs: (i) segmenting the original time series into patches to capture long-term correlations, (ii) different channels are operated independently, which share the same network. 
    \item \textbf{iTransformer}~\cite{liu2023itransformer}: This is the state-of-the-art multivariate time series model. Different from other Transformer-based methods, it employs the attention and feed-forward operation on an inverted dimension, that is, the multivariate correlation. 
    \item \textbf{MAML}~\cite{finn2017model}: Model-Agnostic Meta-Learning is an state-of-the-art meta learning technique. The main idea is to learn a good initialization from various tasks for the target task. 
    \item \textbf{MetaST}MetaST~\cite{yao2019learning}:  It is a urban transfer learning approach, which utilizes long-period data from multiple cities for transfer learning.  by employing a meta-learning approach, it learns a generalized network initialization adaptable to target cities. It also incorporates a pattern-based spatial-temporal memory to capture important patterns.
    \item \textbf{PromptST~\cite{zhang2023promptst}}:  It is the state-of-the-art pre-trianing and prompt-tuning approach for spatio-temporal prediction. 
\end{itemize}

\begin{table}[t!]
\caption{The basic statistics of the used datasets. 
}\label{tbl:dataset_st}
\begin{threeparttable}
\resizebox{1\columnwidth}{!}{
\begin{tabular}{ccccc}
\toprule
Dataset  & Min value & Max value & Mean value & Standard deviation \\ 
\cmidrule(lr){1-1} \cmidrule(lr){2-2}   \cmidrule(lr){3-3}    \cmidrule(lr){4-4}   \cmidrule(lr){5-5} 
TaxiBJ &  0.0 & 1285 & 107 & 133  \\ 
\cmidrule(lr){1-1} \cmidrule(lr){2-2}   \cmidrule(lr){3-3}    \cmidrule(lr){4-4}   \cmidrule(lr){5-5}  
Cellular & 0.0 & 2992532 & 75258 & 149505  \\ 
\cmidrule(lr){1-1} \cmidrule(lr){2-2}   \cmidrule(lr){3-3}    \cmidrule(lr){4-4}   \cmidrule(lr){5-5}  
TaxiNYC-1 & 0.0 & 1517 & 32 & 94\\ 
\cmidrule(lr){1-1} \cmidrule(lr){2-2}   \cmidrule(lr){3-3}    \cmidrule(lr){4-4}   \cmidrule(lr){5-5}   
TaxiNYC-2 & 0.0 & 1283 & 37 & 102 \\ 
\cmidrule(lr){1-1} \cmidrule(lr){2-2}   \cmidrule(lr){3-3}    \cmidrule(lr){4-4}   \cmidrule(lr){5-5}  
BikeNYC-1 & 0.0 & 266 & 9.2 &  18.1 \\ 
\cmidrule(lr){1-1} \cmidrule(lr){2-2}   \cmidrule(lr){3-3}    \cmidrule(lr){4-4}   \cmidrule(lr){5-5}  
BikeNYC-2 & 0.0 & 299 & 4.4 & 14.6\\ 
\cmidrule(lr){1-1} \cmidrule(lr){2-2}   \cmidrule(lr){3-3}    \cmidrule(lr){4-4}   \cmidrule(lr){5-5}   
TDrive &  0.0 & 2681 & 123 & 229 \\ 
\cmidrule(lr){1-1} \cmidrule(lr){2-2}   \cmidrule(lr){3-3}    \cmidrule(lr){4-4}   \cmidrule(lr){5-5}  
Crowd &  0.0 & 593118 & 21656 & 40825 \\ 
\cmidrule(lr){1-1} \cmidrule(lr){2-2}   \cmidrule(lr){3-3}    \cmidrule(lr){4-4}   \cmidrule(lr){5-5}   

TrafficCS &  0.0 & 22.25 & 6.22 & 4.79 \\
\cmidrule(lr){1-1} \cmidrule(lr){2-2}   \cmidrule(lr){3-3}    \cmidrule(lr){4-4}   \cmidrule(lr){5-5}  
TrafficWH &  0.0 & 22.35 & 6.22 & 4.68 \\
\cmidrule(lr){1-1} \cmidrule(lr){2-2}   \cmidrule(lr){3-3}    \cmidrule(lr){4-4}   \cmidrule(lr){5-5}    
TrafficCD & 0.0 & 22.25 &  7.33 & 4.36 \\
\cmidrule(lr){1-1} \cmidrule(lr){2-2}   \cmidrule(lr){3-3}    \cmidrule(lr){4-4}   \cmidrule(lr){5-5}  
TrafficJN & 0.0 & 25.04 & 5.72 & 4.71 \\
\cmidrule(lr){1-1} \cmidrule(lr){2-2}   \cmidrule(lr){3-3}    \cmidrule(lr){4-4}   \cmidrule(lr){5-5}   
TrafficNJ & 0.0 & 24.82 & 5.38 & 4.73\\
\cmidrule(lr){1-1} \cmidrule(lr){2-2}   \cmidrule(lr){3-3}    \cmidrule(lr){4-4}   \cmidrule(lr){5-5}  
TrafficSH & 0.0 & 21.83 & 7.92 & 3.86 \\
\cmidrule(lr){1-1} \cmidrule(lr){2-2}   \cmidrule(lr){3-3}    \cmidrule(lr){4-4}   \cmidrule(lr){5-5}  
TrafficSZ &  0.0 & 22.12 & 5.11 & 4.75  \\
\cmidrule(lr){1-1} \cmidrule(lr){2-2}   \cmidrule(lr){3-3}    \cmidrule(lr){4-4}   \cmidrule(lr){5-5}  
TrafficGZ & 0.0 & 25.16 & 5.26 & 4.79\\
\cmidrule(lr){1-1} \cmidrule(lr){2-2}   \cmidrule(lr){3-3}    \cmidrule(lr){4-4}   \cmidrule(lr){5-5}  
TrafficGY &   0.0 & 28.89 & 5.95 & 7.03 \\
\cmidrule(lr){1-1} \cmidrule(lr){2-2}   \cmidrule(lr){3-3}    \cmidrule(lr){4-4}   \cmidrule(lr){5-5}  
TrafficTJ & 0.0 & 25.24 & 6.32 & 5.05 \\
\cmidrule(lr){1-1} \cmidrule(lr){2-2}   \cmidrule(lr){3-3}    \cmidrule(lr){4-4}   \cmidrule(lr){5-5}  
TrafficHZ & 0.0 & 29.50 & 3.81 & 4.38\\

TrafficZZ &  0.0 & 23.26 & 6.67 & 4.32  \\
\cmidrule(lr){1-1} \cmidrule(lr){2-2}   \cmidrule(lr){3-3}    \cmidrule(lr){4-4}   \cmidrule(lr){5-5}  
TrafficBJ & 0.0 & 22.82 & 6.30 &  4.22  \\
\bottomrule
\end{tabular}}
\end{threeparttable}
\end{table}

\section{Algorithms}\label{sup:alg}

We provide the training algorithm for spatio-temporal  pre-trianing on multiple datasets in Algorithm~\ref{alg:train}. We also present the prompt fine-tuning algorithm in Algorithm~\ref{alg:finetune}.

\section{Implementation Details}\label{sup:imp_detail}

\subsection{Evaluation Metrics}
We use commonly used regression metrics, Mean Absolute Error (MAE) and Root Mean Squared Error (RMSE), to measure the prediction performance. 
Suppose $\bm{Y}=Y_1, ..., Y_M$ are ground truth for real spatio-temporal data, $\bm{\hat{Y}}=\hat{Y}_1, ..., \hat{Y}_N$ are the predicted values by the model, and $N$ is the number of total testing samples, These two metrics can be formulated as follows:

\begin{equation}
\begin{split}
    &\mathrm{RMSE}(\bm{Y}, \bm{\hat{Y}}) = \sqrt{\frac{1}{N}\sum_{i}^{N} \left ( Y_i - \hat{Y}_i \right )^2}, \\
    &\mathrm{MAE}(\bm{Y}, \bm{\hat{Y}}) = \frac{1}{N}\sum_{i}^{N} \left | Y_i - \hat{Y}_i \right |, \\
\end{split}
\end{equation}

\algrenewcommand\algorithmicprocedure{\textbf{Step}}
\algnewcommand{\LineComment}[1]{\State \(\triangleright\) #1}
\begin{algorithm}[t]
\caption{Spatio-temporal Pre-training}\label{alg:train}
\begin{algorithmic}[1]

\State \textbf{Input}:  Dataset $D=\{D_1, D_2, \ldots, D_{M}\}$, base  spatio-temporal prediction model $F$, and loss function $L$.
\State \textbf{Initialize}: Learnable parameters $\theta$ for the model $F$.
\For{$epoch\in\{1,2,\ldots,N_{iter}\}$}
    \State Randomly sample a dataset $D_m$ and a mini-batch $X$ from $D_m$. 
    \State Randomly choose a masking strategy $M$ from the four strategies.
    \State Mask the input $X$ into $X_m$
    \State Compute the reconstructions $\hat{y} \leftarrow F_{\theta}(X_m)$
    \State Compute the MSE loss $\mathcal{L} \leftarrow L(\hat{y},y)$
    \State Update the model's parameters $\theta \leftarrow update(\mathcal{L};\theta)$
\EndFor
\end{algorithmic}
\end{algorithm}

\algrenewcommand\algorithmicprocedure{\textbf{Step}}
\begin{algorithm}[t]
\caption{Prompt Tuning}\label{alg:finetune}
\begin{algorithmic}[1]

\State \textbf{Input}:  Dataset $D=\{D_1, D_2, \ldots, D_{M}\}$, parameters of pre-trained base model $\theta$, and loss function $L$
\State \textbf{Initialize}: Learnable parameters $\phi$ for the prompt network $G$.
\State Load the pretrained model parameters $\theta$.
\State Fix the parameters of the attention and feed-forward layers of the base model $F_\theta$. 
\For{$epoch\in\{1,2,\ldots,N_{iter}\}$}
    \State Randomly sample a dataset $D_m$ and a mini-batch $(X, Y)$ from $D_m$. 
    \State Generate the prompt $P$  for the mini-batch $P \leftarrow G{\phi}(X)$.
    \State Add the prompt to the input space $X_{p} = X + P$. 
    \State Compute the predictions $\hat{y} \leftarrow F_{\theta}(X_p)$
    \State Compute the MSE loss $\mathcal{L} \leftarrow L(\hat{y},Y)$
    \State Update the model's parameters $\gamma \leftarrow update(\mathcal{L};\gamma), \theta \leftarrow update(\mathcal{L};\theta)$
\EndFor
\end{algorithmic}
\end{algorithm}

\subsection{Parameter Settings}

\begin{table*}[t!]
\caption{The parameter details of UniST with different sizes evaluated in ablation studies.}\label{tbl:model_info}
\begin{threeparttable}
\resizebox{1.8\columnwidth}{!}{
\begin{tabular}{ccccc}
\toprule
Model & \#Encoder  Layers & \#Decoder Layers & Hidden Dimension (Encoder) & Decoder Hidden Dimension (Decoder)  \\ 
\cmidrule(lr){1-1} \cmidrule(lr){2-2}   \cmidrule(lr){3-3}    \cmidrule(lr){4-4}   \cmidrule(lr){5-5}
2M Params  & 2 &  2 &  64  & 64  \\ 
\cmidrule(lr){1-1} \cmidrule(lr){2-2}   \cmidrule(lr){3-3}    \cmidrule(lr){4-4}   \cmidrule(lr){5-5}
 8M Params & 4 &3   &  128 & 128   \\ 
\cmidrule(lr){1-1} \cmidrule(lr){2-2}   \cmidrule(lr){3-3}    \cmidrule(lr){4-4}     \cmidrule(lr){5-5}
10M Params &  6 & 4 & 128 & 128 \\ 
\cmidrule(lr){1-1} \cmidrule(lr){2-2}   \cmidrule(lr){3-3}    \cmidrule(lr){4-4}     \cmidrule(lr){5-5}
15M Params & 8 & 8  & 128 & 128     \\ 
\cmidrule(lr){1-1} \cmidrule(lr){2-2}   \cmidrule(lr){3-3}    \cmidrule(lr){4-4}     \cmidrule(lr){5-5}
30M Params & 6 & 6   &  256 & 256    \\ 
\bottomrule
\end{tabular}}
\end{threeparttable}
\end{table*}

Table~\ref{tbl:model_info} shows the parameter details of UniST with different sizes. During the training process, we used the Adam optimizer for gradient-based model optimization. The learning rate of the pre-training is set as 3e-4, and the learning rate of the prompt tuning is set as 5e-5.  The pre-training learning rate is selected via grid searching in a set of $\{1e-3,3e-4,1e-4\}$, and the fine-tuning learning rate is selected in a set of $\{1e-4,5e-5, 1e-5\}$.  Both in pre-training and fine-tuning, we evaluate the model's performance on the validation set every ten epochs ($\sim$all training instances).  We choose the model that performs best on the validation set for evaluations on the testing set.


\subsection{Prompt-Tuning}

The prompt-tuning stage aims to train a effective prompt network, which generates customized prompt for specific spatio-temporal pattern. We propose to leverage four types of spatio-temporal knowledge: (i) spatial closess ($s_c$), (ii) spatial hierarchy ($s_h$), (iii) temporal closeness ($t_c$), and (iv) temporal period ($t_p$). These knowledge-guided features are extracted from the input sequence. 
The input is the historical spatio-temporal sequence, the output is the predicted future spatio-temporal sequence, and the objective is to minimize the distance between the predicted results and real data.  Specifically, we use the widely adopted mean squared error loss function with $l_2$ regularization on the parameters in UniST to prevent over-fitting, which can be formulated as follows

\begin{equation}
    \mathcal{L} = \frac{1}{M}\sum(\hat{y}-y)^2+\lambda\sum_{\theta\in \Theta}\|\mathcal{\theta}\|_2
\end{equation}

\noindent where $\hat{y}$ and $y$ are ground truths and model predictions, respectively; $\Theta$ denotes the set that contains all model parameters.

\subsection{Baseline Implementation}

 We compare UniST with a broad collection of state-of-the-art models for spatio-temporal prediction, which can be categorized into five groups as introduced in Section~\ref{exp:exp_set}. If we consider the scalability to diverse data formats, i.e., different spatio-temporal data shapes, these baselines can be categorized into two groups: (i) approaches that are scalable with different spatio-temporal scales, such as PatchTST, MAML, and MetaST, and (ii) approaches that are non-scalable, including deep urban prediction approaches, video prediction approaches, and iTransformer. Most baselines are not scalable to different data shapes because they require a fixed number of spatial grids or variables, as seen in CNN-based approaches, MLP-based approaches, and multivariate time series models. Due to the varied data shapes, non-scalable baselines cannot be trained using all datasets, so we train separate models for each dataset.

For the scalable baseline, PatchTST~\cite{nie2022time}, it utilizes a channel-independent patch time series Transformer architecture, allowing it to be applied to datasets with varied spatio-temporal shapes. To ensure a fair comparison, we train both separate models and a single "one-for-all" model, as shown in Table~\ref{tbl:short_term}.

Notably, there are two baselines employ pretraining and finetuning: PatchTST~\cite{nie2022time} and PromptST~\cite{zhang2023promptst}. However, PromptST requires a fixed number of nodes, limiting its flexibility across different data formats. In contrast, the channel-independence of PatchTST allows it to handle varied data shapes. While PromptST is a state-of-the-art pre-training and prompt-tuning approach, it lacks generalization ability across different datasets.

\subsection{Experimental Design}

In our experimental design, we incorporate four distinct prediction tasks: short-term prediction, long-term prediction, few-shot prediction, and zero-shot prediction. This design aligns with established practices in foundation models for time series forecasting~\cite{zhou2023one,jin2023time,nie2022time}. The short-term and long-term prediction tasks are conducted without transfer learning settings. In these tasks, the model is trained on a set of $N$ datasets and then evaluated on the corresponding testing sets from these datasets. This setup enables us to directly assess the model's performance across multiple datasets using a single universal model.

Furthermore, the few-shot and zero-shot prediction tasks are designed to evaluate the model's generalization capabilities. In these tasks, the model learns from a set of source datasets to build a pretrained model and a memory pool, which is then utilized for prediction on target datasets. The key difference between the few-shot and zero-shot settings lies in the fine-tuning process on the target dataset. In few-shot prediction, the model undergoes a limited fine-tuning process using a small percentage of the target dataset's training data, while in zero-shot prediction, the model directly applies the pre-trained model and memory pool to make predictions on the target dataset without any fine-tuning.

These four tasks collectively offer a comprehensive evaluation of the model's performance and its ability to generalize across diverse spatio-temporal datasets.

\section{Additional Results}\label{sup:add_result}

\subsection{Analysis of Distribution Shifts}

\begin{figure}[t]
    \centering
    \includegraphics[width=\linewidth]{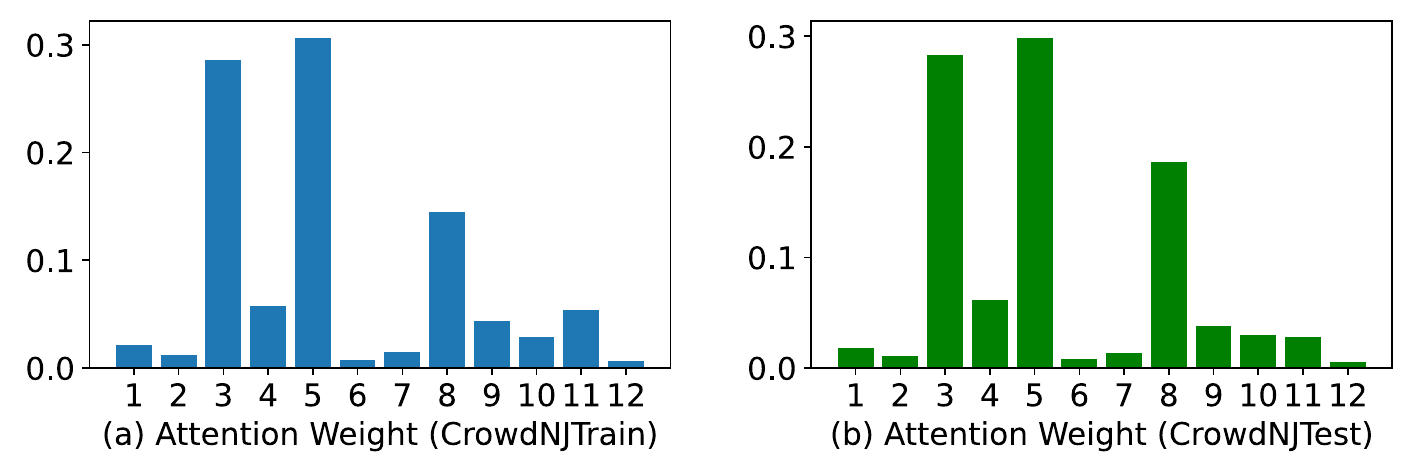}
    \vspace{-3mm}
    \caption{ Comparison of attention weight distribution  between the training set and testing set of the CrowdNJ dataset. The generated prompts assign attention weights on embeddings in the memory pool. }
    \label{fig:prompt_1}
\end{figure}

\begin{figure}[t]
    \centering
    \includegraphics[width=\linewidth]{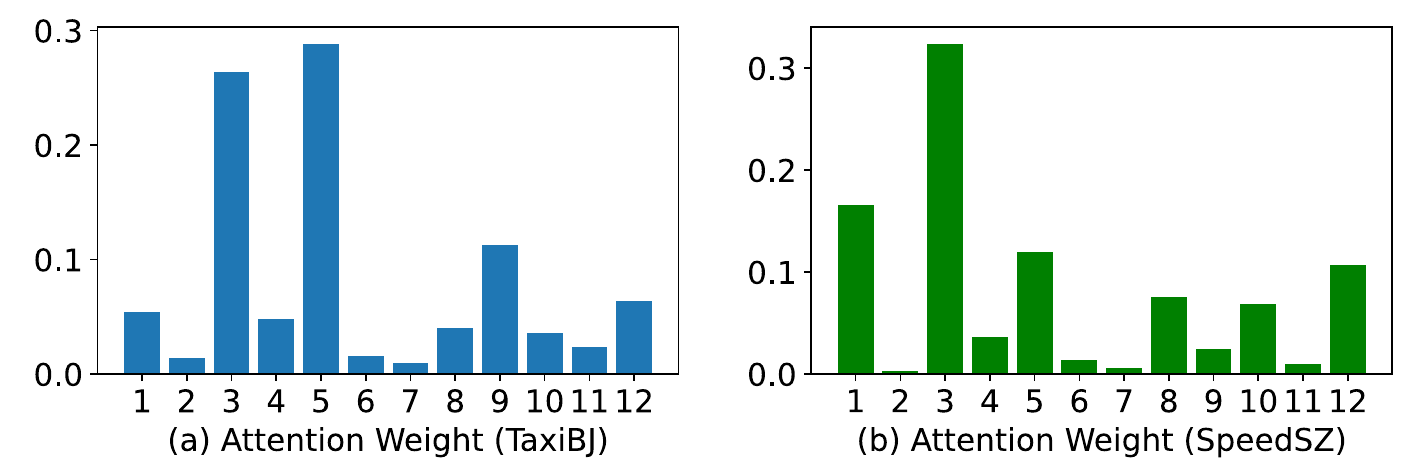}
    \vspace{-3mm}
    \caption{Comparison of attention weight distribution  between the the TaxiBJ dataset and SpeedSZ dataset. The generated prompts assign attention weights on embeddings in the memory pool.  }
    \label{fig:prompt_2}
\end{figure}

Here, we delve into a detailed analysis of the generated prompts across different datasets. For each dataset, we compute the attention weights on the embeddings in the memory pool and visualize the distribution of these weights in Figure~\ref{fig:prompt_1} to Figure~\ref{fig:prompt_2}. We have selected three typical scenarios to explore:

\begin{enumerate}[leftmargin=*]
    \item \textbf{Training and Testing Sets of One Dataset:} This analysis aims to investigate the model's ability to generalize within a familiar dataset. 
    \item \textbf{Two Datasets from Different Domains in the Same City:} Understanding how the model adapts its prompt generation across different but related datasets can provide insights into its domain-specific learning. 
    \item \textbf{Datasets from Different Cities and Domains:} This scenario highlights the model's ability to leverage knowledge learned previously and generate useful prompts adaptively. 
\end{enumerate}

\begin{figure}[t]
    \centering
    \includegraphics[width=\linewidth]{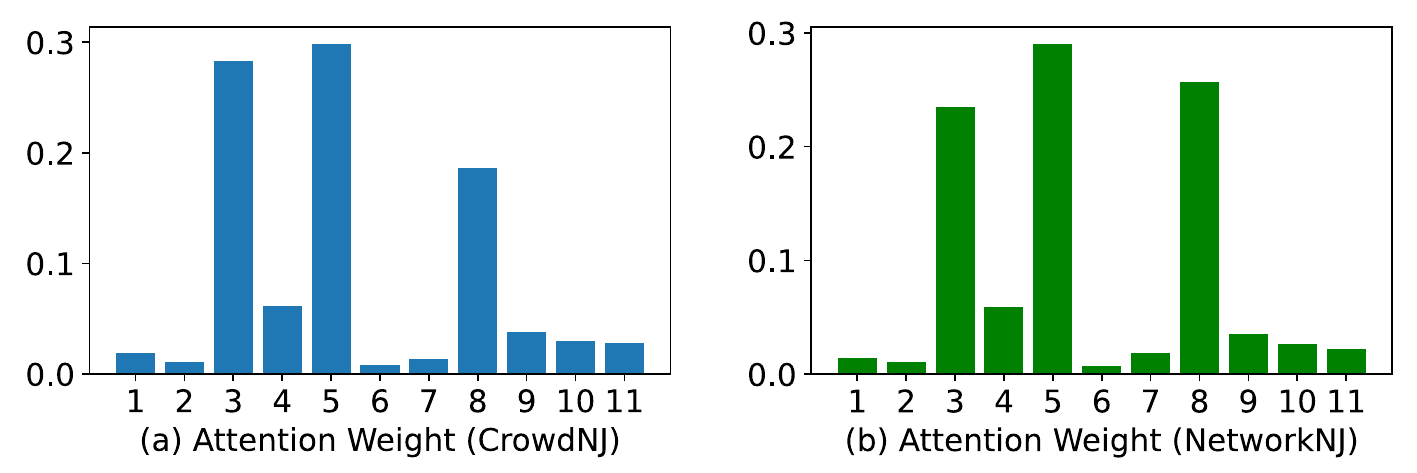}
    \vspace{-3mm}
    \caption{ Comparison of attention weight distribution  between the the CrowdNJ dataset and the CellularNJ dataset. The generated prompts assign attention weights on embeddings in the memory pool.  }
    \label{fig:prompt_3}
\end{figure}

As shown in Figure~\ref{fig:prompt_1} to Figure~\ref{fig:prompt_2}, our analysis reveals compelling insights into the effectiveness of our prompting mechanism in handling distribution shifts. Specifically, we observed that similar prompts are consistently generated for datasets exhibiting similar spatio-temporal patterns. For instance, the prompts generated for the training and testing sets of a single dataset, as well as for the testing sets of two datasets from different domains within the same city, are similar. This consistency in prompt generation suggests that our model effectively captures and leverages the underlying spatio-temporal patterns shared between these datasets. Meanwhile, our model generates distinct prompts for scenarios involving datasets from different cities and domains, indicating its ability to adapt to diverse spatio-temporal contexts. This adaptability is crucial for handling distribution shifts, as it allows the model to flexibly adjust its prompt generation strategy based on the unique characteristics of each dataset.

\begin{table}[t!]
\caption{Performance on different noise levels with sine-cosine positional encoding.}
\label{tbl:noise}
\begin{threeparttable}
\resizebox{1\columnwidth}{!}{
\begin{tabular}{cccccc}
\toprule
 & \multirow{2}{*}{TaxiBJ} & \multirow{2}{*}{Crowd} & \multirow{2}{*}{Cellular} & \multirow{2}{*}{BikeNYC} & \multirow{2}{*}{TrafficSH} \\
 Noise level & & & &  & \\
 \cmidrule(lr){1-1} \cmidrule(lr){2-2} \cmidrule(lr){3-3}  \cmidrule(lr){4-4} \cmidrule(lr){5-5} \cmidrule(lr){6-6} 
0     & 26.841  & 3.00  & 14.294    & 3.506    & 0.6650     \\
 \cmidrule(lr){1-1} \cmidrule(lr){2-2} \cmidrule(lr){3-3}  \cmidrule(lr){4-4} \cmidrule(lr){5-5} \cmidrule(lr){6-6} 
0.1\% & 26.846 & 3.038 & 14.297   & 3.507  & 0.6651    \\
 \cmidrule(lr){1-1} \cmidrule(lr){2-2} \cmidrule(lr){3-3}  \cmidrule(lr){4-4} \cmidrule(lr){5-5} \cmidrule(lr){6-6} 
1\%   & 26.90  & 3.039 & 14.390   & 3.534   & 0.6653    \\
 \cmidrule(lr){1-1} \cmidrule(lr){2-2} \cmidrule(lr){3-3}  \cmidrule(lr){4-4} \cmidrule(lr){5-5} \cmidrule(lr){6-6} 
10\%  & 28.76  & 3.29  & 14.91    & 3.695   & 0.6877   \\
 \cmidrule(lr){1-1} \cmidrule(lr){2-2} \cmidrule(lr){3-3}  \cmidrule(lr){4-4} \cmidrule(lr){5-5} \cmidrule(lr){6-6} 
Best baseline & 27.36 & 3.85 & 16.48 & 3.93 & 0.742\\
\bottomrule
\end{tabular}}
\end{threeparttable}
\end{table}

\begin{table}[t!]
\caption{Performance on different noise levels with learnable positional encoding.}
\label{tbl:noise_learn}
\begin{threeparttable}
\resizebox{1\columnwidth}{!}{
\begin{tabular}{cccccc}
\toprule
 & \multirow{2}{*}{TaxiBJ} & \multirow{2}{*}{Crowd} & \multirow{2}{*}{Cellular} & \multirow{2}{*}{BikeNYC} & \multirow{2}{*}{TrafficSH} \\
 Noise level & & & &  & \\
 \cmidrule(lr){1-1} \cmidrule(lr){2-2} \cmidrule(lr){3-3}  \cmidrule(lr){4-4} \cmidrule(lr){5-5} \cmidrule(lr){6-6} 
0  &  27.02  & 3.31 &  15.054  & 3.609  & 0.686  \\
 \cmidrule(lr){1-1} \cmidrule(lr){2-2} \cmidrule(lr){3-3}  \cmidrule(lr){4-4} \cmidrule(lr){5-5} \cmidrule(lr){6-6} 
0.1\%  & 27.032 & 3.310 & 15.068 & 3.607 & 0.6860\\
 \cmidrule(lr){1-1} \cmidrule(lr){2-2} \cmidrule(lr){3-3}  \cmidrule(lr){4-4} \cmidrule(lr){5-5} \cmidrule(lr){6-6} 
1\%  & 27.29 & 3.589 & 16.544 & 3.696 & 0.6911\\
 \cmidrule(lr){1-1} \cmidrule(lr){2-2} \cmidrule(lr){3-3}  \cmidrule(lr){4-4} \cmidrule(lr){5-5} \cmidrule(lr){6-6} 
10\%  & 43.80 & 11.436 & 70.360 & 8.173 & 1.228\\
 \cmidrule(lr){1-1} \cmidrule(lr){2-2} \cmidrule(lr){3-3}  \cmidrule(lr){4-4} \cmidrule(lr){5-5} \cmidrule(lr){6-6} 
Best baseline & 27.36 & 3.85 & 16.48 & 3.93 & 0.742\\
\bottomrule
\end{tabular}}
\end{threeparttable}
\end{table}

\begin{table*}[t!]
\caption{ Comparison of computational cost and memory cost between different approaches. The training time denotes the time cost to train all instances with one epoch.}
\label{tbl:cost}
\begin{threeparttable}
\resizebox{2.1\columnwidth}{!}{
\begin{tabular}{ccccccccccccccc}
\toprule
 Model   & STResNet & ACFM & STID & STNorm & STGSP & MC-STL & MAU & PredRNN & MIM & SimVP & TAU & PatchTST & iTransformer & UniST \\
\cmidrule(lr){1-1} \cmidrule(lr){2-2} \cmidrule(lr){3-3}  \cmidrule(lr){4-4} \cmidrule(lr){5-5} \cmidrule(lr){6-6}  \cmidrule(lr){7-7} \cmidrule(lr){8-8} \cmidrule(lr){9-9} \cmidrule(lr){10-10} \cmidrule(lr){11-11} \cmidrule(lr){12-12} \cmidrule(lr){13-13} \cmidrule(lr){14-14} \cmidrule(lr){15-15} 
Model Size (M)     &    2.51      &  1.90    &  1.63    &  1.15    &   5.51   &  6.35     &   10.55  &    17.07  &  26.24  &    9.96  &  9.55  &   2.59    &      25.27    &  6.71 \\
\cmidrule(lr){1-1} \cmidrule(lr){2-2} \cmidrule(lr){3-3}  \cmidrule(lr){4-4} \cmidrule(lr){5-5} \cmidrule(lr){6-6}  \cmidrule(lr){7-7} \cmidrule(lr){8-8} \cmidrule(lr){9-9} \cmidrule(lr){10-10} \cmidrule(lr){11-11} \cmidrule(lr){12-12} \cmidrule(lr){13-13} \cmidrule(lr){14-14} \cmidrule(lr){15-15} 
Memory Cost (MB) & 1475 & 1671 & 1715 & 2539 & 1459 &1607 & 1579 & 1065 & 1241 & 1039 & 1075 & 2859 & 2935 & 2875\\ 
\cmidrule(lr){1-1} \cmidrule(lr){2-2} \cmidrule(lr){3-3}  \cmidrule(lr){4-4} \cmidrule(lr){5-5} \cmidrule(lr){6-6}  \cmidrule(lr){7-7} \cmidrule(lr){8-8} \cmidrule(lr){9-9} \cmidrule(lr){10-10} \cmidrule(lr){11-11} \cmidrule(lr){12-12} \cmidrule(lr){13-13} \cmidrule(lr){14-14} \cmidrule(lr){15-15} 
Training Time (min)   &   0.057   &   0.561   &    0.054  &  0.461      &   0.078    &  0.311      &   0.828  &   0.455      &  0.836   &  0.224     &  0.224   &   0.338       &  0.093    & 1.4 (20+ datasets)      \\
\cmidrule(lr){1-1} \cmidrule(lr){2-2} \cmidrule(lr){3-3}  \cmidrule(lr){4-4} \cmidrule(lr){5-5} \cmidrule(lr){6-6}  \cmidrule(lr){7-7} \cmidrule(lr){8-8} \cmidrule(lr){9-9} \cmidrule(lr){10-10} \cmidrule(lr){11-11} \cmidrule(lr){12-12} \cmidrule(lr){13-13} \cmidrule(lr){14-14} \cmidrule(lr){15-15} 
Inference Time (min)  &   0.011       &   0.026   &   0.007   &    0.070    &   0.006    &    0.013    &  0.026   &   0.015      &   0.024  &  0.013     &  0.010   &     0.031     &       0.012  &  0.034  \\
\bottomrule
\end{tabular}}
\end{threeparttable}
\end{table*}

\begin{figure*}[t]
    \centering
    \includegraphics[width=0.8\linewidth]{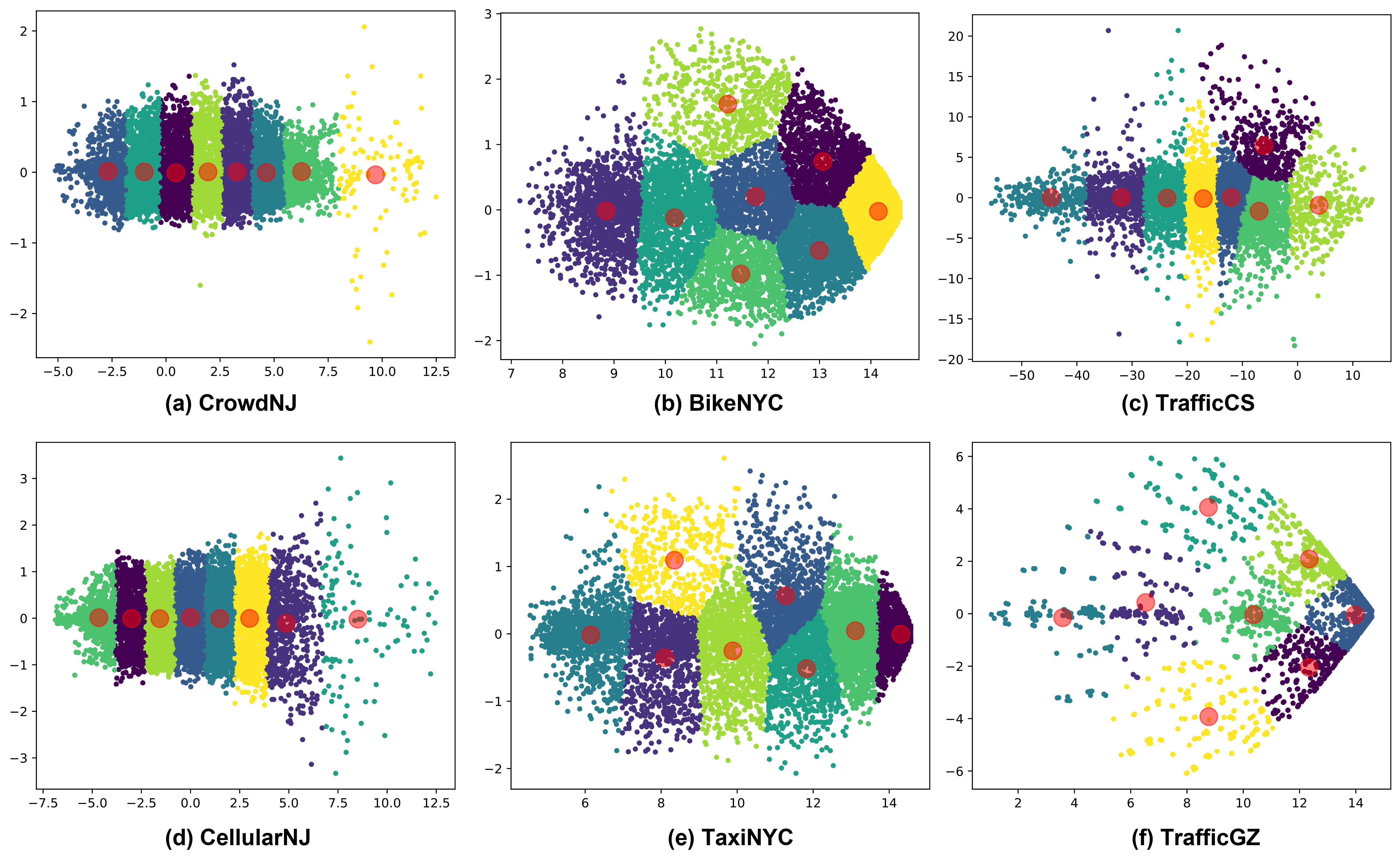}
    \caption{Visualization of different spatio-temporal datasets: Firstly, the high-dimensional data is reduced to a two-dimensional vector using t-SNE. Subsequently, the embeddings are visualized in clusters using the k-means clustering method.}
    \label{fig:tsne}
\end{figure*}

\subsection{Performance under Noise Perturbations}

The model's ability  to handle noisy data is necessary to ensure reliable predictions. Therefore, we conduct experiments to evaluate UniST's robustness against noisy data. Specifically, we introduced Gaussian noise with varying levels of intensity to the input data and assessed UniST's performance under these conditions. We considered three levels of noise: Gaussian noise randomly sampled from a 0.1\% normal distribution, Gaussian noise randomly sampled from a 1\% normal distribution, and Gaussian noise randomly sampled from a 10\% normal distribution. These noise levels represent varying degrees of data corruption, simulating real-world scenarios where data can be noisy or contain irregularities.

The results, as detailed in Table~\ref{tbl:noise}, demonstrate that UniST consistently outperforms baseline models even in the presence of noise perturbations (where the best baseline has no noise perturbation). This suggests that UniST is capable of effectively handling noisy data, which is crucial for ensuring reliable predictions, especially in real-world scenarios where data can be messy or contain irregularities.

Moreover, we examine how different positional encoding methods affect the model's robustness. We compare the use of two positional encoding methods: learnable embeddings and sine-cosine encoding. The results in Table~\ref{tbl:noise_learn} show the performance with learnable embeddings, while Table~\ref{tbl:noise} shows the performance with sine-cosine encoding. Comparing these two sets of results, we observe that sine-cosine encoding exhibits more robust performance against noise perturbations. Specifically, learnable embeddings show a significant performance reduction with increased noise perturbation and perform worse than the best baseline model.

\subsection{Model Efficiency}

Table~\ref{tbl:cost} shows a detailed comparison of the computational and memory costs of UniST against baselines.  The results show that the model size and memory cost of UniST are comparable to those of other approaches. However, due to the multiple data pre-training involved, the training time of UniST is longer compared to other methods. Despite this, UniST consistently outperforms baselines on all datasets with just one model. Thus, we consider the additional training time acceptable given the superior performance achieved.

\subsection{Dataset Similarity}

To assess the similarities among the datasets used in our study, we employed a two-step process. First, we reduced the dimension of the spatio-temporal data using t-SNE, a technique for dimension reduction. This allowed us to visualize the datasets in a lower-dimensional space. Second, we applied the k-means clustering method to the reduced data to identify clusters of similar spatio-temporal patterns.

The results of our visualization revealed interesting insights. We found that certain datasets, such as the Crowd data and Cellular data in Nanjing, exhibited similar spatio-temporal patterns. Similarly, the Bike data and Taxi data in New York City showed similarities in their patterns. However, most datasets from different cities or domains exhibited distinct spatio-temporal patterns, indicating significant distribution shifts. These observations highlight the powerful generalization ability and universality of our approach across datasets with significantly distinct spatio-temporal patterns.

\begin{table}[t!]
\caption{ Ablation studies on four masking strategies.}
\label{tbl:ablation_mask}
\begin{threeparttable}
\resizebox{1\columnwidth}{!}{
\begin{tabular}{cccc}
\toprule
 & Prediction & Imputation &  Spatial extrapolation  \\
  \cmidrule(lr){1-1} \cmidrule(lr){2-2} \cmidrule(lr){3-3}  \cmidrule(lr){4-4} 
Complete  & 0.781  & 0.761 & 0.729 \\
 \cmidrule(lr){1-1} \cmidrule(lr){2-2} \cmidrule(lr){3-3}  \cmidrule(lr){4-4} 
wo/ Random masking  & 0.796  & 1.72 &  0.761 \\
 \cmidrule(lr){1-1} \cmidrule(lr){2-2} \cmidrule(lr){3-3}  \cmidrule(lr){4-4} 
wo/ Tube masking  & 0.787 & 0.788 & 0.817 \\
 \cmidrule(lr){1-1} \cmidrule(lr){2-2} \cmidrule(lr){3-3}  \cmidrule(lr){4-4}  
wo/ Block masking   & 0.785  & 0.773 &  1.02\\
 \cmidrule(lr){1-1} \cmidrule(lr){2-2} \cmidrule(lr){3-3}  \cmidrule(lr){4-4}
wo/ Temporal masking   & 1.44  & 0.772 & 0.742  \\
\bottomrule
\end{tabular}}
\end{threeparttable}
\end{table}

\begin{table*}[t!]
\caption{Performance comparison of short-term prediction  on seven datasets in terms of MAE and RMSE. We use the average prediction errors over all prediction steps. 
}
\label{tbl:short1}
\resizebox{2.1\columnwidth}{!}{
\begin{tabular}{ccccccccccccccc}
\toprule
& \multicolumn{2}{c}{\textbf{TaxiNYC-1}} & \multicolumn{2}{c}{\textbf{BikeNYC-2}} & \multicolumn{2}{c}{\textbf{TaxiNYC-2}} & \multicolumn{2}{c}{\textbf{TrafficBJ}}  & \multicolumn{2}{c}{\textbf{TrafficNJ}} & \multicolumn{2}{c}{\textbf{TrafficWH}} &\multicolumn{2}{c}{\textbf{TrafficSZ}} \\ 
 \cmidrule(lr){2-3} \cmidrule(lr){4-5} \cmidrule(lr){6-7} \cmidrule(lr){8-9} \cmidrule(lr){10-11} \cmidrule(lr){12-13}  \cmidrule(lr){14-15}
 \textbf{Model} & \textbf{RMSE}  & \textbf{MAE}    & \textbf{RMSE}       & \textbf{MAE}      & \textbf{RMSE}       & \textbf{MAE}      & \textbf{RMSE}    & \textbf{MAE}      & \textbf{RMSE}    & \textbf{MAE}    & \textbf{RMSE}  & \textbf{MAE}         & \textbf{RMSE}  & \textbf{MAE}        \\ 
\cmidrule(lr){1-1} \cmidrule(lr){2-3} \cmidrule(lr){4-5} \cmidrule(lr){6-7} \cmidrule(lr){8-9} \cmidrule(lr){10-11} \cmidrule(lr){12-13}  \cmidrule(lr){14-15}

  HA & 57.07 & 18.57 & 15.68 &  7.17 & 52.84 & 15.74 & 1.033& 0.582&	1.593& 0.774&	1.351& 0.645&	0.791& 0.416 \\
 ARIMA  & 55.39 & 20.94 & 25.01 & 13.63 & 62.9 & 29.56  &  1.32&  0.735& 1.30 & 0.709 & 1.51 & 0.748 & 0.821 & 0.445  \\
\cmidrule(lr){1-1} \cmidrule(lr){2-3} \cmidrule(lr){4-5} \cmidrule(lr){6-7} \cmidrule(lr){8-9} \cmidrule(lr){10-11} \cmidrule(lr){12-13}   \cmidrule(lr){14-15}

 STResNet & 29.45& 17.96&	7.18& 3.94	&22.16& 12.06& 0.828& 0.547&	1.03& 0.635&	0.903& 0.568&	0.709& 0.465 \\
 ACFM  & 23.35& 11.54 &	5.99& 3.094  &	14.48 & 6.39 &   0.706& 0.44&	0.888& 0.515&	0.784& 0.471&	0.573& 0.35 \\
 STID & 17.75& 7.03 &	5.70& 2.711 &	17.37 & 7.35 &0.724& 0.431&	0.847& 0.459&	0.78& 0.436&	0.576& 0.33   \\
 STNorm & 21.26 & 8.14&	6.47 & 3.03 &	19.02& 7.17 &  0.727& 0.428&	0.904& 0.476&	0.81& 0.445&	0.666& 0.369\\
 STGSP & 28.13 & 10.29 &14.20 & 7.38 & 29.10 & 10.14  &  0.736& 0.444&	0.883& 0.491&	0.804& 0.473&	0.86& 0.52  \\
 MC-STL &  18.44&9.51 & 6.26 & 3.40 & 16.78 & 8.50 & 0.975& 0.709&	1.13 & 0.78 &	1.1& 0.773&	0.83& 0.615\\
\cmidrule(lr){1-1} \cmidrule(lr){2-3} \cmidrule(lr){4-5} \cmidrule(lr){6-7} \cmidrule(lr){8-9} \cmidrule(lr){10-11} \cmidrule(lr){12-13}   \cmidrule(lr){14-15}
MAU & 28.70 & 11.23 &6.12	& 2.95 &19.38 & 7.27 & 1.12& 0.797&	0.978& 0.545&	1.37 & 0.917&	0.826& 0.523  \\
PredRNN & 16.53 & 5.80 & 6.47 & 3.08 & 19.89 & 7.23 & 0.651& 0.376&	0.852& 0.457&	0.74& 0.421&	0.58& 0.335 \\
MIM  &  18.83 & 6.866& 6.36 & 2.89 &	18.02& 6.56  & 2.62 & 2.14 &	4.65 & 3.39 &	3.86 & 3.15 &	2.22 & 1.40  \\
SimVP & 16.63 & 7.51& 5.96 & 2.92 &	15.10 & 6.54 & 0.664& 0.408&	0.861& 0.481&	0.779& 0.475&	0.583& 0.359\\
TAU  & 16.91 & 6.85&	5.98 & 2.89 &	15.35 & 6.80 & 0.70 & 0.44 &	0.89 & 0.528&	0.747& 0.444&	0.576& 0.353\\
\cmidrule(lr){1-1} \cmidrule(lr){2-3} \cmidrule(lr){4-5} \cmidrule(lr){6-7} \cmidrule(lr){8-9} \cmidrule(lr){10-11} \cmidrule(lr){12-13}   \cmidrule(lr){14-15}
PatchTST & 41.34 & 13.10  & 12.33 &  5.30 & 37.76 &  11.13 &  0.935& 0.512&	1.379& 0.658&	1.17 & 0.561&	0.718& 0.370  \\
iTransformer  & 36.73 &13.11  & 9.86 & 4.50 &33.03 & 11.22 & 0.876& 0.490 &	1.18 & 0.60 &	1.10 & 0.542&	0.718& 0.378 \\
\cmidrule(lr){1-1} \cmidrule(lr){2-3} \cmidrule(lr){4-5} \cmidrule(lr){6-7} \cmidrule(lr){8-9} \cmidrule(lr){10-11} \cmidrule(lr){12-13}   \cmidrule(lr){14-15}
PatchTST(one-for-all) &  44.43 & 14.56 & 13.62 & 6.03  & 41.04 & 12.61 &  0.964 & 0.524 & 1.42 & 0.675 & 1.22 & 0.581 & 0.739 & 0.375 \\
\cmidrule(lr){1-1} \cmidrule(lr){2-3} \cmidrule(lr){4-5} \cmidrule(lr){6-7} \cmidrule(lr){8-9} \cmidrule(lr){10-11} \cmidrule(lr){12-13}   \cmidrule(lr){14-15}
UniST (ours)  &  15.32 & 5.65 & 5.50 & 2.56 & 12.71 & 4.82 & 0.689 & 0.387  & 0.845 & 0.421 &0.762 & 0.396 & 0.513 & 0.264 \\ 
\bottomrule
\end{tabular}}
\end{table*}

\subsection{Additional Ablation Studies}

\subsubsection{Masking Strategies.}

We investigated the contribution of each of the four masking strategies  by comparing the performance when all four strategies are employed with the performance when one of the strategies is removed. We conducted experiments on three spatio-temporal tasks: prediction, imputation, and spatial extrapolation, using the TrafficCD dataset.

\begin{figure}[t]
    \centering
    \includegraphics[width=\linewidth]{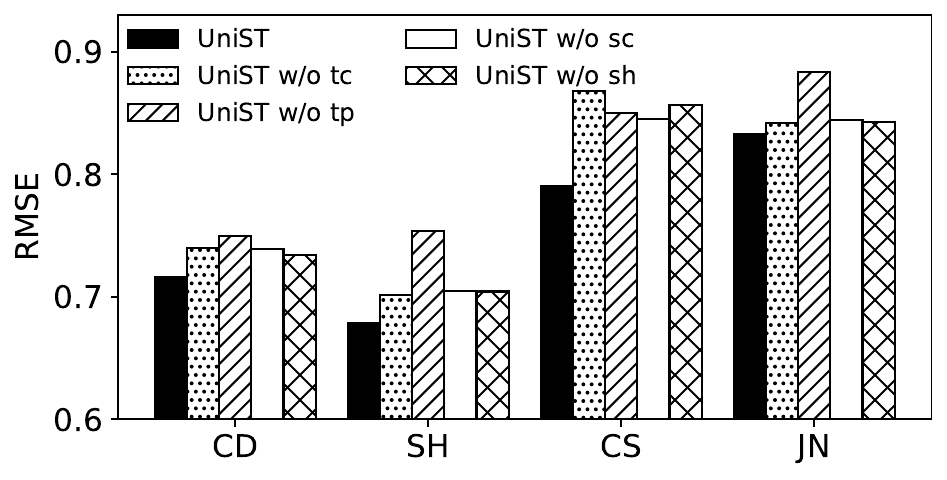}
    \caption{Ablation studies on four types spatial and temporal knowledge extraction $t_c, t_p, s_c$, and $s_h$. }
    \label{fig:ablation_fine}
\end{figure}

The results, shown in Table~\ref{tbl:ablation_mask}, indicate that training with all four masking strategies achieved the best performance across all three tasks. Removing the temporal masking strategy results in the most significant performance decrease for the prediction task, removing the random masking strategy leads to the most significant performance decrease for the imputation task, and removing the block masking strategy results in the most significant performance decrease for the spatial extrapolation task. These results are reasonable as each masking strategy is designed to align with a specific task objective.

It is worth noting that despite the seemingly mismatched nature of some masking strategies with certain spatio-temporal tasks (e.g., random masking vs. prediction, temporal masking vs. imputation, and temporal masking vs. spatial extrapolation), we find that these masking strategies still contribute to the performance of less related tasks. This indicates that the masking strategies not only benefit their intended tasks but also have broader effects on the model's  general learning of spatio-temporal dependencies and dynamics.  For example, while random masking may seem unrelated to causal prediction tasks, it can help the model learn robust features that generalize well across different time points.   Additionally, temporal masking can help the model better understand the temporal dynamics when performing spatial extrapolation.

\begin{table*}[t!]
\vspace{+5mm}
\caption{Performance comparison of short-term prediction  on seven datasets in terms of MAE and RMSE. We use the average prediction errors over all prediction steps. 
}
\label{tbl:short2}
\resizebox{2.1\columnwidth}{!}{
\begin{tabular}{ccccccccccccccc}
\toprule
& \multicolumn{2}{c}{\textbf{TrafficTJ}} & \multicolumn{2}{c}{\textbf{TrafficGY}} & \multicolumn{2}{c}{\textbf{TrafficGZ}} & \multicolumn{2}{c}{\textbf{TrafficZZ}}  & \multicolumn{2}{c}{\textbf{TrafficCS}} & \multicolumn{2}{c}{\textbf{TrafficCD}} &\multicolumn{2}{c}{\textbf{TrafficHZ}} \\ 
 \cmidrule(lr){2-3} \cmidrule(lr){4-5} \cmidrule(lr){6-7} \cmidrule(lr){8-9} \cmidrule(lr){10-11} \cmidrule(lr){12-13}  \cmidrule(lr){14-15}
 \textbf{Model} & \textbf{RMSE}  & \textbf{MAE}    & \textbf{RMSE}       & \textbf{MAE}      & \textbf{RMSE}       & \textbf{MAE}      & \textbf{RMSE}    & \textbf{MAE}      & \textbf{RMSE}    & \textbf{MAE}    & \textbf{RMSE}  & \textbf{MAE}         & \textbf{RMSE}  & \textbf{MAE}        \\ 
\cmidrule(lr){1-1} \cmidrule(lr){2-3} \cmidrule(lr){4-5} \cmidrule(lr){6-7} \cmidrule(lr){8-9} \cmidrule(lr){10-11} \cmidrule(lr){12-13}  \cmidrule(lr){14-15}
 
  HA & 1.61 & 0.824&	1.79 & 0.726&	0.996& 0.52&	1.47 & 0.857&	1.31 & 0.676&	1.12 & 0.668&	0.765& 0.342 \\
 ARIMA  & 2.02 & 1.59 & 1.91 & 1.16 &  1.37 & 0.76 & 1.78 & 0.998 & 1.66 & 0.923 & 1.54 & 0.907 & 0.803 & 0.364  \\
\cmidrule(lr){1-1} \cmidrule(lr){2-3} \cmidrule(lr){4-5} \cmidrule(lr){6-7} \cmidrule(lr){8-9} \cmidrule(lr){10-11} \cmidrule(lr){12-13}   \cmidrule(lr){14-15}

 STResNet & 1.12 & 0.714&	1.32 & 0.799&	0.796& 0.515&	1.03& 0.693&	0.986& 0.651&	0.867& 0.576&	0.669& 0.406 \\
 ACFM  &0.959& 0.574&	1.10 & 0.599&	0.701& 0.418&	0.839& 0.526&	0.842& 0.529&	0.757& 0.493&	0.575& 0.316   \\
 STID & 0.976& 0.549&	1.04 & 0.544&	0.665& 0.362&	0.838& 0.502&	0.855& 0.5&	0.715& 0.44&	0.546& 0.282  \\
 STNorm & 0.973& 0.533&	1.12& 0.508&	0.693& 0.373&	0.885& 0.538&	0.91& 0.511&	0.786& 0.489&	0.556& 0.260  \\
  STGSP & 0.989& 0.572&	1.09 & 0.649&	0.733& 0.419&	0.831& 0.505&	0.978& 0.587&	0.776& 0.497&	0.616& 0.331  \\
 MC-STL & 1.22 & 0.856&	1.82 & 1.36 &	1.04 & 0.775&	1.14 & 0.81 &	1.14 & 0.819&	1.00 & 0.733&	0.842& 0.606\\
\cmidrule(lr){1-1} \cmidrule(lr){2-3} \cmidrule(lr){4-5} \cmidrule(lr){6-7} \cmidrule(lr){8-9} \cmidrule(lr){10-11} \cmidrule(lr){12-13}   \cmidrule(lr){14-15}
MAU & 0.988& 0.549&	1.14 & 0.595&	0.74& 0.415&	1.42& 0.934&	1.31& 0.791&	1.25& 0.919&	0.743& 0.377 \\
PredRNN & 0.971& 0.53&	1.16 & 0.608&	0.71& 0.42&	0.853& 0.508&	0.909& 0.572&	0.815& 0.513&	0.602& 0.288 \\
MIM  & 3.44 & 2.51 &	5.68 & 4.53&	3.43 & 2.80 &	2.05 & 1.56 &	3.57 & 2.71 &	2.75& 2.26 &	1.92 & 1.23  \\
SimVP & 1.00 & 0.597&	1.13 & 0.632&	0.667& 0.399&	0.838& 0.526&	0.835& 0.507&	0.775& 0.495&	0.549& 0.301\\
TAU  & 1.01& 0.606&	1.11& 0.604&	0.65& 0.378&	0.839& 0.527&	0.869& 0.543&	0.768& 0.495&	0.539& 0.289 \\
\cmidrule(lr){1-1} \cmidrule(lr){2-3} \cmidrule(lr){4-5} \cmidrule(lr){6-7} \cmidrule(lr){8-9} \cmidrule(lr){10-11} \cmidrule(lr){12-13}   \cmidrule(lr){14-15}
PatchTST & 1.44 & 0.722&	1.58 & 0.634&	0.894& 0.448&	1.31 & 0.742&	1.18 & 0.599&	1.00& 0.577&	0.696& 0.305  \\
iTransformer  & 1.26 & 0.675&	1.39& 0.621&	0.846& 0.428&	1.19 & 0.696&	1.09 & 0.572&	0.941& 0.541&	0.66& 0.30 \\
\cmidrule(lr){1-1} \cmidrule(lr){2-3} \cmidrule(lr){4-5} \cmidrule(lr){6-7} \cmidrule(lr){8-9} \cmidrule(lr){10-11} \cmidrule(lr){12-13}   \cmidrule(lr){14-15}
PatchTST(one-for-all) & 1.49 & 0.740 & 1.66 & 0.684 & 0.931 & 0.469 & 1.35 & 0.752 & 1.23 & 0.620 & 1.04 & 0.602 &  0.726 & 0.325  \\
\cmidrule(lr){1-1} \cmidrule(lr){2-3} \cmidrule(lr){4-5} \cmidrule(lr){6-7} \cmidrule(lr){8-9} \cmidrule(lr){10-11} \cmidrule(lr){12-13}   \cmidrule(lr){14-15}
UniST (ours)  & 0.958 & 0.510 & 1.03 & 0.458 & 0.648 & 0.325 & 0.832 & 0.482  & 0.791 & 0.423  & 0.711 & 0.415 & 0.530 & 0.236 \\ 
\bottomrule
\end{tabular}}
\end{table*}

\begin{table*}[t!]
\vspace{+5mm}
\caption{Performance comparison of long-term prediction on four datasets in terms of MAE and RMSE. We use the average prediction errors over all prediction steps. 
}
\label{tbl:long1}
\resizebox{1.6\columnwidth}{!}{
\begin{tabular}{ccccccccc}
\toprule
& \multicolumn{2}{c}{\textbf{TaxiBJ}} & \multicolumn{2}{c}{\textbf{Cellular}} & \multicolumn{2}{c}{\textbf{BikeNYC-2}} & \multicolumn{2}{c}{\textbf{TDrive}}  \\ 
 \cmidrule(lr){2-3} \cmidrule(lr){4-5} \cmidrule(lr){6-7} \cmidrule(lr){8-9} 
 \textbf{Model} & \textbf{RMSE}  & \textbf{MAE}    & \textbf{RMSE}       & \textbf{MAE}      & \textbf{RMSE}       & \textbf{MAE}      & \textbf{RMSE}    & \textbf{MAE}        \\ 
\cmidrule(lr){1-1} \cmidrule(lr){2-3} \cmidrule(lr){4-5} \cmidrule(lr){6-7} \cmidrule(lr){8-9} 
  HA      &  74.07 & 43.79 & 77.29 & 31.89 & 15.84 & 7.97 & 144.65 & 72.48  \\
 ARIMA    &  100.76 & 56.04 & 83.66 & 35.96  & 15.29 & 7.25 & 270.05 & 140.80\\
\cmidrule(lr){1-1} \cmidrule(lr){2-3} \cmidrule(lr){4-5} \cmidrule(lr){6-7} \cmidrule(lr){8-9}
 STResNet & 51.36 & 36.08 & 33.87 & 20.87 &   12.73 & 7.16 &  163.88 & 112.27 \\
 ACFM     & 35.49& 22.46 & 26.40 & 13.24 & 13.00 & 7.09 &  88.76 & 42.19\\
 STID     & 36.98 & 23.19 & 22.98 & 11.71 &   12.75 & 8.37 &  83.70& 37.66 \\
 STNorm   & 33.78 & 19.89 &71.05 &32.14 & 12.16 & 5.99 & 100.43 & 49.50 \\
  STGSP   & 70.31 & 42.76 & 67.07 & 31.16 &   14.50 & 7.66 &  83.70 & 37.26 \\
 MC-STL   & 38.23 & 26.86 & 39.74 & 27.04 &  12.72& 7.96 &  100.55 & 59.18 \\
\cmidrule(lr){1-1} \cmidrule(lr){2-3} \cmidrule(lr){4-5} \cmidrule(lr){6-7} \cmidrule(lr){8-9} 
MAU &   85.58& 60.61 & 75.84 & 32.78 &   12.42 & 5.82 &  137.17 & 76.17  \\
PredRNN & 43.89 & 27.42 & 46.68 & 24.96 &    9.72 & 4.37 &  175.32 & 104.79   \\
MIM  &  38.10 & 25.82 &     79.20 & 39.27 &   10.02 & 4.60 & 107.06 & 43.67     \\
SimVP &  33.53 & 19.28 & 23.84 & 12.90 &   10.89 & 5.51 & 91.13& 39.46    \\
TAU  &  34.88 & 19.94 & 23.00 & 12.72 &   11.53 & 6.11   & 91.54 & 41.96   \\
\cmidrule(lr){1-1} \cmidrule(lr){2-3} \cmidrule(lr){4-5} \cmidrule(lr){6-7} \cmidrule(lr){8-9} 
PatchTST & 30.64 & 17.49 & 23.39 & 12.42 & 11.13 & 5.07 & 92.03 & 38.89\\
PatchTST(one-for-all) &  31.58 & 18.67 & 27.94 & 10.89 & 10.71 & 4.74 & 111.56 & 50.57   \\
iTransformer & 32.89 & 18.60 & 29.329 & 11.963 & 11.54 & 5.19 & 93.87 & 40.16\\
\cmidrule(lr){1-1} \cmidrule(lr){2-3} \cmidrule(lr){4-5} \cmidrule(lr){6-7} \cmidrule(lr){8-9} 
UniST (ours) & 30.46 & 17.95  &  20.64 & 10.43 & 11.91 & 5.06 & 90.60 & 37.01 \\ 
\bottomrule
\end{tabular}
}
\end{table*}

\subsubsection{Knowledge-Guide Prompts.}

The prompts play an essential role in our UniST model. Here we investigate whether the designed spatial and temporal properties $s_c, s_h, t_c$, and $t_p$ contribute to the final performance. We use $s_c$ to denote spatial closeness, $s_h$ to denote spatial hierarchy, $t_p$ for temporal periodicity, and $t_c$ for temporal closeness.

we compare the overall design that incorporates all three properties with four degraded versions that individually remove $s_c, s_h, t_c$, or $t_p$ . 
Figure~\ref{fig:ablation_fine} shows the results on four traffic speed datasets. As we can observe, removing any property results in a performance decrease. 
The contributions of each spatial and temporal property vary across different datasets, highlighting the necessity of each property for the spatio-temporal design.

\subsection{Additional Prediction Results}

Table~\ref{tbl:short1}$\sim$Table~\ref{tbl:few_shot_taxibj} report addition prediction results.

\begin{table*}[t!]
\vspace{+10mm}
\caption{Performance comparison in few-shot and zero-shot (only UniST)  settings on the Crowd dataset in terms of MAE and RMSE. 1\% , 5\%, and 10\% denote that only the percentage of training data is utilized.  We use the average prediction errors over all prediction steps. 
}\label{tbl:few_shot_crowd}
\begin{threeparttable}
\resizebox{1.2\columnwidth}{!}{
\begin{tabular}{ccccccc}
\toprule
& \multicolumn{2}{c}{\textbf{10\%}} & \multicolumn{2}{c}{\textbf{5\%}} & \multicolumn{2}{c}{\textbf{1\%}}  \\ 
 \cmidrule(lr){2-3} \cmidrule(lr){4-5}  \cmidrule(lr){6-7} 
 \textbf{Model} & \textbf{RMSE}  & \textbf{MAE}    & \textbf{RMSE}    & \textbf{MAE}   & \textbf{RMSE}    & \textbf{MAE}     \\ 
\cmidrule(lr){1-1} \cmidrule(lr){2-3} \cmidrule(lr){4-5} \cmidrule(lr){6-7} 

ATFM & 19.842 & 11.446  & 19.923 & 11.687 & 21.166 & 12.643 \\
STNorm & 14.668 & 7.050 &  14.884 & 7.723 & 35.959 & 29.585 \\
STID  & 14.676 & 7.280 & 14.975 & 8.671 & 25.905 & 19.610 \\
\cmidrule(lr){1-1} \cmidrule(lr){2-3} \cmidrule(lr){4-5}  \cmidrule(lr){6-7} 
PredRNN & 19.604 & 9.668 & 20.186 & 10.190 & 24.901 & 13.142 \\
SimVP & 14.093 & 7.101 & 14.167 & 8.550 &  14.252 & 8.776 \\
TAU &14.229 & 7.140 & 14.456 & 8.411 & 14.919 & 9.096  \\
\cmidrule(lr){1-1} \cmidrule(lr){2-3} \cmidrule(lr){4-5}  \cmidrule(lr){6-7} 
MAML & 14.089 & 7.180 & 14.795 & 8.154 & 14.334 & 8.608\\
MetaST & 13.801 & 6.847 &  14.220 & 7.442 & 14.242& 7.949\\
PatchTST & 14.060 & 6.787  & 14.142 &  6.811 & 14.491 &  7.227 \\
\cmidrule(lr){1-1} \cmidrule(lr){2-3} \cmidrule(lr){4-5}  \cmidrule(lr){6-7} 
UniST (few-shot)  &13.411 & 6.365 & 13.859 & 6.542 & 13.952 & 6.581  \\
UniST (zero-shot)  & 14.665 & 7.051  & 14.665 & 7.051 & 14.665 & 7.051\\
\bottomrule
\end{tabular}}
\end{threeparttable}
\end{table*}

\begin{table*}[t!]
\vspace{+8mm}
\caption{Performance comparison in few-shot and zero-shot (only UniST)  settings on the BikeNYC dataset in terms of MAE and RMSE. 1\% , 5\%, and 10\% denote that only the percentage of training data is utilized.  We use the average prediction errors over all prediction steps. 
}\label{tbl:few_shot_bike}
\begin{threeparttable}
\resizebox{1.2\columnwidth}{!}{
\begin{tabular}{ccccccc}
\toprule
& \multicolumn{2}{c}{\textbf{10\%}} & \multicolumn{2}{c}{\textbf{5\%}} & \multicolumn{2}{c}{\textbf{1\%}}  \\ 
 \cmidrule(lr){2-3} \cmidrule(lr){4-5}  \cmidrule(lr){6-7} 
 \textbf{Model} & \textbf{RMSE}  & \textbf{MAE}    & \textbf{RMSE}    & \textbf{MAE}   & \textbf{RMSE}    & \textbf{MAE}     \\ 
\cmidrule(lr){1-1} \cmidrule(lr){2-3} \cmidrule(lr){4-5} \cmidrule(lr){6-7} 

ATFM & 8.026 & 3.511 & 10.438 &  4.582 &  11.876 & 5.990\\
STNorm & 7.42 & 2.70 & 10.21 & 4.17 & 12.94 & 5.20\\
STID  &  6.97 & 3.49 & 12.46 & 7.56 & 15.08 & 9.38 \\
\cmidrule(lr){1-1} \cmidrule(lr){2-3} \cmidrule(lr){4-5}  \cmidrule(lr){6-7} 
PredRNN &11.05 & 4.00 & 11.29 & 4.46 & 12.58 & 4.75\\
SimVP &6.570 & 2.691 & 8.525 & 3.174 & 8.661 & 3.721  \\
TAU & 7.06 & 3.07 & 8.74 & 3.28 & 8.50 & 3.72 \\
\cmidrule(lr){1-1} \cmidrule(lr){2-3} \cmidrule(lr){4-5}  \cmidrule(lr){6-7} 
MAML & 6.49 & 2.31 & 8.89 & 3.68 & 8.98 & 3.91\\ 
MetaST &6.21 & 2.18 & 8.22 & 3.03 & 8.58 & 3.60 \\
PatchTST & 9.14 & 2.68 & 10.09 & 2.88 &  9.74 & 3.86 \\
\cmidrule(lr){1-1} \cmidrule(lr){2-3} \cmidrule(lr){4-5}  \cmidrule(lr){6-7} 
UniST  & 5.318 & 1.668 & 6.113 & 1.964 & 7.811 & 2.72\\
UniST (zero-shot)  & 9.06 & 3.63 & 9.06 & 3.63 &  9.06 & 3.63 \\
\bottomrule
\end{tabular}}
\end{threeparttable}
\end{table*}

\begin{table*}[t!]
\vspace{+10mm}
\caption{Performance comparison in few-shot and zero-shot (only UniST) settings on the TaxiBJ dataset in terms of MAE and RMSE. 1\% , 5\%, and 10\% denote that only the percentage of training data is utilized.  We use the average prediction errors over all prediction steps. 
}\label{tbl:few_shot_taxibj}
\begin{threeparttable}
\resizebox{1.2\columnwidth}{!}{
\begin{tabular}{ccccccc}
\toprule
& \multicolumn{2}{c}{\textbf{10\%}} & \multicolumn{2}{c}{\textbf{5\%}} & \multicolumn{2}{c}{\textbf{1\%}}  \\ 
 \cmidrule(lr){2-3} \cmidrule(lr){4-5}  \cmidrule(lr){6-7} 
 \textbf{Model} & \textbf{RMSE}  & \textbf{MAE}    & \textbf{RMSE}    & \textbf{MAE}   & \textbf{RMSE}    & \textbf{MAE}     \\ 
\cmidrule(lr){1-1} \cmidrule(lr){2-3} \cmidrule(lr){4-5} \cmidrule(lr){6-7} 

ATFM & 50.631 & 33.035 & 55.770 & 39.205 & 64.590 & 44.928 \\
STNorm &39.35 & 22.48 & 42.67 & 26.78 & 44.76 & 28.24 \\
STID  & 34.53 & 20.54 & 37.39 & 24.35 & 47.94 & 31.94 \\
\cmidrule(lr){1-1} \cmidrule(lr){2-3} \cmidrule(lr){4-5}  \cmidrule(lr){6-7} 
PredRNN & 84.28 & 58.52 & 97.74 & 73.40 & 92.21 & 66.76 \\
SimVP & 35.114 & 20.87 & 37.42 & 23.131 & 40.465 & 24.95 \\
TAU & 37.70 & 22.69 & 39.77 & 25.73 & 41.98 & 26.48 \\
\cmidrule(lr){1-1} \cmidrule(lr){2-3} \cmidrule(lr){4-5}  \cmidrule(lr){6-7} 
MAML & 36.24 & 20.91 & 36.12 & 23.47 & 40.11 & 24.79 \\ 
MetaST & 35.42 & 18.65 & 35.21 & 21.74 & 39.08 & 23.88\\
PatchTST &  44.03& 22.69 & 44.24 & 22.62 & 46.43 & 24.77  \\
\cmidrule(lr){1-1} \cmidrule(lr){2-3} \cmidrule(lr){4-5}  \cmidrule(lr){6-7} 
UniST  & 27.59 & 15.18 & 31.19 & 17.58 & 35.09 & 20.62\\
UniST (zero-shot)  & 51.4 & 33.1 &  51.4 & 33.1 & 51.4 & 33.1 \\
\bottomrule
\end{tabular}}
\end{threeparttable}
\end{table*}




\end{document}